\documentclass[journal]{IEEEtran}

\usepackage{cite}
\usepackage[nice]{nicefrac}
\usepackage{amsmath,amsfonts} 
\usepackage{graphicx,xcolor}
\usepackage{tabularx,booktabs} 
\usepackage{import}
\usepackage{times,xspace} 
\usepackage{ntheorem}
\usepackage{mathtools}
\usepackage{romannum}
\usepackage{siunitx}
\usepackage{xargs}
\usepackage{balance}
\usepackage{url}
\usepackage{leftidx}
\usepackage[font={footnotesize}]{subcaption}
\usepackage[font={footnotesize}]{caption}
\usepackage{tcolorbox}
\usepackage{algorithm}
\usepackage{setspace}
\usepackage{algpseudocode} 
\usepackage{adjustbox}
\usepackage{enumitem}
\usepackage{makecell}
\usepackage{svg}
\usepackage{multirow}  
\usepackage{stmaryrd} 
\makeatletter

\usepackage[bottom]{footmisc}

\usepackage{pifont}
\newcommand{\cmark}{\ding{51}}%
\newcommand{\xmark}{\ding{55}}%

\definecolor{gridcolor}{RGB}{220,220,220}
\definecolor{lightgray}{RGB}{230,230,230}
\definecolor{darkergray}{RGB}{100,100,100}
\definecolor{background}{RGB}{239,239,239}
\definecolor{reachgray}{RGB}{180,180,180}
\definecolor{reachgraylight}{RGB}{144,144,144}

\newtheorem{definition}{Definition}

\newtheorem{rem}{Remark}
\newcommand{\sledom}{\Relbar\joinrel\mathrel{|}}

\renewtheoremstyle{plain}
{\item{\theorem@headerfont ##1\ ##2\theorem@separator}~}
{\item{\theorem@headerfont ##1\ ##2\ (##3)\theorem@separator}~}
\makeatother
\newtcolorbox{mytransparentbox}[1][]{%
	opacityback=0.65,
	opacityframe=0.0,
	colback=white,
	left=0pt,
	right=0pt,
	top=0pt,
	bottom=0pt,
	boxrule=0pt,
	halign=center,
	valign=center,
	#1
}
\usepackage{hyperref}
\algrenewcommand\algorithmicrequire{\textbf{Input:}}
\algrenewcommand\algorithmicensure{\textbf{Output:}}
\algnewcommand\algorithmicinput{\textbf{Optional Input:}}
\algnewcommand\INPUT{\item[\algorithmicinput]}
\newif\ifdraft
\draftfalse  

\ifdraft
\else
\fi
\usepackage[symbols, toc,nonumberlist, sanitizesort]{glossaries}

\newglossary[glignoredl]{ignored}{glignored}{glignoredin}{Ignored Glossary}

\newglossary{time}{gls1}{glo1}{Time}
\newglossary{longitudinalplanning}{gls2}{glo2}{Longitudinal Trajectory Planning}
\newglossary{lateralplanning}{gls3}{glo3}{Lateral Trajectory Planning}
\newglossary{cosy}{gls4}{glo4}{Coordinate Systems}
\newglossary{misc}{gls5}{glo5}{Miscellaneous}
\newglossary{state}{gls6}{glo6}{State Variables of the Ego Vehicle}
\newglossary{refpath}{gls7}{glo7}{Reference Path}
\newglossary{shape}{gls8}{glo8}{Ego Vehicle Shape}
\newglossary{obs}{gls9}{glo9}{Obstacles}
\newglossary{reach}{gls10}{glo10}{Reachability Analysis}
\newglossary{drivingcorridor}{gls11}{glo11}{Driving Corridor Identification}
\newglossary{constraints}{gls12}{glo12}{Collision Avoidance Constraints}

\makenoidxglossaries
\glsnoexpandfields

\newcommand{\shorteq}{%
	\settowidth{\@tempdima}{-}%
	\resizebox{\@tempdima}{\height}{=}%
}

\newcommand{\ndec}{\glslink{ndec}{\alpha}}
\newglossaryentry{ndec}{
	name={\ensuremath{\ndec}},
	description={Number of tactical decisions},
	sort=5 a,
	type=misc
}

\newcommand{\nobs}{\glslink{nobs}{\sigma}}
\newglossaryentry{nobs}{
	name={\ensuremath{\nobs}},
	description={Number of obstacles},
	sort=5 b,
	type=misc
}

\newcommand{\cartesian}{\mathtt{G}}
\newcommand{\local}{\mathtt{L}}
\newcommand{\vehiclefixed}{\mathtt{V}}
\newcommandx{\cosy}[1]{F^{#1}}
\newcommand{\transformation}{T}

\newcommand{\curvframe}{\glslink{curvframe}{\cosy{\local}}}
\newglossaryentry{curvframe}{
	name={\ensuremath{\curvframe}},
	description={Local, curvilinear coordinate frame},
	sort=b,
	type=cosy
}

\newcommandx{\localvar}[1]{\glslink{localvar}{\leftidx{^{\local}}{#1}}}
\newglossaryentry{localvar}{
	name={\ensuremath{\localvar{\square}}},
	description={Variable $\square$ in $\curvframe$},
	parent=curvframe,
	sort=b,
	type=cosy
}

\newcommand{\cartesianframe}{\glslink{cartesianframe}{\cosy{\cartesian}}}
\newglossaryentry{cartesianframe}{
	name={\ensuremath{\cartesianframe}},
	description={Global, Cartesian coordinate frame},
	sort=b,
	type=cosy
}

\newcommandx{\globalvar}[1]{\glslink{globalvar}{\leftidx{^{\cartesian}}{#1}}}
\newglossaryentry{globalvar}{
	name={\ensuremath{\globalvar{\square}}},
	description={Variable $\square$ in $\cartesianframe$},
	parent=cartesianframe,
	sort=b,
	type=cosy
}

\newcommand{\vehicleframe}{\glslink{vehicleframe}{\cosy{\vehiclefixed}}}
\newglossaryentry{vehicleframe}{
	name={\ensuremath{\vehicleframe}},
	description={Vehicle-fixed coordinate frame},
	sort=b,
	type=cosy
}

\newcommand{\curvframelondir}{\glslink{curvframelondir}{\zeta}}
\newglossaryentry{curvframelondir}{
	name={\ensuremath{\curvframelondir}},
	description={Longitudinal direction in $\curvframe$},
	parent=curvframe,
	sort=b,
	type=cosy
}

\newcommand{\curvframelatdir}{\glslink{curvframelatdir}{\eta}}
\newglossaryentry{curvframelatdir}{
	name={\ensuremath{\curvframelatdir}},
	description={Lateral direction in $\curvframe$},
	parent=curvframe,
	sort=b,
	type=cosy
}

\newcommandx{\transformationcurvcartesian}[1][1={\slon}, usedefault]{\glslink{transformationcurvcartesian}{\transformation_{\local}^{\cartesian}\left(#1\right)}}
\newglossaryentry{transformationcurvcartesian}{
	name={\ensuremath{\transformationcurvcartesian}},
	description={Transformation from $\curvframe$ to $\cartesianframe$},
	sort=b b,
	type=cosy
}

\newcommandx{\transformationvehiclecartesian}[1][1=\orientation, usedefault]{\glslink{transformationvehiclecartesian}{\transformation_{\vehiclefixed}^{\cartesian}\left(#1\right)}}
\newglossaryentry{transformationvehiclecartesian}{
	name={\ensuremath{\transformationvehiclecartesian}},
	description={Transformation from $\vehicleframe$ to $\cartesianframe$},
	sort=b b,
	type=cosy
}

\newcommand{\maxval}[1]{\glslink{maxval}{\overline{#1}}}
\newglossaryentry{maxval}{
	name={\ensuremath{\maxval{\square}}},
	description={Maximum admissible value of a variable $\square$},
	sort=6a,
	type=state
}

\newcommand{\minval}[1]{\glslink{minval}{\underline{#1}}}
\newglossaryentry{minval}{
	name={\ensuremath{\minval{\square}}},
	description={Minimum admissible value of a variable $\square$},
	sort=6b,
	type=state
}

\newcommand{\pos}{s}
\newcommand{\velocity}{v}
\newcommand{\acceleration}{a}
\newcommand{\jerk}{j}

\newcommand{\slon}{\glslink{slon}{\pos_{\curvframelondir}}}
\newglossaryentry{slon}{
	name={\ensuremath{\slon}},
	description={Longitudinal position in $\curvframe$},
	sort=6c,
	type=state
}
\newcommandx{\slonk}[1][1=\timestep, usedefault]{
	\glslink{slonk}{\pos_{{\curvframelondir, #1}}}}
\newglossaryentry{slonk}{
	name={\ensuremath{\slonk}},
	description={Longitudinal position $\slon$ at time step $\timestep$},
	sort=6c,
	parent=slon,
	type=state
}

\newcommand{\dslon}{\glslink{dslon}{\dot{\pos}_{\curvframelondir}}}
\newglossaryentry{dslon}{
	name={\ensuremath{\dslon}},
	description={Time derivative of longitudinal position in $\curvframe$},
	type=ignored,
}

\newcommand{\sminlon}{\glslink{sminlon}{\minval{\pos}_{\curvframelondir}}}
\newglossaryentry{sminlon}{
	name={\ensuremath{\sminlon}},
	description={Minimum value of $\slon$},
	sort=6c,
	type=state,
	parent=slon
}

\newcommandx{\sminlonk}[1][1=\timestep, usedefault]{\glslink{sminlonk}{\minval{\pos}_{\curvframelondir, #1}}}
\newglossaryentry{sminlonk}{
	name={\ensuremath{\sminlonk}},
	description={Minimum value of $\slon$ at time step $\timestep$},
	sort=6c,
	type=state,
	parent=sminlon
}

\newcommand{\smaxlon}{\glslink{smaxlon}{\maxval{\pos}_{\curvframelondir}}}
\newglossaryentry{smaxlon}{
	name={\ensuremath{\smaxlon}},
	description={Maximum value of $\slon$},
	sort=6c,
	type=state,
	parent=slon
}

\newcommandx{\smaxlonk}[1][1=\timestep, usedefault]{\glslink{smaxlonk}{\maxval{\pos}_{\curvframelondir, #1}}}
\newglossaryentry{smaxlonk}{
	name={\ensuremath{\smaxlonk}},
	description={Maximum value of $\slon$ at time step $\timestep$},
	sort=6c,
	type=state,
	parent=smaxlon
}

\newcommand{\vlon}{\glslink{vlon}{\velocity_{\curvframelondir}}}
\newglossaryentry{vlon}{
	name={\ensuremath{\vlon}},
	description={Longitudinal velocity in $\curvframe$},
	sort=6d,
	type=state
}
\newcommandx{\vlonk}[1][1=\timestep, usedefault]{\glslink{vlonk}{\velocity_{\curvframelondir, #1}}}
\newglossaryentry{vlonk}{
	name={\ensuremath{\vlonk}},
	description={Longitudinal velocity $\vlon$ at time step $\timestep$},
	sort=6d,
	type=state,
	parent=vlon
}

\newcommand{\vminlon}{\glslink{vminlon}{\minval{\velocity}_{\curvframelondir}}}
\newglossaryentry{vminlon}{
	name={\ensuremath{\vminlon}},
	description={Minimum value of $\vlon$},
	sort=6d,
	type=state,
	parent=vlon
}

\newcommand{\vmaxlon}{\glslink{vmaxlon}{\maxval{\velocity}_{\curvframelondir}}}
\newglossaryentry{vmaxlon}{
	name={\ensuremath{\vmaxlon}},
	description={Maximum value of $\vlon$},
	sort=6d,
	type=state,
	parent=vlon
}

\newcommand{\alon}{\glslink{alon}{\acceleration_{\curvframelondir}}}
\newglossaryentry{alon}{
	name={\ensuremath{\alon}},
	description={Longitudinal acceleration in $\curvframe$},
	sort=6e,
	type=state,
}

\newcommandx{\alonk}[1][1=\timestep, usedefault]{\acceleration_{\curvframelondir, #1}}
\newglossaryentry{alonk}{
	name={\ensuremath{\alonk}},
	description={Longitudinal acceleration $\alon$ at time step $\timestep$},
	sort=6e,
	type=state,
	parent=alon
}

\newcommand{\aminlon}{\glslink{aminlon}{\minval{\acceleration}_{\curvframelondir}}}
\newglossaryentry{aminlon}{
	name={\ensuremath{\aminlon}},
	description={Minimum value of $\alon$},
	sort=6e,
	type=state,
	parent=alon
}

\newcommand{\amaxlon}{\glslink{amaxlon}{\maxval{\acceleration}_{\curvframelondir}}}
\newglossaryentry{amaxlon}{
	name={\ensuremath{\amaxlon}},
	description={Maximum value of $\alon$},
	sort=6e,
	type=state,
	parent=alon
}

\newcommand{\jlon}{\glslink{jlon}{\jerk_{\curvframelondir}}}
\newglossaryentry{jlon}{
	name={\ensuremath{\jlon}},
	description={Longitudinal jerk in $\curvframe$},
	sort=6f,
	type=state,
}

\newcommandx{\jlonk}[1][1=\timestep, usedefault]{\glslink{jlonk}{\jerk_{\curvframelondir, #1}}}
\newglossaryentry{jlonk}{
	name={\ensuremath{\jlonk}},
	description={Longitudinal jerk $jlonk$ at time step $\timestep$},
	sort=6f,
	type=state,
	parent=jlon
}

\newcommand{\slat}{\glslink{slat}{\pos_{\curvframelatdir}}}
\newglossaryentry{slat}{
	name={\ensuremath{\slat}},
	description={Lateral position in $\curvframe$},
	sort=6g,
	type=state
}

\newcommandx{\slatk}[1][1=\timestep, usedefault]{
	\glslink{slatk}{\pos_{{\curvframelatdir, #1}}}}
\newglossaryentry{slatk}{
	name={\ensuremath{\slatk}},
	description={Lateral position $\slat$ at time step $\timestep$},
	sort=6g,
	type=state,
	parent=slat
}

\newcommand{\dslat}{\glslink{dslat}{\dot{\pos}_{\curvframelatdir}}}
\newglossaryentry{dslat}{
	name={\ensuremath{\dslat}},
	description={Time derivative of lateral position in $\curvframe$},
	type=ignored
}

\newcommand{\vlat}{\glslink{vlat}{\velocity_{\curvframelatdir}}}
\newglossaryentry{vlat}{
	name={\ensuremath{\vlat}},
	description={Lateral velocity in $\curvframe$},
	sort=6h,
	type=state
}

\newcommandx{\vlatk}[1][1=\timestep, usedefault]{\glslink{vlatk}{\velocity_{\curvframelatdir, #1}}}
\newglossaryentry{vlatk}{
	name={\ensuremath{\vlatk}},
	description={Lateral velocity $\vlat$ at time step $\timestep$},
	sort=6h,
	type=state,
	parent=vlat
}

\newcommand{\vminlat}{\glslink{vminlat}{\minval{\velocity}_{\curvframelatdir}}}
\newglossaryentry{vminlat}{
	name={\ensuremath{\vminlat}},
	description={Minimum value of $\vlat$},
	sort=6h,
	type=state,
	parent=vlat
}

\newcommand{\vmaxlat}{\glslink{vmaxlat}{\maxval{\velocity}_{\curvframelatdir}}}
\newglossaryentry{vmaxlat}{
	name={\ensuremath{\vmaxlat}},
	description={Minimum value of $\vlat$},
	sort=6h,
	type=state,
	parent=vlat
}

\newcommand{\alat}{\glslink{alat}{\acceleration_{\curvframelatdir}}}
\newglossaryentry{alat}{
	name={\ensuremath{\alat}},
	description={Lateral acceleration in $\curvframe$},
	sort=6i,
	type=state
}

\newcommandx{\alatk}[1][1=\timestep, usedefault]{\acceleration_{\curvframelatdir, #1}}
\newglossaryentry{alatk}{
	name={\ensuremath{\alatk}},
	description={Lateral acceleration $\alatk$ at time step $\timestep$},
	sort=6i,
	type=state,
	parent=alat
}

\newcommand{\aminlat}{\glslink{aminlat}{\minval{\acceleration}_{\curvframelatdir}}}
\newglossaryentry{aminlat}{
	name={\ensuremath{\aminlat}},
	description={Minimum value of $\alat$},
	sort=6i,
	type=state,
	parent=alat
}

\newcommand{\amaxlat}{\glslink{amaxlat}{\maxval{\acceleration}_{\curvframelatdir}}}
\newglossaryentry{amaxlat}{
	name={\ensuremath{\amaxlat}},
	description={Maximum value of $\alat$},
	sort=6i,
	type=state,
	parent=alat
}

\newcommand{\sx}{\glslink{sx}{\pos_{x}}}
\newglossaryentry{sx}{
	name={\ensuremath{\sx}},
	description={Longitudinal position in $\cartesianframe$},
	sort=6j,
	type=state
}

\newcommand{\sy}{\glslink{sy}{\pos_{y}}}
\newglossaryentry{sy}{
	name={\ensuremath{\sy}},
	description={Lateral position in $\cartesianframe$},
	sort=6k,
	type=state
}

\newcommand{\orientation}{\glslink{orientation}{\theta}}
\newglossaryentry{orientation}{
	name={\ensuremath{\orientation}},
	description={Orientation in $\cartesianframe$},
	sort=6l,
	type=state
}

\newcommandx{\orientationk}[1][1=\timestep, usedefault]{\glslink{orientationk}{\theta_{#1}}}
\newglossaryentry{orientationk}{
	name={\ensuremath{\orientationk}},
	description={Orientation $\orientation$ at time step $\timestep$},
	sort=6l,
	type=state,
	parent=orientation
}

\newcommand{\curvature}{\glslink{curvature}{\kappa}}
\newglossaryentry{curvature}{
	name={\ensuremath{\curvature}},
	description={Curvature in $\cartesianframe$},
	sort=6l,
	type=state
}

\newcommandx{\curvaturek}[1][1=\timestep, usedefault]{\curvature_{#1}}
\newglossaryentry{curvaturek}{
	name={\ensuremath{\curvaturek}},
	description={Curvature $\curvature$ at time step $\timestep$},
	sort=6l,
	type=state,
	parent=curvature
}

\newcommand{\refpathsym}{\Gamma}
\newcommandx{\refpath}[1][1=\slon, usedefault]{\glslink{refpath}{\refpathsym(#1)}}
\newglossaryentry{refpath}{
	name={\ensuremath{\refpath}},
	description={Reference path},
	sort=7a,
	type=refpath
}

\newcommand{\reforientation}{\glslink{reforientation}{\orientation_\refpathsym}}
\newglossaryentry{reforientation}{
	name={\ensuremath{\reforientation}},
	description={Orientation of reference path},
	sort=7b,
	type=refpath
}

\newcommand{\refcurvature}{\glslink{refcurvature}{\curvature_\refpathsym}}
\newglossaryentry{refcurvature}{
	name={\ensuremath{\refcurvature}},
	description={Curvature of reference path},
	sort=7c,
	type=refpath
}

\newcommand{\minrefcurvature}{\glslink{minrefcurvature}{\minval{\curvature}_\refpathsym}}
\newglossaryentry{minrefcurvature}{
	name={\ensuremath{\minrefcurvature}},
	description={Minimum curvature of reference path},
	sort=7d,
	type=refpath
}

\newcommand{\maxrefcurvature}{\glslink{maxrefcurvature}{\maxval{\curvature}_\refpathsym}}
\newglossaryentry{maxrefcurvature}{
	name={\ensuremath{\maxrefcurvature}},
	description={Maximum curvature of reference path},
	sort=7f,
	type=refpath
}

\newcommand{\idxcentercircle}{i}
\newcommand{\centerpoint}{c}

\newcommand{\wheelbase}{\glslink{wheelbase}{\ell}}
\newglossaryentry{wheelbase}{
	name={\ensuremath{\wheelbase}},
	description={Wheelbase},
	sort=8a,
	type=shape
}

\newcommand{\radius}{\glslink{radius}{r}}
\newglossaryentry{radius}{
	name={\ensuremath{\radius}},
	description={Radius of the circles approximating the vehicle shape},
	sort=8b,
	type=shape
}

\newcommandx{\centercircle}[2][1=\idxcentercircle, 2=\timestep, usedefault]{\glslink{centercircle}{\centerpoint^{(#1)}_{#2}}\glslink{centercirclek}{}}
\newglossaryentry{centercircle}{
	name={\ensuremath{\centercircle[][ ]}},
	description={Center of the $\idxcentercircle$-th circle approximating the vehicle shape},
	sort=8c,
	type=shape
}
\newglossaryentry{centercirclek}{
	name={\ensuremath{\centercircle}},
	description={Center $\centercircle$ at time step $\timestep$},
	sort=8c,
	type=shape,
	parent=centercircle,
}

\newcommandx{\globalcentercirclek}[2][1=\idxcentercircle, 2=\timestep, usedefault]{\glslink{globalcentercirclek}{\globalvar{\centerpoint}^{(#1)}_{#2}}}
\newglossaryentry{globalcentercirclek}{
	name={\ensuremath{\globalcentercirclek}},
	description={Center $\centercircle$ in $\cartesianframe$ at time step $\timestep$},
	sort=8c,
	type=shape,
	parent=centercircle
}

\newcommandx{\globalcentercirclerayk}[2][1=\idxcentercircle, 2=\timestep, usedefault]{\glslink{globalcentercirclerayk}{\globalvar{\hat{\centerpoint}}^{(#1)}_{#2}}}
\newglossaryentry{globalcentercirclerayk}{
	name={\ensuremath{\globalcentercirclerayk}},
	description={Center $\centercircle$ in $\cartesianframe$ at time step $\timestep$},
	sort=8c,
	type=shape,
	parent=centercircle
}

\newcommandx{\localcentercirclek}[2][1=\idxcentercircle, 2=\timestep, usedefault]{\glslink{localcentercirclek}{\localvar{\centerpoint}^{(#1)}_{#2}}}
\newglossaryentry{localcentercirclek}{
	name={\ensuremath{\localcentercirclek}},
	description={Center $\centercircle$ in $\curvframe$ at time step $\timestep$},
	sort=8c,
	type=shape,
	parent=centercircle
}

\newcommand{\conttime}{\glslink{time}{t}}
\newglossaryentry{time}{
	name={\ensuremath{\conttime}},
	description={Time},
	sort=1 a,
	type=time,
}

\newcommand{\dt}{\glslink{dt}{\Delta \conttime}}
\newglossaryentry{dt}{
	name={\ensuremath{\dt}},
	description={Time increment},
	sort=1 b,
	type=time,
}

\newcommand{\thorizon}{\glslink{thorizon}{\timestep_f}}
\newglossaryentry{thorizon}{
	name={\ensuremath{\thorizon}},
	description={Final time step},
	sort=1 c,
	type=time
}

\newcommand{\tinit}{\glslink{tinit}{\timestep_0}}
\newglossaryentry{tinit}{
	name={\ensuremath{\tinit}},
	description={Initial time step},
	sort=1 d,
	type=time
}

\newcommand{\timestep}{\glslink{timestep_b}{k}}
\newglossaryentry{timestep_a}{
	name=\ensuremath{\timestep},
	description={Discrete time},
	sort=1 f,
	type=time,
}
\newglossaryentry{timestep_b}{
	name={\ensuremath{\square_k}},
	description={Variable $\square$ at time step $\timestep$},
	parent=timestep_a,
	sort=1 g,
	type=time,
}

\newcommand{\lon}{\mathrm{lon}}
\newcommand{\lat}{\mathrm{lat}}
\newcommand{\R}{\mathcal{R}}

\newcommand{\systemmatrix}{\glslink{systemmatrix}{A}}
\newglossaryentry{systemmatrix}{
	name={\ensuremath{\systemmatrix}},
	description={System matrix},
	sort=c e
}

\newcommand{\inputmatrix}{\glslink{inputmatrix}{B}}
\newglossaryentry{inputmatrix}{
	name={\ensuremath{\inputmatrix}},
	description={Input matrix},
	sort=c f
}

\newcommand{\x}{\glslink{x}{x}}
\newglossaryentry{x}{
	name={\ensuremath{\x}},
	description={State},
	sort=c g
}

\newcommand{\systeminput}{\glslink{systeminput}{u}}
\newglossaryentry{systeminput}{
	name={\ensuremath{\systeminput}},
	description={Input},
	sort=c h
}

\newcommand{\X}{\glslink{X}{\mathcal{X}}}
\newglossaryentry{X}{
	name={\ensuremath{\X}},
	description={Admissible set of states},
	sort=c i
}

\newcommand{\U}{\glslink{U}{\mathcal{U}}}
\newglossaryentry{U}{
	name={\ensuremath{\U}},
	description={Admissible set of inputs},
	sort=c j
}

\newcommandx{\statetrajectory}[1][1=i, usedefault]{\glslink{statetrajectory}{X_{#1}}}
\newglossaryentry{statetrajectory}{
	name={\ensuremath{\statetrajectory[ ]}},
	description={State trajectory},
	sort=c k
}

\newcommandx{\inputtrajectory}[1][1=i, usedefault]{\glslink{inputtrajectory}{U_{#1}}}
\newglossaryentry{inputtrajectory}{
	name={\ensuremath{\inputtrajectory[ ]}},
	description={Input trajectory},
	sort=c l
}

\newcommand{\Alon}{\glslink{Alon}{\systemmatrix_{\lon}}}
\newglossaryentry{Alon}{
	name={\ensuremath{\Alon}},
	description={System matrix},
	type=longitudinalplanning,
	sort=2 a
}
\newcommandx{\Alonk}[1][1=\timestep, usedefault]{\glslink{Alonk}{\systemmatrix_{\lon, #1}}}
\newglossaryentry{Alonk}{
	name={\ensuremath{\Alonk}},
	description={System matrix at time step $\timestep$},
	parent=Alon,
	type=longitudinalplanning,
	sort=2 a
}

\newcommandx{\Blon}{\glslink{Blon}{\inputmatrix_{\lon}}}
\newglossaryentry{Blon}{
	name={\ensuremath{\Blon}},
	description={Input matrix trajectory planning},
	type=longitudinalplanning,
	sort=2 b
}
\newcommandx{\Blonk}[1][1=\timestep, usedefault]{\glslink{Blonk}{\inputmatrix_{\lon, #1}}}
\newglossaryentry{Blonk}{
	name={\ensuremath{\Blonk}},
	description={Input matrix at time step $\timestep$},
	parent=Blon,
	type=longitudinalplanning,
	sort=2 b
}

\newcommand{\xlon}{\glslink{xlon}{\x_\lon}}
\newglossaryentry{xlon}{
	name={\ensuremath{\xlon}},
	description={State},
	type=longitudinalplanning,
	sort=2 c
}

\newcommandx{\xlonk}[1][1=\timestep, usedefault]{\glslink{xlonk}{\x_{{\lon, #1}}}}
\newglossaryentry{xlonk}{
	name={\ensuremath{\xlonk}},
	description={State at time step $\timestep$},
	parent=xlon,
	type=longitudinalplanning,
	sort=2 c
}

\newcommand{\ulon}{\glslink{ulon}{\systeminput_\lon}}
\newglossaryentry{ulon}{
	name={\ensuremath{\ulon}},
	description={Input},
	type=longitudinalplanning,
	sort=2 e
}

\newcommandx{\ulonk}[1][1=\timestep, usedefault]{\glslink{ulonk}{\systeminput_{\lon,#1}}}
\newglossaryentry{ulonk}{
	name={\ensuremath{\ulonk}},
	description={Input at time step $\timestep$},
	parent=ulon,
	type=longitudinalplanning,
	sort=2 e
}

\newcommandx{\Xlonk}[1][1=\timestep, usedefault]{\glslink{Xlonk}{\X_{\lon, #1}}}
\newglossaryentry{Xlonk}{
	name={\ensuremath{\Xlonk}},
	description={Set of admissible states at time step $\timestep$},
	type=longitudinalplanning,
	sort=2 f
}

\newcommandx{\Ulonk}[1][1=\timestep, usedefault]{\glslink{Ulonk}{\U_{\lon, #1}}}
\newglossaryentry{Ulonk}{
	name={\ensuremath{\Ulonk}},
	description={Set of admissible inputs at time step $\timestep$},
	type=longitudinalplanning,
	sort=2 g
}

\newcommandx{\statetrajectorylon}{\glslink{statetrajectorylon}{\statetrajectory[\lon]}}
\newglossaryentry{statetrajectorylon}{
	name={\ensuremath{\statetrajectorylon}},
	description={Longitudinal state trajectory},
	type=longitudinalplanning,
	sort=2 h
}

\newcommand{\costlon}{\glslink{costlon}{J_\lon}}
\newglossaryentry{costlon}{
	name={\ensuremath{\costlon}},
	description={Cost function},
	type=longitudinalplanning,
	sort=2 i
}

\newcommand{\Alat}{\glslink{Alat}{\systemmatrix_{\lat}}}
\newglossaryentry{Alat}{
	name={\ensuremath{\Alat}},
	description={System matrix},
	sort=3 a,
	type=lateralplanning,
}
\newcommandx{\Alatk}[1][1=\timestep, usedefault]{\glslink{Alatk}{\systemmatrix_{\lat, #1}}}
\newglossaryentry{Alatk}{
	name={\ensuremath{\Alatk}},
	description={System matrix at time step $\timestep$},
	sort=3 b,
	parent=Alat,
	type=lateralplanning,
}

\newcommandx{\Blat}{\glslink{Blat}{\inputmatrix_{\lat}}}
\newglossaryentry{Blat}{
	name={\ensuremath{\Blat}},
	description={Input matrix},
	sort=3 c,
	type=lateralplanning,
}
\newcommandx{\Blatk}[1][1=\timestep, usedefault]{\glslink{Blatk}{\inputmatrix_{\lat, #1}}}
\newglossaryentry{Blatk}{
	name={\ensuremath{\Blatk}},
	description={Input matrix at time step $\timestep$},
	sort=3 d,
	parent=Blat,
	type=lateralplanning,
}

\newcommand{\xlat}{\glslink{xlat}{\x_\lat}}
\newglossaryentry{xlat}{
	name={\ensuremath{\xlat}},
	description={State},
	sort=3 f,
	type=lateralplanning,
}

\newcommandx{\xlatk}[1][1=\timestep, usedefault]{\glslink{xlatk}{\x_{{\lat, #1}}}}
\newglossaryentry{xlatk}{
	name={\ensuremath{\xlatk}},
	description={State at time step $\timestep$},
	sort=3 g,
	parent=xlat,
	type=lateralplanning,
}

\newcommand{\ulat}{\glslink{ulat}{\systeminput_\lat}}
\newglossaryentry{ulat}{
	name={\ensuremath{\ulat}},
	description={Input},
	sort=3 h,
	type=lateralplanning,
}

\newcommandx{\ulatk}[1][1=\timestep, usedefault]{\glslink{ulatk}{\systeminput_{\lat,#1}}}
\newglossaryentry{ulatk}{
	name={\ensuremath{\ulatk}},
	description={Input at time step $\timestep$},
	sort=3 i,
	type=lateralplanning,
	parent=ulat
}

\newcommandx{\Xlatk}[1][1=\timestep, usedefault]{\glslink{Xlatk}{\X_{\lat, #1}}}
\newglossaryentry{Xlatk}{
	name={\ensuremath{\Xlatk}},
	description={Set of admissible states at time step $\timestep$},
	type=lateralplanning,
	sort=3 j
}

\newcommandx{\Ulatk}[1][1=\timestep, usedefault]{\glslink{Ulatk}{\U_{\lat, #1}}}
\newglossaryentry{Ulatk}{
	name={\ensuremath{\Ulatk}},
	description={Set of admissible inputs at time step $\timestep$},
	type=lateralplanning,
	sort=3 k
}

\newcommandx{\statetrajectorylat}{\glslink{statetrajectorylat}{\statetrajectory[\lat]}}
\newglossaryentry{statetrajectorylat}{
	name={\ensuremath{\statetrajectorylat}},
	description={Lateral state trajectory},
	type=lateralplanning,
	sort=3 l
}

\newcommand{\costlat}{\glslink{costlat}{J_\lat}}
\newglossaryentry{costlat}{
	name={\ensuremath{\costlat}},
	description={Cost function},
	type=lateralplanning,
	sort=3 m
}

\newcommandx{\obstaclesk}[1][1=\globalvar, usedefault]{\glslink{obstaclesk}{#1{\mathcal{O}}_{\timestep}}}
\newglossaryentry{obstaclesk}{
	name={\ensuremath{\obstaclesk}},
	description={Occupancy sets of all obstacles in $\cartesianframe$ at time step $\timestep$},
	type=obs,
	sort=9a
}

\newcommandx{\obstaclescirclek}[1][1=\globalvar, usedefault]{\glslink{obstaclescirclek}{#1{\mathcal{O}}^{\basiccircleshape}_{\timestep}}}
\newglossaryentry{obstaclescirclek}{
	name={\ensuremath{\obstaclescirclek}},
	description={Occupancy sets $\obstaclesk$ dilated with circle $\basiccircleshape$},
	type=obs,
	sort=9c
}

\newcommand{\Areach}{\glslink{Areach}{\systemmatrix_{\R}}}
\newglossaryentry{Areach}{
	name={\ensuremath{\Areach}},
	description={System matrix},
	sort=10a,
	type=reach,
}

\newcommand{\Breach}{\glslink{Breach}{\inputmatrix_{\R}}}
\newglossaryentry{Breach}{
	name={\ensuremath{\Breach}},
	description={Input matrix},
	sort=10b,
	type=reach,
}

\newcommand{\xreach}{\glslink{xreach}{\x_{\R}}}
\newglossaryentry{xreach}{
	name={\ensuremath{\xreach}},
	description={State},
	sort=10c,
	type=reach,
}

\newcommandx{\xreachk}[1][1=\timestep, usedefault]{\glslink{xreachk}{\x_{\R, #1}}}
\newglossaryentry{xreachk}{
	name={\ensuremath{\xreachk}},
	description={State $\xreach$ at time step $\timestep$},
	sort=10c,
	type=reach,
	parent=xreach
}

\newcommand{\ureach}{\glslink{ureach}{\glslink{ureach}{\systeminput_{\R}}}}
\newglossaryentry{ureach}{
	name={\ensuremath{\ureach}},
	description={Input},
	sort=10d,
	type=reach,
}

\newcommandx{\ureachk}[1][1=\timestep, usedefault]{\glslink{ureachk}{\systeminput_{\R, #1}}}
\newglossaryentry{ureachk}{
	name={\ensuremath{\ureachk}},
	description={Input $\ureach$ at time step $\timestep$},
	sort=10d,
	type=reach,
	parent=ureach
}

\newcommandx{\Xreachk}[1][1=\timestep, usedefault]{\glslink{Xreachk}{\X_{\R, #1}}}
\newglossaryentry{Xreachk}{
	name={\ensuremath{\Xreachk}},
	description={Set of admissible states at time step $\timestep$},
	type=reach,
	sort=10e
}

\newcommandx{\Ureachk}[1][1=\timestep, usedefault]{\glslink{Ureachk}{\U_{\R, #1}}}
\newglossaryentry{Ureachk}{
	name={\ensuremath{\Ureachk}},
	description={Set of admissible inputs at time step $\timestep$},
	type=reach,
	sort=10f
}

\newcommandx{\forbiddenset}[1][1=\timestep, usedefault]{\glslink{forbiddenset}{\mathcal{F}_{#1}}}
\newglossaryentry{forbiddenset}{
	name={\ensuremath{\forbiddenset}},
	description={Set of forbidden states at time step $\timestep$},
	type=reach,
	sort=10g
}

\newcommandx{\egooccupancy}[2][1={(\xreachk)}, 2=\globalvar, usedefault]{\glslink{egooccupancy}{#2{\mathcal{Q}}#1}}
\newglossaryentry{egooccupancy}{
	name={\ensuremath{\egooccupancy}},
	description={Occupancy of ego vehicle in $\cartesianframe$ at state $\xreachk$},
	type=reach,
	sort=10h
}

\newcommand{\exact}{\mathtt{e}}
\newcommand{\exactval}[1]{#1^{\exact}}

\newcommandx{\exactreach}[1][1=\timestep, usedefault]{\glslink{exactreach}{\exactval{\R}_{#1}}}
\newglossaryentry{exactreach}{
	name={\ensuremath{\exactreach}},
	description={Exact reachable set at time step $\timestep$},
	type=reach,
	sort=10i
}

\newcommandx{\reachk}[1][1=\timestep, usedefault]{\glslink{reachk}{\R_{#1}}}
\newglossaryentry{reachk}{
	name={\ensuremath{\reachk}},
	description={Approx. reachable set at time step $\timestep$},
	type=reach,
	sort=10j
}

\newcommand{\D}{\mathcal{D}}
\newcommandx{\exactdrivablearea}[1][1=\timestep, usedefault]{\glslink{exactdrivablearea}{\exactval{\D}_{#1}}}
\newglossaryentry{exactdrivablearea}{
	name={\ensuremath{\exactdrivablearea}},
	description={Exact drivable area at time step $\timestep$},
	type=reach,
	sort=10k
}

\newcommandx{\drivableareak}[1][1=\timestep, usedefault]{\glslink{drivableareak}{\D_{#1}}}
\newglossaryentry{drivableareak}{
	name={\ensuremath{\drivableareak}},
	description={Approx. drivable area at time step $\timestep$},
	type=reach,
	sort=10l
}

\newcommand{\graph}{\mathcal{G}}

\newcommand{\graphreachability}{\glslink{graphreachability}\graph_{\R}}
\newglossaryentry{graphreachability}{
	name={\ensuremath{\graphreachability}},
	description={Reachability graph},
	type=reach,
	sort=10m
}

\newcommand{\idxreach}{i}
\newcommand{\baseset}{\R}

\newcommandx{\sBik}[2][1=(\idxreach), 2=\timestep, usedefault]{\glslink{sBik}{\baseset^{#1}_{#2}}}
\newglossaryentry{sBik}{
	name={\ensuremath{\sBik}},
	description={$\idxreach$-th base set at time step $\timestep$},
	type=reach,
	sort=10n
}

\newcommand{\polytope}{\mathcal{P}}
\newcommandx{\sPikl}[3][1=(\idxreach), 2=\timestep, usedefault]{\glslink{sPikllondir}{\polytope^{#1}_{#3, #2}} \glslink{sPikllatdir}{}}
\newglossaryentry{sPikllondir}{
	name={\ensuremath{\sPikl{\curvframelondir}}},
	description={$\idxreach$-th polytope in the $(\slon, \vlon)$ plane at time step $\timestep$},
	type=reach,
	sort=10o
}
\newglossaryentry{sPikllatdir}{
	name={\ensuremath{\sPikl{\curvframelatdir}}},
	description={$\idxreach$-th polytope in the $(\slat, \vlat)$ plane at time step $\timestep$},
	type=reach,
	sort=10p
}

\newcommand{\aabb}{\D}
\newcommandx{\sAik}[2][1=(\idxreach), 2=\timestep, usedefault]{\glslink{sAik}{\aabb^{#1}_{#2}}}
\newglossaryentry{sAik}{
	name={\ensuremath{\sAik}},
	description={Projection of $\sBik$ onto position domain},
	type=reach,
	sort=10q
}

\newcommand{\prop}{\mathtt{prop}}
\newcommandx{\propreachk}[1][1=\timestep, usedefault]{\glslink{propreachk}{\R^{\prop}_{#1}}}
\newglossaryentry{propreachk}{
	name={\ensuremath{\propreachk}},
	description={Reachable set after propagation step},
	type=reach,
	parent=reachk
}

\newcommandx{\propdrivableareak}[1][1=\timestep, usedefault]{\glslink{propdrivableareak}{\D^{\prop}_{#1}}}
\newglossaryentry{propdrivableareak}{
	name={\ensuremath{\propdrivableareak}},
	description={Drivable area after propagation step},
	type=reach,
	parent=drivableareak
}

\newcommandx{\sBPik}[2][1=(\idxreach), 2=\timestep, usedefault]{\glslink{sBPik}{\baseset^{\prop#1}_{#2}}}
\newglossaryentry{sBPik}{
	name={\ensuremath{\sBPik}},
	description={Propagated base set at time step $\timestep$},
	type=reach,
	parent=sBik
}

\newcommandx{\sAPik}[2][1=(\idxreach), 2=\timestep, usedefault]{\glslink{sAPik}{\aabb^{\prop#1}_{#2}}}
\newglossaryentry{sAPik}{
	name={\ensuremath{\sAPik}},
	description={Projection of $\sBPik$ onto position domain},
	type=reach,
	parent=sAik
}

\newcommandx{\sPPikl}[3][1=(\idxreach), 2=\timestep, usedefault]{\glslink{sPPikllondir}{\polytope^{\prop#1}_{#3, #2}} \glslink{sPPikllatdir}{}}
\newglossaryentry{sPPikllondir}{
	name={\ensuremath{\sPPikl{\curvframelondir}}},
	description={$\idxreach$-th propagated polytope},
	type=reach,
	parent=sPikllondir
}
\newglossaryentry{sPPikllatdir}{
	name={\ensuremath{\sPPikl{\curvframelatdir}}},
	description={$\idxreach$-th propagated polytope},
	type=reach,
	parent=sPikllatdir
}

\newcommand{\repartitioned}{\mathtt{rprt}}
\newcommandx{\repartitioneddrivableareak}[1][1=\timestep, usedefault]{\glslink{repartitioneddrivableareak}{\D^{\repartitioned}_{#1}}}
\newglossaryentry{repartitioneddrivableareak}{
	name={\ensuremath{\repartitioneddrivableareak}},
	description={Drivable area after re-partitioning step},
	type=reach,
	parent=drivableareak,
	sort=10l
}

\newcommandx{\sARik}[2][1=(\idxreach), 2=\timestep, usedefault]{\glslink{sARik}{\aabb^{\repartitioned#1}_{#2}}}
\newglossaryentry{sARik}{
	name={\ensuremath{\sARik[(q)]}},
	description={Part of drivable area after re-partitioning step},
	type=reach,
	parent=sAik,
}

\newcommandx{\sPPhatikl}[3][1=(\idxreach), 2=\timestep, usedefault]{\hat{\polytope}^{#1}_{#3, #2}}

\newcommand{\enlargement}{\glslink{enlargement}{\mathcal{A}^\epsilon}}
\newglossaryentry{enlargement}{
	name={\ensuremath{\enlargement}},
	description={Enlargement to consider shape of ego vehicle},
	type=reach,
	sort=10r
}

\newcommand{\idxconnectedset}{n}
\newcommand{\connected}{\mathcal{C}}

\newcommand{\connectedset}{\glslink{connectedset}{\mathcal{C}}}
\newglossaryentry{connectedset}{
	name={\ensuremath{\connectedset}},
	description={Connected set in the position domain},
	type=drivingcorridor,
	sort=11a
}

\newcommandx{\connectedsetk}[1][1=\timestep, usedefault]{\glslink{connectedsetk}{\connectedset_{#1}}}
\newglossaryentry{connectedsetk}{
	name={\ensuremath{\connectedsetk}},
	description={Connected set at time step $\timestep$},
	type=drivingcorridor,
	parent=connectedset
}

\newcommandx{\connectedsetkn}[2][1=\timestep, 2=\idxconnectedset, usedefault]{\glslink{connectedsetkn}{\connectedset^{(#2)}_{#1}}}
\newglossaryentry{connectedsetkn}{
	name={\ensuremath{\connectedsetkn[][i]}},
	description={$i$-th connected set at time step $\timestep$},
	type=drivingcorridor,
	parent=connectedset
}

\newcommandx{\graphconnectedcomponents}[1][1= , usedefault]{\glslink{graphconnectedcomponents}{\graph_{\connected#1}}}
\newglossaryentry{graphconnectedcomponents}{
	name={\ensuremath{\graphconnectedcomponents}},
	description={Graph storing connected sets},
	type=drivingcorridor,
	sort=11c
}

\newcommand{\node}{\glslink{node}{\mathtt{n}}}
\newglossaryentry{node}{
	name={\ensuremath{\node}},
	description={Node in $\graphconnectedcomponents$},
	type=drivingcorridor,
	parent=graphconnectedcomponents
}

\newcommand{\dc}{C}

\newcommand{\dclon}{\glslink{dclon}{\dc_{\lon}}}
\newglossaryentry{dclon}{
	name={\ensuremath{\dclon}},
	description={Longitudinal driving corridor},
	type=drivingcorridor,
	sort=11e
}

\newcommandx{\dclonk}[1][1=\timestep, usedefault]{\glslink{dclonk}\dc_{\lon, #1}}
\newglossaryentry{dclonk}{
	name={\ensuremath{\dclonk}},
	description={Longitudinal driving corridor at time step $\timestep$},
	type=drivingcorridor,
	parent=dclon
}

\newcommand{\dclat}{\glslink{dclat}{\dc_{\lat}}}
\newglossaryentry{dclat}{
	name={\ensuremath{\dclat}},
	description={Lateral driving corridor},
	type=drivingcorridor,
	sort=11f
}

\newcommandx{\dclatk}[1][1=\timestep, usedefault]{\glslink{dclatk}\dc_{\lat, #1}}
\newglossaryentry{dclatk}{
	name={\ensuremath{\dclatk}},
	description={Lateral driving corridor at time step $\timestep$},
	type=drivingcorridor,
	parent=dclat
}

\newcommandx{\parentset}[1][1=k-1, usedefault]{\mathcal{D}^{\mathrm{parents}}_{#1}}

\newcommand{\distance}{d}
\newcommandx{\latdiscircle}[2][1=\idxcentercircle, 2=\timestep, usedefault]{\glslink{latdiscircle}{\distance^{(#1)}_{#2}}}
\newglossaryentry{latdiscircle}{
	name={\ensuremath{\latdiscircle}},
	description={Lateral distance of $\centercircle$ from $\refpath$ at time step $\timestep$},
	type=constraints,
	sort=12a
}

\newcommandx{\dmink}[2][1=\idxcentercircle, 2=\timestep, usedefault]{\glslink{dmink}{\minval{\distance}^{(#1)}_{#2}}}
\newglossaryentry{dmink}{
	name={\ensuremath{\dmink}},
	description={Minimum admissible value of $\latdiscircle$},
	type=constraints,
	parent=latdiscircle
}

\newcommandx{\dmaxk}[2][1=\idxcentercircle, 2=\timestep, usedefault]{\glslink{dmaxk}{\maxval{\distance}^{(#1)}_{#2}}}
\newglossaryentry{dmaxk}{
	name={\ensuremath{\dmaxk}},
	description={Maximum admissible value of $\latdiscircle$},
	type=constraints,
	parent=latdiscircle
}

\newcommandx{\straightlineik}[3][1=\idxcentercircle, 2=\timestep, usedefault]{\glslink{straightlineik}{g^{(#1)}_{#2}#3}}
\newglossaryentry{straightlineik}{
	name={\ensuremath{\straightlineik{}}},
	description={Straight line perpendicular to $\refpath$ going through $\centercircle$},
	type=constraints,
	sort=12b
}

\newcommandx{\intersectingposik}[2][1=\idxcentercircle, 2=\timestep, usedefault]{\glslink{intersectingposik}{\mathcal{Y}^{(#1)}_{#2}}}
\newglossaryentry{intersectingposik}{
	name={\ensuremath{\intersectingposik}},
	description={Positions in $\dclon$ intersecting with $\straightlineik{}$},
	type=constraints,
	sort=12c
}

\newcommand{\idxinterval}{q}
\newcommandx{\intervaliqk}[3][1=\idxcentercircle, 2=\timestep, 3=\idxinterval, usedefault]{\glslink{intervaliqk}{\mathcal{I}^{(#1)}_{#2,#3}}}
\newglossaryentry{intervaliqk}{
	name={\ensuremath{\intervaliqk}},
	description={$\idxinterval$-th interval of positions in $\dclon$ intersecting $\straightlineik{}$},
	type=constraints,
	sort=12d
}

\newcommandx{\validintervals}[1][1=\idxcentercircle,usedefault]{\mathcal{I}^{(#1)}_{\mathtt{v}}}

\newcommand{\placeholdersys}{\glslink{placeholdersys}{\mathrm{sys}}}
\newglossaryentry{placeholdersys}{
	name={\ensuremath{\placeholdersys}},
	description={Placeholder for a variable},
	sort=a
}

\newcommandx{\proj}[2][1=\lozenge, usedefault]{\glslink{proj}{\mathrm{proj}_{#1}\!(#2 )}}
\newglossaryentry{proj}{
	name={\ensuremath{\proj[\lozenge]{\x}}},
	description={Projection operator that maps the state $\x$ onto its elements $\lozenge$},
	sort=z
}

\newcommandx{\overlap}[1][1=\sAik, usedefault]{\glslink{overlap}{\mathrm{overlap}(#1 )}}
\newglossaryentry{overlap}{
	name={\ensuremath{\overlap}},
	description={Returns all indices $\idxreach$ of the propagated sets $\sBPik$ that overlap with $\sAik$},
	sort=z
}

\newcommand{\convexhull}{\glslink{convexhull}{\mathrm{convexhull}}}
\newglossaryentry{convexhull}{
	name={\ensuremath{\convexhull}},
	description={Computes the convex hull},
	sort=z
}

\newcommandx{\sminlonARik}[2][1=q, 2=\timestep, usedefault]{\minval{\pos}^{\repartitioned(#1)}_{\curvframelondir, #2}}
\newcommandx{\smaxlonARik}[2][1=q, 2=\timestep, usedefault]{\maxval{\pos}^{\repartitioned(#1)}_{\curvframelondir, #2}}
\newcommandx{\sminlatARik}[2][1=q, 2=\timestep, usedefault]{\minval{\pos}^{\repartitioned(#1)}_{\curvframelatdir, #2}}
\newcommandx{\smaxlatARik}[2][1=q, 2=\timestep, usedefault]{\maxval{\pos}^{\repartitioned(#1)}_{\curvframelatdir, #2}}

\newcommand{\approxsymb}{\mathtt{C}}

\newcommandx{\refcurvaturecirclei}[1][1=i, usedefault]{\curvature_{\refpathsym,#1}^{\approxsymb}}

\usepackage{tikz} 
\newcommand\copyrighttext{%
	\footnotesize \copyright 2025 IEEE. Personal use of this material is permitted. Permission from IEEE must be obtained for all other uses, in any current or future media, including reprinting/republishing this material for advertising or promotional purposes, creating new collective works, for resale or redistribution to servers or lists, or reuse of any copyrighted component of this work in other works.}
\newcommand\copyrightnotice{%
	\begin{tikzpicture}[remember picture,overlay]
		\node[anchor=south,yshift=5pt] at (current page.south) {\fbox{\parbox{\dimexpr\textwidth-\fboxsep-\fboxrule\relax}{\copyrighttext}}};
	\end{tikzpicture}%
}
\begin{document}

\title{Traffic-Rule-Compliant Trajectory Repair\\ via Satisfiability Modulo Theories \\and Reachability Analysis} 

\author{Yuanfei Lin, Zekun Xing, Xuyuan Han, and Matthias Althoff%
	\thanks{
	Manuscript received Month Date, Year; revised Month Date, Year.
	}
\thanks{The authors are with the Department of Computer Engineering, Technical University of Munich, 85748 Garching, Germany. } 
	\thanks{{\tt\footnotesize \{yuanfei.lin, zekun.xing, xuyuan.han, althoff\} @tum.de} \textit{ (Corresponding author: Yuanfei Lin.)}}
}

\markboth{Journal of XX,~Vol.~XX, No.~X, Month~Year}%
{Shell \MakeLowercase{\textit{et al.}}: Bare Demo of IEEEtran.cls for IEEE Journals}

\maketitle

\begin{abstract}
	Complying with traffic rules is challenging for automated vehicles, as numerous rules need to be considered simultaneously. If a planned trajectory violates traffic rules, it is common to replan a new trajectory from scratch. We instead propose a trajectory repair technique to save computation time. 
	By coupling satisfiability modulo theories with set-based reachability analysis, we determine if and in what manner the initial trajectory can be repaired. Experiments in high-fidelity simulators and in the real world demonstrate the benefits of our proposed approach in various scenarios. Even in complex environments with intricate rules, we efficiently and reliably repair rule-violating trajectories, enabling automated vehicles to swiftly resume legally safe operation in real time.
\end{abstract}


\IEEEpeerreviewmaketitle
\section{Introduction}
\label{sec:introduction}
\IEEEPARstart{J}{ust} like human drivers, automated vehicles must explicitly comply with traffic rules to ensure their safe operation on roads \cite{nomoretraffic}. 
If this is done carefully, automated driving can gain the necessary public trust, and responsible operators can mitigate potential liability claims in case of accidents. However, despite the paramount importance of rule compliance, existing motion planning algorithms, as highlighted in recent surveys \cite{Gonzalez2015, Paden2016, claussmann2019review, mehdipour2023formal}, either overlook traffic rules, incorporate them implicitly using deep learning approaches \cite{huang2020survey}, or address only a limited selection \cite{xiao2021rulebased, kochdumper2024real}.  \copyrightnotice

\begin{figure}[t!]
	\centering
	\vspace{2mm}
	\def\svgwidth{1\columnwidth}\footnotesize
	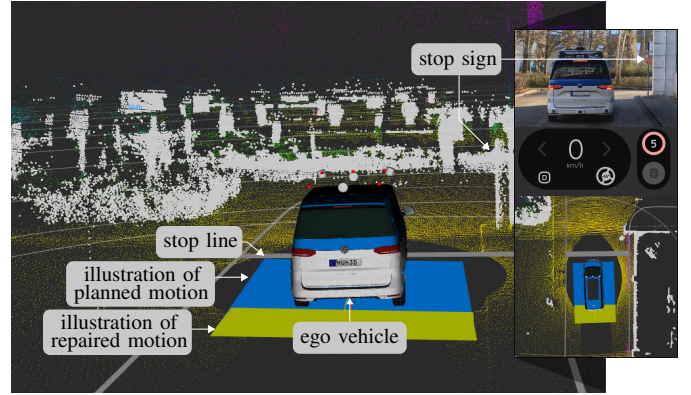
	\caption{
		Snapshot from our real-world driving experiments. Our trajectory repair approach dynamically adjusts planned trajectories that violate traffic rules in real time. For example, it enforces the stop-line rule by requiring the vehicle to halt before the marked line for the specified duration. The motion plans are illustrated by widening the path to be followed for improved visibility and are referenced to the rear axle. The project site, including code and experiment animations, is available at \href{https://commonroad.github.io/repair-to-drive/}{https://commonroad.github.io/repair-to-drive/}.
	}
	\label{fig:intro}
\end{figure}

Runtime verification (aka {\it monitoring}) is an efficient method to verify if the planned trajectory aligns with expected behaviors \cite{rizaldi2017formalising, du2020online, sahin2020autonomous}. If planned trajectories are monitored and identified as rule-violating, it is common to replan them within a receding-horizon planning framework. Nevertheless, replanning the entire trajectory is often undesired and inefficient, since:
\begin{enumerate}
	\item replanning typically requires substantially more computational resources than focusing on adjusting specific trajectory segments; and
	\item frequent replanning can lead to unpredictable and inconsistent behavior, potentially causing confusion among other traffic participants \cite[Sec. I]{werling2012optimal}, {whereas repair enables localized adjustments with greater continuity}.
\end{enumerate}
Therefore, we propose a trajectory repair framework that selectively refines a planned trajectory to address violations of traffic rules (see Fig.~\ref{fig:intro}). 

\subsection{Literature Overview}
\label{sec:literature}
We categorize existing works on rule-informed motion planning, trajectory replanning, and repair techniques for automated vehicles. 
{Since our work focuses on traffic rules derived from legal resources whose formalization remains fixed during online deployment, we do not review works addressing dynamic modification of specifications \cite{fainekos2011revising, alur2013counter, pacheck2023physically}}.
\subsubsection{Rule-Informed Motion Planning}\label{sec:ri_plan}
Traffic rules are initially described in natural language, which usually introduces ambiguity and unintended interpretations. Formal methods address this issue by using rigorous mathematical formalization
 \cite{mehdipour2023formal}. In particular, temporal logic is extensively used to formalize safety requirements and complex traffic rules. Examples include linear temporal logic (LTL)~\cite{Rizaldi2018,Esterle2020}, metric temporal logic (MTL) \cite{Maierhofer2020a, krasowski2021temporal, maierhofer2022formalization}, signal temporal logic (STL) \cite{Arechiga2019, zhong2023guided, meng2024diverse}, and their variants \cite{vasile2017minimum, dokhanchi2018evaluating}.

Several studies integrate LTL specifications into the motion planning problem by formulating them as automata and use rapidly-exploring random trees (RRTs) \cite{lavalle2001rapidly} to plan trajectories that satisfy the specifications \cite{castro2013incremental, barbosa2019integrated, vasile2020reactive}. The automata-based approaches are computationally expensive, and the RRT algorithm is only probabilistically complete \cite{karaman2010incremental}. 
In contrast, MTL and STL have the advantage of quantitatively evaluating the degree of satisfaction or violation using {robustness} (aka \textit{robustness degree}) \cite{FAINEKOS20094262, donze2010robust}. Thus, we can maximize the level of satisfaction by encoding the MTL and STL formulas as mixed-integer constraints and including the robustness in the cost function of optimization-based planners \cite{belta2019formal}.  But this approach scales poorly with an increasing number of integer variables.
Recent efforts leverage {smooth approximations,} neural-network-based methods, or control barrier functions to generate trajectories satisfying STL tasks with high computational efficiency~{\cite{Pant2017, lindemann2018control,liu2021recurrent, leung2023backpropagation}}.
However, they 
 cannot easily consider complex specifications due to their assumptions, such as the existence of valid control barrier functions \cite{lindemann2018control,liu2021recurrent}.
In addition, the authors of \cite{irani2021computing, liu2023specification, lercher2024specification} propose using reachability analysis and model checking to prune the search space of motion planners by enforcing compliance with traffic rules.  
{Furthermore, STL formulas are employed in \cite{zhong2023guided, meng2024diverse} to guide trajectory generation for controllable traffic simulations.} 
\subsubsection{Trajectory Replanning}
In classical motion planning, incremental techniques are widely employed to replan trajectories by dynamically adapting the motion planner to changing environments \cite{Paden2016}. 
Search algorithms, such as lifelong planning A* \cite{koenig2004lifelong}, D*  \cite{stentz1995optimal}, 
and anytime repairing A* \cite{likhachev2003ara}, are incremental variants of the original A* algorithm. 
Similarly, many sampling-based motion planners \cite{ferguson2006replanning, karaman2011anytime, otte2015rrt, connell2017dynamic, chandler2017online} use incremental strategies to progressively explore the solution space.
In task planning with LTL specifications, several studies \cite{guo2013revising, guo2015multi, lahijanian2016iterative} promote an incremental method for mending the product automaton, which combines an abstract model of the system and the specification automaton. 
To ensure the safety of black-box motion planners, the authors of \cite{shao2021reachability} use reachability analysis 
to verify safety and replan unsafe trajectories by sampling from the feasible configuration space. Likewise, the approach proposed in \cite{lin2022road} replans actions suggested by a reinforcement learning agent when they violate traffic rules. Furthermore, the work presented in \cite{drplanner} harnesses the power of large language models to enhance the performance of motion planners.
However, these methods are tightly integrated within the motion planner, which limits their flexibility.
\subsubsection{Trajectory Repair}
Trajectory repair differs from replanning by relying solely on the output of the motion planner, either by making local adjustments or branching off from it.
\paragraph{Local Trajectory Repair}\label{subsec:repair_initial}
Local trajectory repair typically uses the planned trajectory as a reference to improve its quality when it fails to meet the given criteria. For example, the authors of \cite{kalakrishnan2011stomp} employ an optimization procedure that incorporates trajectory smoothness as an additional constraint.
A similar regime is outlined in \cite{lau2009kinodynamic,phillips2011sipp, zucker2013chomp,  usenko2017real, liu2017planning, ding2018trajectory, missura2022fast}, where search algorithms, optimizers, or parametric curves are used to repair the initial plan. In addition, the planned trajectory can be locally deformed to avoid collisions, either by modeling it as an elastic band 
 \cite{Roesmann2012TEB}
or through an affine transformation~\cite{Pham.2012}. 
However, when addressing the nonholonomic motion of a vehicle, 
solving the two-point boundary value problem -- finding a trajectory that satisfies given initial and final conditions -- poses significant challenges, especially in real-time applications.

\paragraph{Branching Off From Trajectories} 
When the initially planned result maintains a sufficiently high quality, it can be partially reused, and a new partial trajectory can be branched off to meet all requirements. The works in \cite{Ziegler2014, ding2021epsilon, chen2022interactive, wang2023interaction, li2023marc, tong2023robust, peters2023contingency} bear conceptual similarities to our approach in this regard.  They account for potential future scenario evolutions, resulting in trajectories branching off for each alternative future scenario.
Notably, the authors of \cite{li2023marc} highlight the benefits of delaying branching points as much as possible, which reduces the number of constraints in the trajectory optimization problem and leaves more reaction time for the vehicle to handle potential hazards. This also aligns with the preference of system designers for safety systems to intervene at the latest possible point in time \cite{pek2020fail}.
However, their selection of branching points is not suitable for rule-compliant trajectory repair, as these are either defined solely by divergences in future predictions \cite{li2023marc}, determined through a static choice  \cite{ding2021epsilon, chen2022interactive}, or lack well-defined frameworks  \cite{Ziegler2014, peters2023contingency}. To address these limitations, we proposed using criticality measures in our previous work \cite{YuanfeiLin2021, wang2024, yuanfei2022}, as they objectively assess behavioral safety and offer a systematic approach for determining the appropriate timing of interventions. To efficiently obtain a satisfactory solution, satisfiability modulo theories (SMT) \cite{Barrett2018} is used to determine 
whether and how the violated rules can be satisfied.
\subsection{Contributions and Outline}
\label{sec:contribution}
Building upon our previous work \cite{YuanfeiLin2021, yuanfei2022, wang2024}, we present a trajectory repair approach to promote compliance with formalized traffic rules of an automated vehicle, referred to as the \textit{ego vehicle} from now on. 
{By refining several submodules, our approach achieves the following combination of capabilities that, to the best of our knowledge, has not been demonstrated previously in the literature:}
\begin{enumerate}
	\item {trajectory repair} for traffic rules formalized in STL with arbitrary temporal operators; 
	\item model predictive robustness {is used for the first time as a heuristic} for NP-complete Boolean satisfiability (SAT) solving within an SMT solver. {This improves efficiency and enables the retrieval of reasonable robustness values without excessive manual tuning.}
	\item set-based reachability analysis {is applied for the first time to} {systematically} compute  spatio-temporal constraints {supporting the repair of trajectories};
	\item {reduction of the} computation time in more critical scenarios with smaller {repair} solution spaces;
	\item {consideration of} arbitrary traffic scenarios, covering diverse driving environments, such as interstates and intersections; and
	\item {evaluation in} both open-loop and closed-loop traffic simulations, as well as in real-world scenarios.
\end{enumerate}

This article is structured as follows: After introducing the preliminaries and problem statement in Sec.~\ref{sec:preliminaries}, we present an overview of our approach in Sec.~\ref{sec:overview}. 
The abstraction process for propositional rule formulas is described in Sec.~\ref{sec:rule_monitoring}, followed by SAT solving and trajectory repair in  Sec.~\ref{sec:sat_solver} and Sec.~\ref{sec:T_solver}, respectively. Finally,
we evaluate our concept in Sec.~\ref{sec:experiment} before drawing conclusions in Sec.~\ref{sec:conclusions}.

\section{Preliminaries and Problem Statement}
\label{sec:preliminaries}

\subsection{System Description and Notations}\label{subsec:system}
An index $k\in\mathbb{N}_0$ corresponds to a discrete time step $t_k=k\Delta t$, where $\Delta t\in\mathbb{R}_+$ is a fixed time increment. The motion of vehicles is modeled as a discrete-time system:
\begin{equation}\label{eq:motion_veh}
	\boldsymbol{x}_{k+1} = f(\boldsymbol{x}_k, \boldsymbol{u}_k),
\end{equation}
where $\boldsymbol{x}_k\in\mathbb{R}^{n_x}$ is the state vector and $\boldsymbol{u}_k\in\mathbb{R}^{n_u}$ is the input vector. The state and input are bounded by sets of admissible values: $\forall k\in[0, h]\colon \boldsymbol{x}_k\in \mathcal{X}_k$ and $\forall k\in[0, h-1]\colon \boldsymbol{u}_k\in\mathcal{U}_k$, where $h\in\mathbb{N}_+$ is the final time step. We denote the solution of (\ref{eq:motion_veh}) at time step $k$ as $\chi\big(k, \boldsymbol{x}_0, \boldsymbol{u}_{[0,k-1]}\big)$, given an initial state $\boldsymbol{x}_0$ and an input trajectory $\boldsymbol{u}_{[0,k-1]}$ for the time interval $[0, k-1]$. {Note that for $k=0$, we set $\chi\big(k, \boldsymbol{x}_0, \boldsymbol{u}_{[0,k-1]}\big) = \boldsymbol{x}_0$.}
For simplicity, we sometimes write $\chi$ instead of $\chi\big([0,h], \boldsymbol{x}_0, \boldsymbol{u}_{[0, h-1]}\big)$. 

\begin{figure}[t!]
	\centering
	\def\svgwidth{0.85\columnwidth}\footnotesize
	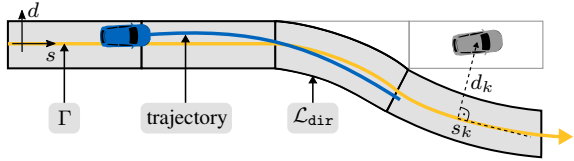
	\caption{Vehicle localization in a curvilinear coordinate system defined with respect to the reference path $\Gamma$, which represents the centerline of $\mathcal{L}_{\mathtt{dir}}$.
	}
	\vspace{-2mm}
	\label{fig:road_network}
\end{figure}

The road network $\mathcal{L}$ is composed of a set of lanelets~\cite{Bender2014}, each defined by polylines that establish its left and right boundaries. Given a planned or predicted trajectory, we compute a reference path $\Gamma$ \cite{Gerald2024}, which uniquely corresponds to a sequence of lanelets $\mathcal{L}_{\mathtt{dir}} \subset \mathcal{L}$ aligned with the driving direction.
As depicted in Fig.~\ref{fig:road_network}, a curvilinear coordinate system can be derived from $\Gamma$ for locating the vehicle using its longitudinal position $s$ and lateral deviation $d$ from $\Gamma$ at $s$ \cite{Gerald2024}. 
Let $\square$ represent a variable.
 The initial and repaired values of $\square$ are denoted by $\square^{\mathtt{ini}}$ and $\square^{\mathtt{rep}}$, respectively. Values associated with the ego vehicle are represented by $\square_{\mathtt{ego}}$, and those associated with an obstacle $\mathtt{obs}\in\mathcal{B}$ by $\square_{\mathtt{obs}}$, where $\mathcal{B}$ denotes the set of rule-relevant obstacles. Unless explicitly specified, all values refer to the ego vehicle. The concatenated states of vehicles at time step $k$ is denoted by  $\omega_k\coloneqq[\boldsymbol{x}^T_{\mathtt{ego}, k}, \boldsymbol{x}^T_{\mathtt{obs}_1, k},\dots,\boldsymbol{x}^T_{\mathtt{obs}_{|\mathcal{B}|}, k}]^T\in\mathcal{X}_{k}^{|\mathcal{B}|+1}$. 

\subsection{Signal Temporal Logic}
For traffic rule evaluation, we consider a discrete-time signal $\boldsymbol{\omega}\coloneqq\omega_0, \dots, \omega_k ,\dots, \omega_{h}$. 
Given formulas $\varphi$, $\varphi_1$, and $\varphi_2$, the syntax of STL 
is defined as \cite[Sec.~2.1]{bartocci2018specification}:
\begin{equation}\label{eq:stl_semantics}
	\varphi \coloneqq p \ |\ \lnot \varphi \ |\ \varphi_1\vee \varphi_2 \ |\ 
	 \varphi_1\mathbf{S}_{[a,b]} \varphi_2\  |\
	  \varphi_1\mathbf{U}_{[a,b]} \varphi_2,
\end{equation}
where $p \coloneqq \alpha(\omega_k)>0$ 
is an atomic predicate, defined via an {evaluation function} $\alpha\colon \mathcal{X}_k^{|\mathcal{B}| + 1} \rightarrow \mathbb{R}$. 
The symbols $\lnot$ and $\vee$ denote Boolean \textit{negation} and \textit{disjunction} operators. Additionally,  $\varphi_1\mathbf{S}_{[a,b]}\varphi_2$ and
 $\varphi_1\mathbf{U}_{[a,b]}\varphi_2$ represent the temporal \textit{since} and 
 \textit{until} operators, respectively, with a time bound of $[a, b]$, where $a, b\in\mathbb{R}_0$ and $b\geq a$. 
 The  logical \textit{True} and \textit{False} are denoted as $\top$ and $\bot$, respectively, and the valuation of a formula $\varphi$ is denoted as $\llbracket\varphi\rrbracket$, e.g., $\llbracket\varphi\rrbracket=\top$.
To simplify notation, we omit the interval from temporal operators when it extends to the end of the input signal, i.e., $[0, h]$. 
If a signal $\boldsymbol{\omega}$ complies with $\varphi$ at time step $k$, we write $\omega_k\models\varphi$. If not, we write $\omega_k\not\models\varphi$.
Additional temporal logic operators can be constructed from (\ref{eq:stl_semantics})~(see \cite[Sec.~2.1]{bartocci2018specification}), such as $\varphi_1\wedge \varphi_2\coloneqq\lnot(\lnot\varphi_1\vee \lnot\varphi_2)$ (\textit{conjunction}), $\varphi_1\Rightarrow \varphi_2\coloneqq\lnot\varphi_1\vee\varphi_2$ (\textit{implication}), 
$\mathbf{P}\varphi\coloneqq\bot\mathbf{S}\varphi$ (\textit{previous}),
 $\mathbf{O}_{[a,b]}\varphi\coloneqq\top\mathbf{S}_{[a,b]}\varphi$ (\textit{once}), $\mathbf{H}_{[a,b]}\varphi\coloneqq\lnot\mathbf{O}_{[a,b]}\lnot\varphi$ (\textit{historically}), $\mathbf{F}_{[a,b]}\varphi\coloneqq\top\mathbf{U}_{[a,b]}\varphi$ (\textit{eventually}), and $\mathbf{G}_{[a,b]}\varphi\coloneqq\lnot\mathbf{F}_{[a,b]}\lnot\varphi$ (\textit{globally}). {Furthermore, STL formulas can be seamlessly expressed in canonical forms such as negation normal form (NNF) \cite{sadraddini2015robust} -- where negations are applied only to predicates -- and conjunctive normal form (CNF), which is a conjunction of disjunctions of (negated) predicates.}
We now define the STL robustness.

\begin{definition}[STL Robustness {\cite[Def. 3]{donze2010robust}}]\label{def:rob}
The robustness  $\rho_{\varphi}(\boldsymbol{\omega}, k)$ of an STL formula  $\varphi$ (see (\ref{eq:stl_semantics})) with respect to a signal $\boldsymbol{\omega}$ at time step $k$ is defined as:
\vspace{-1mm}\begin{align*}
		\rho_{\lnot\varphi}(\boldsymbol{\omega}, k) &\coloneqq 	-\rho_{\varphi}(\boldsymbol{\omega}, k),\\
		\rho_{\varphi_1\vee\varphi_2}(\boldsymbol{\omega}, k) &\coloneqq \max\big(	\rho_{\varphi_1}(\boldsymbol{\omega}, k),  \rho_{\varphi_2}(\boldsymbol{\omega}, k) \big),\\
		\rho_{\varphi_1\mathbf{U}_{[a,b]}\varphi_2}(\boldsymbol{\omega}, k) &\coloneqq 	\max_{k'\in[k+a,k+b]}\big( \min\big(	\rho_{\varphi_2}(\boldsymbol{\omega}, k'),\\&\hspace{25mm} 	
		\min_{\tau\in[k,k']} \rho_{\varphi_1}(\boldsymbol{\omega}, \tau)\big)\big),\\\vspace{-1mm}
		\rho_{\varphi_1\mathbf{S}_{[a,b]}\varphi_2}(\boldsymbol{\omega}, k) &\coloneqq 	\max_{k'\in[k-a,k-b]}\big( \min\big(	\rho_{\varphi_2}(\boldsymbol{\omega}, k'),\\&\hspace{25mm} \vspace{-1mm}	
		\min_{\tau\in[k,k']} \rho_{\varphi_1}(\boldsymbol{\omega}, \tau)\big)\big).
\end{align*}
\end{definition}
{Note that the signal $\boldsymbol{\omega}$  is assumed sufficiently long for proper operator evaluation; otherwise, a finitary interpretation is required \cite[Sec.~2.1]{bartocci2018specification}.} 

\subsection{Definitions for Trajectory Repair}
\label{subsec:definitions}

Without loss of generality, we henceforth assume that all STL formulas are expressed in NNF, and omit the definition of negation for simplicity.
	{We denote the branching time step of the planned trajectory as $k_{\mathrm{cut}}\in[0, h]$, whose corresponding state is referred to as the \textit{cut-off state} \cite[Def.~1]{yuanfei2022}, from which the repaired trajectory begins.} To establish an upper bound for $k_{\mathrm{cut}}$, we define:
\begin{definition}[Time-To-Violation {\cite[Def. 4]{yuanfei2022}}]\label{def:tv}
	The time-to-violation $\mathtt{TV}_\varphi$ at time step $k$ for a trajectory $\chi$ 
	with respect to an STL formula $\varphi$ is defined as:
	\allowdisplaybreaks{
		\vspace{-1mm}\begin{align*}
			\mathtt{TV}_{p} (\boldsymbol{u}_{[0,h-1]}, k)&\coloneqq 
			\begin{cases}
				k  & \text{if}\  \chi\big(k, \boldsymbol{x}_0, \boldsymbol{u}_{[0,k-1]}\big) \not\models p,\\
				\infty &\text{otherwise},
			\end{cases}\nonumber \\
		\mathtt{TV}_{\varphi_1\vee\varphi_2} (\boldsymbol{u}_{[0,h-1]}, k) &\coloneqq \max\big(\mathtt{TV}_{\varphi_1} (\boldsymbol{u}_{[0,h-1]}, k),\\
		 & \qquad \qquad \mathtt{TV}_{\varphi_2}(\boldsymbol{u}_{[0,h-1]}, k)\big),\\
			\mathtt{TV}_{\varphi_1\mathbf{U}_{[a, b]}\varphi_2}(\boldsymbol{u}_{[0,h-1]}, k)  & \coloneqq 
			\max_{{k'\in[k+a,k+b]}}\big(\\ 
			 \min(
			\mathtt{TV}_{\varphi_2}(\boldsymbol{u}&{_{[0,h-1]}, k'),\!\min_{\tau\in[k, k']}  \!\mathtt{TV}_{\varphi_1}(\boldsymbol{u}_{[0,h-1]}, \tau)
			)\big).}
		\end{align*}}
\end{definition}
{As shown in Def.~\ref{def:rob}, the past-time operator $\mathbf{S}$ corresponds to its future-time counterpart $\mathbf{U}$ when reversing time; thus, its time-to-violation can be defined analogously.} We can then derive $\mathtt{TV}_\varphi$ for other operators, such as
\allowdisplaybreaks{
\begin{align*}
	\mathtt{TV}_{\mathbf{F}_{[a, b]}\varphi}(\boldsymbol{u}_{[0,h-1]}, k) &\coloneqq \max_{k+a \leq k'\leq k+b} \!\mathtt{TV}_{\varphi}(\boldsymbol{u}_{[0,h-1]}, k'),\\
	\mathtt{TV}_{\mathbf{G}_{[a, b]}\varphi}(\boldsymbol{u}_{[0,h-1]}, k) &\coloneqq \min_{k+a \leq k'\leq k+b} \!\mathtt{TV}_{\varphi}(\boldsymbol{u}_{[0,h-1]}, k').
\end{align*}}

\noindent {Similarly to STL robustness, for incomplete signals, weak or strong semantics are used depending on the operator \cite[Sec.~III]{bauer2010comparing}. A common convention in the literature is to apply weak semantics to $\mathbf{G}_{[a, b]}\varphi$, where $\mathtt{TV}_{\varphi} (\boldsymbol{u}_{[0,h-1]}, k')$ is set to $\infty$ for all $k'>h$ \cite[Sec.~2.1]{bartocci2018specification}.}
With this, we formally define:
\begin{definition}[Time-To-Comply {\cite[Def. 5]{yuanfei2022}}]\label{def:tc}
	The time-to-comply $\mathtt{TC}_\varphi$ is the latest time step before $\mathtt{TV}_\varphi$ at which a trajectory $\chi^{\mathtt{ini}}$ complying with $\varphi$ exists:
\begin{align*}
	 \mathtt{TC}_{\varphi}(\boldsymbol{u}^{\mathtt{ini}}_{[0,h-1]})\coloneqq \max \bigg(\!\Big\{\!-&\infty\!\Big\} \cup \Big\{ 
	  k \in [0, \mathtt{TV}_\varphi(\boldsymbol{u}^{\mathtt{ini}}_{[0,h-1]}, 0)] \, \Big| \, \\ \notag \exists \boldsymbol{u}_{[k,h-1]}\colon &\forall k' \in [k,h-1]\colon\boldsymbol{u}_{k'}\in\mathcal{U}_{k'}, \\ & \mathtt{TV}_\varphi ([\boldsymbol{u}^{\mathtt{ini}}_{[0,k-1]}, \boldsymbol{u}_{[k,h-1]} ],0 ) \!=\! \infty  \!\Big\} \!\bigg).
\end{align*}
\end{definition}
	\begin{figure}[t!]
	\centering
	\def\svgwidth{0.86\columnwidth}\footnotesize
	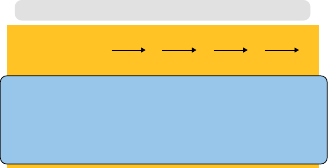
	\caption{Example computations of $\mathtt{TV}_{\varphi}$. A circle represents a state, with its marking indicating whether the formula is satisfied or violated. {For evaluating $\mathbf{G}_{[0, 5]}(\varphi_1\lor\varphi_2)$, we have $\mathtt{TV}_{\varphi_1\lor\varphi_2} (\boldsymbol{u}_{[0,h-1]}, k')=\infty$ for all $k'>4$.}} 
	\vspace{-2mm}
	\label{fig:tv_example}
\end{figure}
Hereafter, we omit the input dependency of $\mathtt{TV}_\varphi$ and $\mathtt{TC}_\varphi$ for brevity. If no violation of $\varphi$ occurs, then $\mathtt{TC}_\varphi = \mathtt{TV}_\varphi = \infty$.
Fig.~\ref{fig:tv_example} illustrates example signals that violate the given formulas.
Finally, we define reachable sets, which {help us find} rule-compliant trajectories starting from the cut-off state: 

\begin{definition}[Specification-Compliant Reachable Sets {\cite[Sec. II-C]{lercher2024specification}}] 
	The specification-compliant reachable set $\mathcal{R}^{\mathtt{e}}_{k}$ at time step $k$ is the set of states that can be reached from the set of initial states $\mathcal{X}_{0}$ {by following trajectories that comply with $\varphi$}:
\begin{align*}
	{\mathcal{R}^{\mathtt{e}}_{k}(\mathcal{X}_{0},\varphi)\coloneqq \bigg\{\chi\big(k, \boldsymbol{x}_0, \boldsymbol{u}_{[0,k-1]}\big) \ \bigg|\ \bigg(\exists \boldsymbol{x}_{0} \in \mathcal{X}_{0}, }&\\  
	\forall k'\!\in[0,k-1],
	\exists \boldsymbol{u}_{k'}\in\mathcal{U}_{k'}\! \colon \chi\big(k', \boldsymbol{x}_0, \boldsymbol{u}_{[0,k'-1]}\big) {\in \mathcal{X}_{k'}} \bigg)\\  
	{\bigwedge \chi\big([0,k], \boldsymbol{x}_0, \boldsymbol{u}_{[0,k-1]}	\big) \models \varphi \bigg\}.} &
\end{align*}
	\vspace{-8mm}
\end{definition}

\subsection{Problem Formulation}\label{subsuc:prob}

We focus on traffic rules formalized in STL,  where the only difference from MTL is that atomic propositions $\sigma$ in MTL have to be expressed as predicates $p$ in STL \cite{bartocci2018specification}. 
Once $\chi^{\mathtt{ini}}$ violates a rule $\varphi$, i.e., $\mathtt{TV}_\varphi\neq\infty$,  our repairer aims to first determine the cut-off time step\footnote{For multiple violated rules $\varphi_1,\varphi_2,\dots,\varphi_{n_{\mathrm{v}}}$, they are conjoined using $\wedge$ as $\varphi=\bigwedge_{\mathrm{v}=1}^{n_{\mathrm{v}}} \varphi_{\mathrm{v}}$. The corresponding $\mathtt{TV}_\varphi$ is defined as the minimum value among those assigned to the rules (cf. Def.~\ref{def:tv}), i.e., $\mathtt{TV}_\varphi=\min_{\mathrm{v}\leq n_{\mathrm{v}}} \mathtt{TV}_{\varphi_{\mathrm{v}}}$.}:

\allowdisplaybreaks{
  \begin{subequations}\label{eq:problem_def}
  	\begin{equation}
  				{k_{\mathrm{cut}} =}  \ {\mathtt{TC}_{\varphi}},\label{eq:kprime}
  	\end{equation}
  	  		\text{and then solves the following optimization problem:}
\begin{align}
\min_{{\boldsymbol{u}_{[k_{\mathrm{cut}},h-1]}}} \label{eq:cost} 	\sum_{\tau=k_{\mathrm{cut}}}^{h}& J(\boldsymbol{x}_\tau, \boldsymbol{u}_\tau)\\
	\hspace{-7.5mm} \text{subject to} \hspace{1cm}
	\forall \tau\in[k_{\mathrm{cut}} , h-1]\colon & (\ref{eq:motion_veh}),\ \boldsymbol{u}_\tau\in \mathcal{U}_\tau,\label{eq:system_dyn}\\
\boldsymbol{x}_{k_{\mathrm{cut}}} = \boldsymbol{x}^{\mathtt{ini}}_{k_{\mathrm{cut}}},\	\forall \tau\in[k_{\mathrm{cut}} +1, h]\colon & \boldsymbol{x}_\tau\in \mathcal{R}_{\tau}^{\mathtt{e}}(\mathcal{X}^{\mathtt{ini}}_{k_{\mathrm{cut}}},\varphi). \label{eq:after_tc}
\end{align}
  \end{subequations}}

\noindent 
{As a result, we obtain the repaired trajectory as $\chi^{\mathtt{rep}}=\chi\big([0,h], \boldsymbol{x}_0, [\boldsymbol{u}^{\mathtt{ini}}_{[0, k_{\mathrm{cut}} -1]}, \boldsymbol{u}_{[k_{\mathrm{cut}},h-1]}]\big)$.}
Informally, our first goal is to  {ensure rule compliance by modifying the initial trajectory as late as possible at time step
$k_{\mathrm{cut}}$, while respecting dynamic constraints, such as jerk limits (cf. (\ref{eq:kprime}) and Def.~\ref{def:tc}).} 
 Subsequently, we optimize the input trajectory $\boldsymbol{u}_{[k_{\mathrm{cut}}, h-1]}$ under the system constraints in (\ref{eq:system_dyn}), minimizing the cost function $J$ (cf. (\ref{eq:cost})) and ensuring rule compliance by maintaining the solution within the reachable sets in (\ref{eq:after_tc}). {The main challenge lies in ensuring that both the selection of $k_{\mathrm{cut}}$ over infinitely many control inputs and the computation of the remaining trajectory satisfy diverse traffic rules formalized in STL, while also being performed efficiently.}
\section{Overview of the Trajectory Repair Approach}
\label{sec:overview}
{Since traffic rules must hold at all times, all formalized rules in \cite{Maierhofer2020a, krasowski2021temporal, maierhofer2022formalization} begin with a globally temporal operator $\mathbf{G}$.}
Within the scope of the outermost $\mathbf{G}$ operator, these rules are typically constructed to comprehensively cover all relevant situations, often expressed through disjunctions or implications. Consequently, satisfying only parts of the formula is {often} sufficient to ensure that the entire formula is satisfied, {such as when subformulas are connected by disjunctions.}
To this end, we use a lazy SMT solver \cite{sebastiani2007lazy} to improve solving efficiency by reasoning about smaller subformulas. The solver combines a SAT solver -- responsible for proposing truth valuations $\phi$ for the Boolean abstraction $\varphi^\mathtt{P}$ of $\varphi$ -- and a $\mathcal{T}$-solver, which checks whether the valuations $\phi$ are consistent with the background theory $\mathcal{T}$; in our case, it
	corresponds to verifying whether a solution to $\varphi=\phi$ exists.

\begin{figure}[t!]
	\centering
	\def\svgwidth{0.98\columnwidth}\footnotesize
	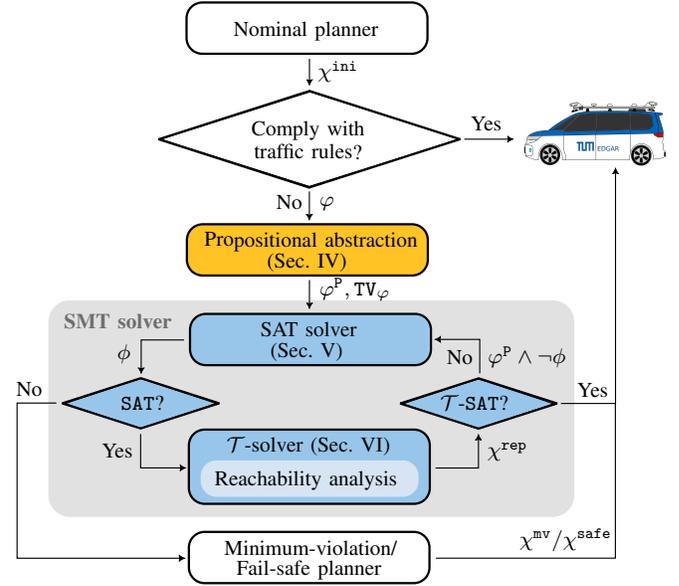
	\caption{Trajectory repair flowchart for a given nominal planner. 
	} 
	\vspace{-4mm}
	\label{fig:flowchart}
\end{figure}

Fig.~\ref{fig:flowchart} presents an overview of our trajectory repair framework running in every planning cycle. Our repair approach is implemented as an additional layer, together with a traffic rule monitor, between the nominal planner and the control layer. Our framework is designed to be agnostic to the choice of nominal planner and works as follows: 
Offline, we abstract the STL formulas of formalized traffic rules into propositional representations $\varphi^\mathtt{P}$ (see Sec.~\ref{sec:rule_monitoring}). 
In the online repair process, the framework leverages the abstraction $\varphi^\mathtt{P}$ of the violated rules  $\varphi$ to compute $\mathtt{TV}_\varphi$ (cf. Def.~\ref{def:tv}) and to initiate the SMT solver, starting with the SAT solver determining whether the abstracted problem is satisfiable, aka $\mathtt{SAT}$ (see Sec.~\ref{sec:sat_solver}). If the SAT solver finds a satisfying valuation $\phi$, the $\mathcal{T}$-solver determines the cut-off state and performs reachability analysis to solve (\ref{eq:problem_def}) (see Sec.~\ref{sec:T_solver}). 
If successful, it generates the repaired trajectory $\chi^{\mathtt{rep}}$, referred to as $\mathcal{T}$-$\mathtt{SAT}$. 
Otherwise, the framework refines the abstraction $\varphi^\mathtt{P}$ by adding $\phi$ as a conflicting clause $\lnot\phi$ and iterates the process. If the SAT solver ultimately proves the problem unsatisfiable, the framework concludes that no feasible repaired trajectory exists.  In such a scenario,  minimum-violation motion planning \cite{halder2022minimum, veer2023receding} complements our approach by offering a solution $\chi^{\mathtt{mv}}$ that minimizes the extent of required rule relaxation. 
Another alternative is a fail-safe trajectory $\chi^{\mathtt{safe}}$, which can be executed to ensure safety for an infinite time horizon \cite{pek2020fail}.

\section{Propositional Traffic Rule Abstraction}
\label{sec:rule_monitoring}
We abstract formalized traffic rules offline into a propositional logic formula that is sufficiently expressive for the SMT solver, while enabling effective online trajectory repair. To achieve this, we generalize the approach outlined in~\cite{yuanfei2022} by introducing three key steps:  formula rewriting (see Sec.~\ref{subsec:Flattening}), distributive decomposition (see Sec.~\ref{subsec:Deduction}), and CNF conversion (see Sec.~\ref{subsec:CNF}).
\subsection{Formula Rewriting}\label{subsec:Flattening} 
To obtain a more compact formula with reduced nesting, the STL formula $\varphi$ is rewritten into  $\varphi^\mathtt{F}$ in NNF \cite{sadraddini2015robust}. 	

	 \vspace{2mm}
	 \noindent \textbf{Running Example:} Consider the stop-line rule $\varphi_{\text{IN1}}$ from \cite{maierhofer2022formalization}: \textit{The ego vehicle must stop in front of the associated stop line for at least a duration of 
	 $t_\mathrm{slw}\in\mathbb{R}_+$ before entering the intersection}. This rule is formalized in STL as follows:
\allowdisplaybreaks{
	\begin{align}\label{eq:rule_rin1}
		\varphi_{\text{IN1}} = &\ \mathbf{G}\bigg( \left( \mathrm{passing\_stop\_line}(\boldsymbol{x}_{\mathtt{ego}}) \right.\nonumber \\
		&\qquad\qquad\wedge \mathrm{at\_traffic\_sign\_stop}(\boldsymbol{x}_{\mathtt{ego}}) \nonumber\\
		&\qquad\qquad\left. \wedge \lnot \mathrm{relevant\_traffic\_light}(\boldsymbol{x}_{\mathtt{ego}}) \right) \\
		&\Rightarrow \mathbf{O}\left( \mathbf{H}_{[0, t_\mathrm{slw}]} \left( \mathrm{stop\_line\_in\_front}(\boldsymbol{x}_{\mathtt{ego}}) \right. \right.\nonumber \\
		&\qquad\qquad\qquad\quad  \wedge \left. \left.\! \mathrm{in\_standstill}(\boldsymbol{x}_{\mathtt{ego}}) \right) \right) \bigg),\nonumber
\end{align}}
with
\begin{equation*}
	\begin{aligned}
		\mathrm{passing\_stop\_line}(\boldsymbol{x}_{\mathtt{ego}}) \coloneqq  &\ \mathbf{P}\left(\mathrm{stop\_line\_in\_front}(\boldsymbol{x}_{\mathtt{ego}})\right)\\
		&\wedge \lnot \mathrm{stop\_line\_in\_front}(\boldsymbol{x}_{\mathtt{ego}}).
	\end{aligned}
\end{equation*}
After rewriting, we obtain $\varphi_{\text{IN1}}^\mathtt{F}$ in NNF as:
\allowdisplaybreaks{
	\begin{align}\label{eq:flattening_ex}
		\varphi_{\text{IN1}}^\mathtt{F} = &\ \mathbf{G}\bigg(  \mathbf{P}\left(\lnot\mathrm{stop\_line\_in\_front}(\boldsymbol{x}_{\mathtt{ego}})\right) \nonumber\\
		&\qquad\vee  \mathrm{stop\_line\_in\_front}(\boldsymbol{x}_{\mathtt{ego}}) \nonumber\\
		&\qquad\vee \lnot \mathrm{at\_traffic\_sign\_stop}(\boldsymbol{x}_{\mathtt{ego}}) \\
		&\qquad\vee \mathrm{relevant\_traffic\_light}(\boldsymbol{x}_{\mathtt{ego}}) \nonumber\\
		&\qquad\vee \mathbf{O}\left( \mathbf{H}_{[0, t_\mathrm{slw}]} \left( \mathrm {stop\_line\_in\_front}(\boldsymbol{x}_{\mathtt{ego}}) \right. \right.\nonumber\\
		&\qquad\qquad\ \wedge\left.\left.\! \mathrm{in\_standstill}(\boldsymbol{x}_{\mathtt{ego}}) \right) \right) \bigg).\nonumber
		\end{align}}

\subsection{Distributive Decomposition}\label{subsec:Deduction} To further decompose $\varphi^\mathtt{F}$ into subformulas interconnected by disjunctions and conjunctions in a Boolean structure, denoted as $\varphi^\mathtt{D}$, we use the distributive properties of temporal operators \cite[Sec. 3.6]{warford2020calculational} \cite[Prop. 1]{zhang2023modularized}. For instance, for $\mathbf{G}$, we have:
	\begin{equation}
	\begin{aligned}\label{eq:approxi}
			\mathbf{G}(\varphi_1 \wedge \varphi_2) &\equiv \mathbf{G}(\varphi_1) \wedge \mathbf{G}(\varphi_2),\\
			 \mathbf{G}(\varphi_1 \lor \varphi_2) & \sledom  \mathbf{G}(\varphi_1) \lor \mathbf{G}(\varphi_2)\text{\footnotemark}.
		\end{aligned}	\end{equation}\footnotetext{The reversed entailment symbol $\sledom$  indicates that the formula on the right entails the formula on the left.}
	
	\noindent \textbf{Running Example:} Applying distributive decomposition, we reformulate (\ref{eq:flattening_ex}) as:
	\allowdisplaybreaks{
		\begin{align}\label{eq:ded_ex}
			\varphi_{\text{IN1}}^\mathtt{D} \sledom &\ \mathbf{G}\left(  \mathbf{P}\left(\lnot\mathrm{stop\_line\_in\_front}(\boldsymbol{x}_{\mathtt{ego}}) \right)\right)\nonumber \\
			&\lor \mathbf{G}\left(  \mathrm{stop\_line\_in\_front}(\boldsymbol{x}_{\mathtt{ego}})\right)\nonumber  \\
			&\lor \mathbf{G}\left( \lnot \mathrm{at\_traffic\_sign\_stop}(\boldsymbol{x}_{\mathtt{ego}}) \right) \\
			&\lor \mathbf{G}\left( \mathrm{relevant\_traffic\_light}(\boldsymbol{x}_{\mathtt{ego}}) \right)\nonumber \\
			&\lor \mathbf{G}\left( \mathbf{O}\left( \mathbf{H}_{[0, t_\mathrm{slw}]} \left( \mathrm {stop\_line\_in\_front}(\boldsymbol{x}_{\mathtt{ego}}) \right) \right. \right.\nonumber \\
			&\qquad\qquad \wedge \left. \left. \mathbf{H}_{[0, t_\mathrm{slw}]} \left( \mathrm{in\_standstill}(\boldsymbol{x}_{\mathtt{ego}}) \right) \right) \right).\nonumber
		\end{align}}
	\vspace{-5mm}
		\begin{rem}\label{rem:dis}
			As each subformula typically corresponds to a maneuver or a traffic condition (see (\ref{eq:ded_ex}) and Sec.~\ref{subsec:tc}), the approximation in (\ref{eq:approxi}) neglects scenarios where the vehicle frequently changes maneuvers or traffic conditions shift rapidly -- a situation that is uncommon over short planning horizons \cite[Sec. II-B]{Paden2016}.
		Moreover, traffic rules often represent causal relationships between subformulas, making it natural -- and often acceptable in practice -- to satisfy only {a subset of subformulas} within a globally scoped formula, {e.g., it suffices to satisfy $\varphi_2$ in $\mathbf{G}(\varphi_1\Rightarrow\varphi_2)$ or $\mathbf{G}(\varphi_1\lor\varphi_2\lor\dots)$.} 
		Therefore, the overapproximation has minimal impact on repair performance, as validated in \cite[Sec. V-D]{yuanfei2022}. \vspace{-2mm}
	\end{rem}
\subsection{CNF Conversion}\label{subsec:CNF}
To convert the decomposed formula $\varphi^\mathtt{D}$ into a form suitable for SAT solving, we replace subformulas connected by logical connectives with propositional variables $\sigma_j$, $j \in \mathbb{N}_+$. In contrast, subformulas containing nested temporal operators are kept intact and not further substituted, thereby preserving their structural integrity. This selective substitution simplifies the transformation of $\varphi^\mathtt{D}$ into an equivalent propositional formula $\varphi^\mathtt{P}$ in CNF  using, e.g., the \textit{Tseitin} transformation~\cite{tseitin1983complexity}.

	\vspace{2mm}
	\noindent \textbf{Running Example:} After converting  $\varphi_{\text{IN1}}^\mathtt{D}$ in (\ref{eq:ded_ex}) in CNF, the input for the SAT solver of rule $\varphi_{\text{IN1}}$ is: 
	\begin{equation}\label{eq:cnf_sat}
		\varphi_{\text{IN1}}^\mathtt{P} = \sigma_1\vee  \sigma_2 \vee \sigma_3 \vee  \sigma_4 \vee  \sigma_5,
	\end{equation}\vspace{-5mm}
	\begin{align*}
		\hspace{-3.5mm}\text{with}\enspace \enspace\enspace \sigma_1 &\coloneqq \mathbf{G}\left( 
		 \mathbf{P}\left(\lnot\mathrm{stop\_line\_in\_front}(\boldsymbol{x}_{\mathtt{ego}}) \right)\right), \\
		\sigma_2 &\coloneqq \mathbf{G}\left( \mathrm{stop\_line\_in\_front}(\boldsymbol{x}_{\mathtt{ego}}) \right), \\
		\sigma_3 &\coloneqq \mathbf{G}\left( \lnot \mathrm{at\_traffic\_sign\_stop}(\boldsymbol{x}_{\mathtt{ego}}) \right), \\
		\sigma_4 &\coloneqq \mathbf{G}\left( \mathrm{relevant\_traffic\_light}(\boldsymbol{x}_{\mathtt{ego}}) \right), \\
		\sigma_5 &\coloneqq \mathbf{G}\left( \mathbf{O}\left( \mathbf{H}_{[0, t_\mathrm{slw}]} \left( \mathrm {stop\_line\_in\_front}(\boldsymbol{x}_{\mathtt{ego}}) \right) \right. \right. \\
		&\qquad\qquad\ \wedge \left. \left. \mathbf{H}_{[0, t_\mathrm{slw}]} \left( \mathrm{in\_standstill}(\boldsymbol{x}_{\mathtt{ego}}) \right) \right) \right).
	\end{align*}
The violating truth valuations for the propositions in $\varphi_{\text{IN1}}^\mathtt{P}$ are  $\llbracket \sigma_1\rrbracket = \llbracket \sigma_2\rrbracket = \llbracket \sigma_3\rrbracket =\llbracket \sigma_4\rrbracket = \llbracket \sigma_5\rrbracket = \bot$.

\section{SAT Solver}
\label{sec:sat_solver}
We employ the popular Davis-Putnam-Logemann-Loveland (DPLL) algorithm \cite{davis1962machine} to solve the Boolean satisfiability of $\varphi^\mathtt{P}$ online. 
Since the robustness quantifies the extent of violation or satisfaction, it can serve as a heuristic to determine the order of evaluated atomic propositions. The evaluation of STL robustness fundamentally relies on the robustness of predicates, which, however, is not included in Def.~\ref{def:rob}. Therefore, we begin by computing the model predictive robustness for predicates in Sec.~\ref{subsec:mpr}, followed by the SAT solving process in Sec.~\ref{subsec:sat_solving}.
\subsection{Model Predictive Robustness}\label{subsec:mpr}
To incorporate the dynamics of underlying system models (e.g., (\ref{eq:motion_veh})) and streamline robustness evaluation across various predicates, we use model predictive robustness \cite{YuanfeiLinMPR}.
The fundamental concept is to systematically assess the model capability for adhering to rule predicates. 
We now present the overall algorithm for computing robustness step by step, including details on adapting predicates for different types of roads. 
\subsubsection{Overall Algorithm}
\begin{figure}[t!]
	\centering
	\vspace{.5mm}
	\begin{subfigure}[b]{1\columnwidth}
		\centering
		\def\svgwidth{0.99\columnwidth}
		\footnotesize
		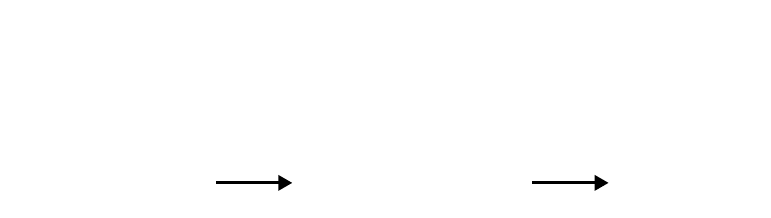
		\caption{Computation scheme for the predicates $p_1$ and $p_2$.}
		\vspace{1.5mm}
		\label{fig:mpr}
	\end{subfigure}
	\begin{subfigure}[b]{0.24\textwidth}
		\centering
		\def\svgwidth{0.99\columnwidth}
		\footnotesize
		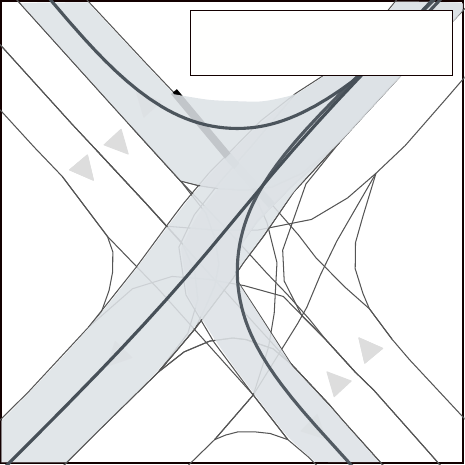
		\caption{Prediction model.}
		\label{fig:mpr_intersection1}
	\end{subfigure}
	\hfill
	\begin{subfigure}[b]{0.24\textwidth}
		\centering
		\def\svgwidth{0.99\columnwidth}
		\footnotesize
		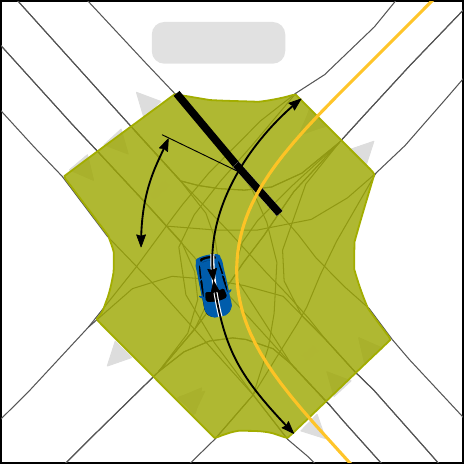
		\caption{Feature variables.}
		\label{fig:mpr_intersection2}
	\end{subfigure}
	\caption{Computation of model predictive robustness. (a) illustrates the computation scheme for predicates $p_1$, $p_2$, etc., in both offline and online modes. (b) displays the trajectory sampling results, where the states are labeled according to the predicate $\mathrm{stop\_line\_in\_front}(\boldsymbol{x}_{\mathtt{ego}})$.  (c) presents selected feature variables relevant to intersections.}
	\vspace{-4mm}
	\label{fig:mpr_intersection}
\end{figure}

An accurate distribution of future vehicle behaviors is essential in model predictive robustness \cite[(5)]{YuanfeiLinMPR}; however, obtaining such a distribution is nontrivial.
Therefore, we divide the computation of model predictive robustness into two stages: offline learning incorporating approximations of  possible future vehicle behaviors and online inference (cf. Fig.~\ref{fig:mpr}).

	The robustness is computed offline using real traffic data. To effectively explore the configuration space, a Monte Carlo simulation is used to sample a representative set of potential future states $\mathcal{X}^{\mathtt{MC}}$ for the ego vehicle over a finite prediction horizon, as described in \cite[Sec.~III-D]{YuanfeiLinMPR} (see Fig.~\ref{fig:mpr_intersection1}).
Future trajectories of other traffic participants are represented by recorded trajectories from the dataset. 
	For each predicate $p$, 
	model predictive robustness is computed by evaluating the relative frequency of future compliant states $\mathcal{X}^{\mathtt{c}}$ with an STL monitor \cite[Def.~2]{YuanfeiLinMPR}\footnote{For the sake of brevity, we omit the normalization step.}:
	\begin{equation*}
				\begin{aligned}
		\rho_{p}(\boldsymbol{\omega}, k) \coloneqq
		\begin{dcases}
		\frac{\lvert\mathcal{X}^{\mathtt{c}}\rvert}{\lvert\mathcal{X}^{\mathtt{MC}}\rvert}	& \ \text{if}\ \alpha(\omega_k)>0,\\
			-\frac{\lvert\mathcal{X}^{\mathtt{MC}}\rvert-\lvert\mathcal{X}^{\mathtt{c}}\rvert}{\lvert\mathcal{X}^{\mathtt{MC}}\rvert}	& \ \text{otherwise,}
		\end{dcases}
				\end{aligned}
	\end{equation*}
	where the sign denotes the satisfaction or violation of $p$ at time step $k$, {as determined by the evaluation function $\alpha(\omega_k)$}.
	Subsequently, pairs of feature variables that encompass relevant vehicular and environmental information based on $\omega_k$ (see Tab.~\ref{tab:feature_variables}) together with their corresponding robustness values $\rho_p(\boldsymbol{\omega}, k)$ are stored.  
	These stored pairs are then used to train individual Gaussian process (GP) \cite{williams2006gaussian} regression models for each predicate $p$, improving computational efficiency and reducing noise in robustness computation. These pretrained models are later utilized in online applications with the same feature variables. 

\renewcommand{\arraystretch}{1.12}
\begin{table}[t!]\small
	\begin{center}
		\vspace{.5mm}
		\caption{Feature variable definitions. All values presented are in SI units and at time step $k$ unless otherwise specified. To compute $\Delta_{\flat}$ or $\Delta_{\mathcal{L}}$, we use the signed distance from the vehicle center to its closest point at the road boundary $\flat$ or the bounds of 
				the lanelets $\mathcal{L}_{\mathtt{dir}}$.}
			\label{tab:feature_variables}
			\begin{tabular}{@{}p{1.6cm}p{4.8cm}p{1.59cm}@{}}
				\toprule[1.pt]
				\textbf{Feat. var.}  & \textbf{Description} & \textbf{Location} \\
				\midrule
				\multicolumn{3}{l}{\small \textbf{Rule-related}} \\
				$2\llbracket p_k \rrbracket - 1$ & Characteristic function\footnotemark \cite[Def. 1]{YuanfeiLinMPR} & All\\
				\midrule
				\multicolumn{3}{l}{\textbf{Ego-vehicle-related}} \\
				
				$\ell_{\mathtt{ego}}$, $w_{\mathtt{ego}}$	     & Vehicle length and width & All\\
				$\boldsymbol{x}_{\mathtt{ego}}$, $\boldsymbol{u}_{\mathtt{ego}}$ & State and input & All\\
				$\Delta_{\mathcal{L}_{\mathtt{l}}, \mathtt{ego}}$, $\Delta_{\mathcal{L}_{\mathtt{r}}, \mathtt{ego}}$            & Left and right distance to the left and right boundary of the lanelets $\mathcal{L}_{\mathtt{dir}}$  & All\\
				$\Delta_{\flat_{\mathtt{l}}, \mathtt{ego}}$, $\Delta_{\flat_{\mathtt{r}}, \mathtt{ego}}$ & Distance to the left and right road boundary& Interstates\\
				$s^{\text{entry}}_{\mathtt{ego}}$, $s^{\text{exit}}_{\mathtt{ego}}$ & Longitudinal distances to the intersection area & Intersections\\
				$s^{\text{stop}}_{\mathtt{ego}}$ & Longitudinal distance to the stop line & Intersections\\
				\midrule
				\multicolumn{3}{l}{\textbf{Other-vehicle-related}} \\
				
				$\ell_{\mathtt{obs}}$, $w_{\mathtt{obs}}$				         & Vehicle length and width  & All\\
				$\boldsymbol{x}_{\mathtt{obs}}$ & State & All\\
				$\Delta_{\mathcal{L}_{\mathtt{l}}, \mathtt{obs}}$, $\Delta_{\mathcal{L}_{\mathtt{r}}, \mathtt{obs}}$            & Left and right distance to the left and right boundary of the lanelets $\mathcal{L}_{\mathtt{dir}}$ & All\\
				$\Delta_{\flat_{\mathtt{l}}, \mathtt{obs}}$, $\Delta_{\flat_{\mathtt{r}}, \mathtt{obs}}$ & Distance to the left and right road boundary & Interstates\\
				$s^{\text{entry}}_{\mathtt{obs}}$, $s^{\text{exit}}_{\mathtt{obs}}$ & Longitudinal distances to the intersection area & Intersections\\
				\midrule
				\multicolumn{3}{l}{\textbf{Ego-other-relative}} \\
				
				$\Delta s_{\mathtt{ego}}$, $\Delta d_{\mathtt{ego}}$ & Relative longitudinal and lateral distance along $\Gamma_{\mathtt{ego}}$ & All \\
				$\Delta s_\mathtt{obs}$, $\Delta d_\mathtt{obs}$ & Relative longitudinal and lateral distance along $\Gamma_{\mathtt{obs}}$& Intersections \\
				$\Delta v_s, \Delta v_d$ & Relative velocity in the longitudinal and lateral directions along respective $\Gamma$ of each vehicle 
				& All\\
				\bottomrule[1.pt]
			\end{tabular}
			\vspace{-4mm}
		\end{center}
	\end{table}
\subsubsection{Extension to Intersections}
As the work \cite{YuanfeiLinMPR} primarily focuses on interstates, its application to intersections 
necessitates careful adjustments to both the trajectory sampling process in the prediction model and the feature variables.
\begin{table*}[!t]\centering\footnotesize
	
	\caption{Description of time-to-maneuver and the corresponding predicate categories. 
	}
	\renewcommand{\arraystretch}{1.2}
	\begin{tabular}{@{}llll@{}} \toprule
		\textbf{Time-to-maneuver  \cite{lin2023commonroad}} & \textbf{Description} & \textbf{Category \cite{irani2021computing}} & \textbf{Predicates \cite{Maierhofer2020a, maierhofer2022formalization}}\\ \midrule
		\makecell[l]{\textbf{Time-to-brake} \textbf{(TTB)}} & \makecell[l]{Full braking with maximum deceleration.}\vspace{0.2mm} & \multirow{2.8}{*}{\makecell[l]{\textbf{Longitudinal} \textbf{position}, \textbf{velocity}}} & \multirow{2}{*}{\makecell[l]{ $\mathrm{in\_conflict\_area}$, \\ $\mathrm{keeps\_lane\_speed\_limit}$, \\ $\mathrm{stop\_lane\_in\_front}$,  
				$\hdots$}} \\ 
		\makecell[l]{\textbf{Time-to-kick-down} \textbf{(TTK)}} & \makecell[l]{Full acceleration until reaching the maximum \\velocity and then maintaining the velocity.} &  & \\
		\makecell[l]{\textbf{Time-to-steer (TTS)}} & \makecell[l]{Full steering to reach a certain lateral offset or \\ a specified orientation change.\vspace{0.2mm}}  & \makecell[l]{\bf {Lateral position}}& \makecell[l]{$\mathrm{in\_same\_lane}$, $\mathrm{left\_of}$, \\$\mathrm{turning\_left}$, $\hdots$} \\
		\makecell[l]{\textbf{Time-to-maintain-velocity} \textbf{(TTMV)}}  & \makecell[l]{Maintaining a constant velocity.} & \textbf{Acceleration} & $\mathrm{brakes\_abruptly}$, $\hdots$\\
		\bottomrule
	\end{tabular}
	\label{tab:def_comp_man}
	\vspace{-2mm}
\end{table*}

	\paragraph{\it Trajectory Sampling}
	 It is important to note that vehicles at intersections exhibit a greater variety of velocity changes compared to those on interstates. Therefore, when the velocity falls below a certain threshold $v_{\text{switch}}\in \mathbb{R}_+$, the lateral trajectory generation is jointly considered with the longitudinal motion rather than being treated as an independent process \cite[Rem.~5]{werling2012optimal}.
Furthermore, the sampling process must account for all possible driving lanes within the road network, as illustrated in Fig.~\ref{fig:mpr_intersection1}.

	\paragraph{\it Feature Variables} 	
	Given the increased complexity of the road network at intersections compared to interstates, especially regarding regulatory elements, it is necessary to extend the features in \cite[Tab.~I]{YuanfeiLinMPR}. 
 Tab.~\ref{tab:feature_variables} summarizes the updated feature variables, categorized according to various locations, with some details provided as follows:  
	To effectively analyze intersection-related features, the intersection area is characterized as the region consisting of lanelets that are mutually accessible to vehicles approaching the intersection from various directions \cite{krasowski2022safe}, as shown in Fig.~\ref{fig:mpr_intersection2}. 	
	Consequently, the distance to the intersection area of a vehicle is represented by the longitudinal distance to its entry and exit points, denoted as $s^{\text{entry}}$ and $s^{\text{exit}}$, respectively, along its reference path $\Gamma$. Furthermore, if stop lines are present within $\mathcal{L}_{\mathtt{dir}}$, the longitudinal distance $s^{\text{stop}}_{\mathtt{ego}}$  from the ego vehicle to the nearest one along $\Gamma$ is chosen. Otherwise, $s^{\text{stop}}_{\mathtt{ego}}$ is assigned a default value \cite[Sec. III-A]{krasowski2022safe}. The other-vehicle-related and ego-other-relative feature variables are only applicable when the predicate involves other vehicles.
	Given that the other vehicle might travel in different directions at intersections, we additionally measure the relative distance between the vehicles along the reference path of the other vehicle $\Gamma_\mathtt{obs}$, denoted by $\Delta s_\mathtt{obs}$ and $\Delta d_\mathtt{obs}$. 
	\footnotetext{The function equals $1$ when $\llbracket p_k \rrbracket$ is $\top$; otherwise, it equals $-1$.} 

\subsection{SAT Solving}\label{subsec:sat_solving}

		Prior to exploration, the propositions are sorted in ascending order based on their absolute robustness values $\left|\rho_{\sigma_j}\right|$ over the time interval $[\mathtt{TV}_\varphi, h]$, where the rule violation occurs (cf. Def.~\ref{def:tv}). {This sorting strategy utilizes the monotonicity property of model predictive robustness \cite[Prop.~3]{YuanfeiLinMPR}: we consider smaller robustness values as an indication that the current truth valuation is more likely to change, thereby guiding the DPLL algorithm to flip predicate values with minimal effort.} The DPLL algorithm then systematically explores these sorted propositions by iteratively assigning truth values, performing unit propagation, and backtracking when conflicts arise~\cite{nieuwenhuis2006solving}. 
		Consequently, a partial solution $\phi$ -- consisting of truth valuations for a subset of propositions required to satisfy $\varphi^\mathtt{P}$  -- is incrementally constructed. 

	\vspace{2mm}
\noindent \textbf{Running Example:} We assume that the ranked order of the propositions in (\ref{eq:cnf_sat}) is $\sigma_2< \sigma_1< \sigma_5< \sigma_3< \sigma_4$.  In the first iteration of the SMT solver, the DPLL algorithm produces the partial solution $\llbracket \sigma_2\rrbracket=\top$ in the SAT solver, meaning the requirement is  $\mathbf{G}\left( \mathrm{stop\_line\_in\_front}(\boldsymbol{x}_{\mathtt{ego}}) \right)$.

\begin{figure}[t!]
	\centering
	\def\svgwidth{0.7\columnwidth}\footnotesize
	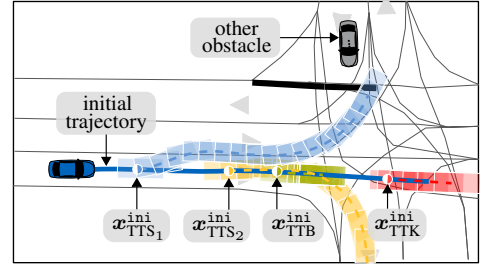
	\caption{Illustration of evasive maneuvers in Tab.~\ref{tab:def_comp_man}, depicting the starting points of each maneuver along with their associated occupancies. The TTS is demonstrated by reaching a specified lateral offset (TTS$_1$) and achieving a defined orientation change (TTS$_2$).}
	\vspace{-4mm}
	\label{fig:ttx}
\end{figure}
\section{$\mathcal{T}$-Solver}
\label{sec:T_solver}
In the $\mathcal{T}$-Solver, we solve (\ref{eq:problem_def}) by replacing $\varphi$ with the SAT solution $\phi$. 
Following  \cite[Alg.~2]{yuanfei2022}, we first determine $\mathtt{TC}_{\phi}$ -- based on $\phi$ and $\mathtt{TV}_\varphi$ -- in Sec.~\ref{subsec:tc}. 
 Next, we compute the specification-compliant reachable sets starting from $\mathtt{TC}_{\phi}$ in Sec.~\ref{subsec:reach}, followed by solving a convex optimization problem to obtain the repaired trajectory in Sec.~\ref{subsec:repair}. 

\subsection{Time-To-Comply Computation}\label{subsec:tc}
Since considering all possible reachable states to calculate the exact $\mathtt{TC}_{\phi}$ (cf. Def.~\ref{def:tc}) is computationally intractable, we employ an underapproximation that focuses on a selected set of maneuvers $\mathcal{M}$ \cite[Sec. IV-C]{yuanfei2022}. The goal is to alter the truth valuations of the atomic propositions $\phi_r \subseteq \phi$, which differ in their values at  $\mathtt{TV}_\varphi$\footnote{We use $\mathtt{TV}_\varphi$ instead of $[\mathtt{TV}_\varphi, h]$ as the violation valuations may change or resolve throughout the interval. If multiple rules $\varphi=\bigwedge_{\text{v}=1}^{n_\text{v}} \varphi_\text{v}$ are violated, the valuations are compared individually to those at $\mathtt{TV}_{\varphi_\text{v}}$.}   from the case where the propositional formula is satisfied, i.e., $\llbracket \varphi^\mathtt{P}\rrbracket = \top$. 
 The maneuvers are automatically determined by the categories relative to the domains of the predicates 
within $\phi_r$, as outlined in Tab.~\ref{tab:def_comp_man} and  illustrated in Fig.~\ref{fig:ttx}. 
Note that predicates within past-time temporal operators cannot guide actions for future compliance.
 Each maneuver is associated with a time-to-maneuver \cite{lin2023commonroad}, representing the latest possible time at which the maneuver can still be executed to comply with $\phi$ and calculated using \cite[Alg.~2]{Tamke.2011}. 
 According to  Def.~\ref{def:tc},  $\mathtt{TC}_{\phi}$ is then underapproximated by the maximum time-to-maneuver among the set $\mathcal{M}$. 

	\vspace{2mm}
\noindent \textbf{Running Example:} For the obtained SAT solution $\phi$ where $\llbracket \sigma_2 \rrbracket = \top$, i.e., $\phi_r=\phi$, the time-to-maneuver is the minimum of TTB and TTK, as the predicate $\mathrm{stop\_line\_in\_front}(\boldsymbol{x}_{\mathtt{ego}})$ belongs to the category of longitudinal position (cf.~Tab.~\ref{tab:def_comp_man}). 

\subsection{Specification-Compliant Reachability Analysis}\label{subsec:reach}
After the cut-off time step ${\mathtt{TC}_{\phi}}$, we adopt the approach proposed in \cite{lercher2024specification} to compute the reachable sets complying with $\phi$ in (\ref{eq:after_tc}) as the search space within the time interval $[\mathtt{TC}_{\phi}, h]$; however, any other reachability analysis can be used as well. 
 Since computing exact reachable sets is infeasible for efficiency reasons \cite{platzer2007image}, we use {an} 
overapproximation $\mathcal{R}_{\tau}$ of $\mathcal{R}^{\mathtt{e}}_{\tau}$ that encloses all kinematically feasible and specification-compliant trajectories. {The overapproximation guarantees that no valid trajectories are excluded. As soon as the reachable set becomes empty, we terminate the $\mathcal{T}$-solving process prematurely.} 
Note that the distributivity of $\mathbf{G}$ over $\lor$ in (\ref{eq:approxi}) imposes stricter rule requirements (cf. Rem.~\ref{rem:dis}). To mitigate its impact of overly shrinking the valid solution space, we follow the syntactic timing separation theorem \cite[Thm.~1]{zhang2023modularized} and relax $\mathbf{G}\varphi$ in the proposition when the overapproximation
is used during the propositional abstraction in Sec.~\ref{sec:rule_monitoring}\footnote{{Assuming we obtain the SAT solution of $\mathbf{G}(\varphi_1 \lor \varphi_2 \lor \dots)$ as any of the disjuncts, e.g., $\mathbf{G}\varphi_1$, relaxing the disjunct can help include parts of the solution space that are excluded by the overapproximation in  (\ref{eq:approxi}), thereby reducing conservativeness.}}:
\begin{align*}
	\mathbf{G}\varphi \models&\ 
	\mathbf{G}_{[0, \mathtt{TC}_{\phi}]} \varphi \wedge	\mathbf{G}_{[\mathtt{TC}_{\phi}, \mathtt{TV}_\varphi]} \varphi \wedge \mathbf{G}_{[\mathtt{TV}_\varphi, h]} \varphi \\  \models& \  \mathbf{G}_{[0, \mathtt{TC}_{\phi}]} \varphi \wedge \mathbf{G}_{[\mathtt{TV}_\varphi, h]}\varphi,
\end{align*}
where the requirement $\mathbf{G}_{[\mathtt{TC}_{\phi}, \mathtt{TV}_\varphi]}$ is omitted and the reachable sets remain overapproximated with respect to $\mathbf{G}\varphi$.  
In addition, collision avoidance is included as a default specification over the entire time horizon for safety concerns by conjuncting it with the SAT solution $\phi$
\cite{lercher2024specification}\footnote{For the interval  $[0, \mathtt{TC}_{\phi}]$, it is assumed that $\chi^{\mathtt{ini}}$ is 	kinematically feasible and collision-free.}. 
With this, we perform automaton-based model checking on the fly \cite{lercher2024specification} to compute the specification-compliant reachable sets. {This involves rewriting the formula in LTL over finite traces, e.g., using the tool Spot \cite{duret2016spot}.} 

\begin{figure}[t!]
	\centering
	\vspace{0mm}
	\def\svgwidth{0.95\columnwidth}\footnotesize
	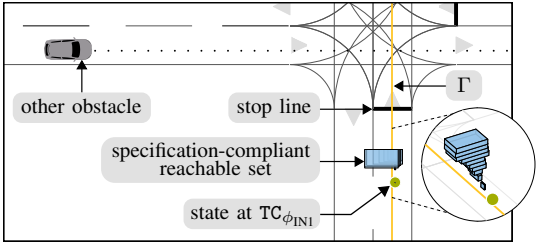
	\caption{{Examples of specification-compliant reachable sets, which are projected onto the position domain.}}
	\vspace{-3mm}
	\label{fig:reach_stop_line}
	
\end{figure}

\vspace{2mm}
\noindent \textbf{Running Example:} Given the SAT solution $\llbracket \sigma_2 \rrbracket = \top$, Fig.~\ref{fig:reach_stop_line} shows the computed reachable sets over the time interval $[\mathtt{TC}_{\phi},h]$ satisfying $\mathbf{G}_{[0, \mathtt{TC}_{\phi}]} \left(\mathrm{stop\_line\_in\_front}(\boldsymbol{x}_{\mathtt{ego}})\right)  \wedge \mathbf{G}_{[\mathtt{TV}_\varphi, h]}\left(\mathrm{stop\_line\_in\_front}(\boldsymbol{x}_{\mathtt{ego}}) \right)$, which significantly reduce the search space for finding repaired trajectories compared to the entire configuration space.

\subsection{Optimization-Based Trajectory Repair}\label{subsec:repair}
From the specification-compliant reachable sets, we first extract an optimal driving corridor \cite[Sec. II-E]{liu2023specification}. We then formulate a convex optimization problem \cite{boyd2004convex} whose solution yields repaired trajectories confined to this corridor and is characterized by fast convergence. 	
Building on this, the optimization problem in (\ref{eq:problem_def}) over the time interval $[\mathtt{TC}_{\phi},h]$ 
 is solved using the approach detailed in \cite{manzinger2020using}. 
{At each time step between $\mathtt{TC}_{\phi}$ and $h$, the vehicle motion is decomposed into longitudinal and lateral directions along the reference path $\Gamma$ \cite[Sec. II-D]{manzinger2020using}. 
 	The convex cost function  $J$ can be chosen, e.g., to favor comfort by penalizing high accelerations and jerks \cite[(12) and (18)]{pek2020fail}.}
{Moreover, the rule constraints are defined by the interval between the lower and upper bounds of the reachable sets.}
Since the reachable sets are 
{overapproximated}, the computed trajectory may exhibit slight violations of $\varphi$. Thus, we only return the repaired trajectory if it is verified to be rule-compliant.

\section{Experimental Results}
\label{sec:experiment}
In this section, we demonstrate the effectiveness of our framework in addressing multiple rule violations and handling complex, real-world scenarios. Following the implementation details in Sec.~\ref{sec:imp_details}, we focus on repairing multiple interstate rule violations in Sec.~\ref{subsec:multi}.  We then show the scalability of our approach to intersection traffic rules in Sec.~\ref{subsec:intersection}, compare it with related work in Sec.~\ref{subsec:comp}, and provide further discussions in Sec.~\ref{subsec:disuccsion}. Finally, we validate our method through closed-loop planning tests conducted within the realistic simulator in Sec.~\ref{subsec:carla} and in the real world in Sec.~\ref{subsec:realworld}.
\subsection{Implementation Details}\label{sec:imp_details}

\subsubsection{General Settings} Our approach is implemented in Python, and simulations were conducted on a single thread using a machine equipped with an AMD EPYC 7763 64-core processor and 2TB of RAM. In our implementation, we use GPyTorch \cite{gardner2018gpytorch} to model and solve GP regression and RTAMT \cite{yamaguchi2024rtamt} to monitor rules and evaluate robustness. The CommonRoad-CriMe toolbox \cite{lin2023commonroad} is employed to derive the cut-off state, while the CommonRoad-Reach toolbox \cite{reach} is used to compute the reachable sets from that state. Finally, the trajectory optimization problem is solved using the Gurobi solver \cite{gurobi} to obtain the repaired trajectory.
Tab.~\ref{tab:params} lists the selected parameters, while all other parameters remain consistent with those defined in the original paper.

\begin{table}[!t]\centering\small
	
\caption{General parameters for the numerical experiments (all variables are expressed in SI units).}
	
	\renewcommand{\arraystretch}{1.06}
	\begin{tabular}{@{}ll@{}} \toprule
		\textbf{Description} & \textbf{Notation and value}\\ \midrule
		\multicolumn{1}{l}{\noindent\bf{Model predictive robustness}} &\\
		Number of samples & $1500$ \\
		Prediction horizon & $1.56\unit{\second}$ \\
		Velocity sampling interval & $[\max(0, v_k\! -\! 17.25), v_k\! +\! 17.25]\!\!$\\
		Lateral position sampling interval & \\
		\enspace\enspace High speed mode & $[d_k - 5, d_k + 5]$ \\
		\enspace\enspace Low speed mode & $[d_k - 1.5, d_k + 1.5]$\\
		Derivative of lateral position& \\
		\enspace\enspace High speed mode & $[-3, 3]\unit{m/s}$ \\
		\enspace\enspace Low speed mode & $[-0.2, 0.2]\unit{m/s}$ \\
		\makecell[l]{Switching velocity} & $v_{\text{switch}}=4\unit{m/s}$\\ 
		 \midrule
		\multicolumn{1}{l}{\noindent\bf{Traffic rule monitoring}} & \\
		Waiting time duration & $t_{\mathrm{slw}} = 3\unit{s}$ (cf. (\ref{eq:rule_rin1}))\\
		Time to restore a safe distance & $t_\text{c} = 3\unit{s}$ (cf. (\ref{eq:rule_rG13})) \\
		Waiting time, entry delay& \makecell[l]{$t_{\mathrm{ia}} = 0.5\unit{s}$,  $t_{\mathrm{ib}} = 1\unit{s}$ \\(cf. Sec.~\ref{subsec:priority})}\\
		\midrule
		\multicolumn{1}{l}{\noindent\bf{Dataset evaluation}} &\\
		Time step size & $\Delta t$ = $0.2\unit{s}$\\
		Maximum planning horizon & $h = 20$\\
	 \midrule
		\multicolumn{1}{l}{\noindent\bf{CARLA simulation}} & \\
		Time step size, planning horizon  & $\Delta t$ = $0.1\unit{s}$, $h = 30$\\
		Replanning cycle & $1.0\unit{s}$\\\midrule
		\multicolumn{1}{l}{\noindent\bf{Real-world experiments}} & \\
		Time step size, 	planning horizon & $\Delta t$ = $0.1\unit{s}$, $h = 50$\\
		Replanning cycle & $0.2\unit{s}$ \\
		\bottomrule
	\end{tabular}	\vspace{-1mm}
	\label{tab:params}
\end{table}

\subsubsection{Vehicle Models}\label{subsec:vm} 
For model predictive robustness computation and trajectory optimization, we use the vehicle models presented in \cite{pek2020fail}: the longitudinal dynamics are represented as a fourth-order integrator, while the lateral dynamics are modeled by a linearized kinematic single-track model, with both formulated in the curvilinear coordinate system.
\subsubsection{Traffic Rules and Scenarios}\label{subsec:scenario} 
We adopt the formalized German interstate \cite{Maierhofer2020a} and intersection \cite{maierhofer2022formalization} traffic rules. To train the GP regression model for online computation of model predictive robustness, we use scenarios derived from the highD \cite{highd} and inD \cite{bock2020ind} datasets. These drone-captured datasets provide recorded traffic data from highway and urban locations in Germany, ensuring consistency with the formalized traffic rules.
During the evaluation, we convert the raw data into CommonRoad scenarios \cite{Althoff2017a} using an open-source data converter\footnote{\url{https://commonroad.in.tum.de/tools/dataset-converters}}, with each scenario limited to a maximum duration of $4\unit{s}$. 
 We focus on scenarios containing rule violations and designate the violating vehicle as the ego vehicle.

\newcommand{\minus}{\scalebox{0.75}[1.0]{$-$}}
\renewcommand{\cellgape}{\Gape[0.2em]}
\begin{table}[!t]\centering\small
	\caption{Key parameters obtained during trajectory repair.}
	\renewcommand{\arraystretch}{1.1}
	\begin{tabular}{@{}l@{\hskip 0.06in}ll@{\hskip 0.05in}l@{}} \toprule 
		{\textbf{Parameter}} & & \multicolumn{2}{@{}l}{\textbf{Values}}\\ \midrule
		\multicolumn{4}{l}{\textbf{Multi-rule violation}} \\
	{${\mathtt{TV}_{\varphi}}$} & &	\multicolumn{2}{@{}l}{$13$ ($\mathtt{TV}_{\varphi_{\text{G1}}} =13$, $\mathtt{TV}_{\varphi_{\text{G3}}} =15$)}\\
	\makecell[l]{${\rho_{\sigma_j}}$
		{over}\\ ${[\mathtt{TV}_\varphi, h]}$ }& & \multicolumn{2}{@{}l}{\makecell[l]{$\sigma_1 \colon\minus0.351$, $\sigma_2\colon\minus0.971$,  $\sigma_3 \colon\minus0.236$, $\sigma_{4} \colon\minus0.295$,\\
			$\sigma_{5} \colon0.692$, $\sigma_{6} \colon0.786$, $\sigma_{7} \colon0.903$, $\sigma_{8} \colon\minus0.032$}}\\ 
	\multirow{2.5}{*}{\makecell[c]{ \textit{1. iteration}\\\cmark}} & \multicolumn{2}{l}{$\phi$} & {\makecell[l]{$\llbracket \sigma_3\rrbracket =\top$, $ \llbracket \sigma_5\rrbracket=\top$, $
			\llbracket \sigma_6\rrbracket=\top$, \\$\llbracket\sigma_{7}\rrbracket=\top$, $\llbracket\sigma_{8}\rrbracket=\top$}}\\
	& \multicolumn{2}{l}{$\mathtt{TC}_{\phi}$} & $\max\{\text{TTB}, \text{TTK}, \text{TTS}\} = \text{TTB} = 12$\\ 
	\midrule
	\multicolumn{4}{l}{\textbf{Stop-line rule}}\\ 
	$\mathtt{TV}_{\varphi}$ & & \multicolumn{2}{@{}l}{$15$}\\
	\makecell[l]{$\rho_{\sigma_j}$
		{over}\\ $[\mathtt{TV}_\varphi, h]$ }&  & \multicolumn{2}{@{}l}{\makecell[l]{$\sigma_1 \colon\minus0.001$, $\sigma_2\colon\minus0.968$,  $\sigma_3 \colon\minus1.000$, $\sigma_{4} \colon\minus1.000$, \\
			$\sigma_{5} \colon\minus0.970$}}\\ 
	{\makecell[c]{ \textit{1. iteration}\\\xmark}} & \multicolumn{2}{l}{$\phi$, $\mathtt{TC}_{\phi}$} & $\llbracket \sigma_1\rrbracket =\top$, $\minus\infty$\\
	\multirow{2}{*}{\makecell[c]{ \textit{2. iteration}\\\cmark}} & \multicolumn{2}{l}{$\phi$} & $\llbracket \sigma_1\rrbracket =\bot$, $\llbracket \sigma_2\rrbracket =\top$\\ 
	& \multicolumn{2}{l}{$\mathtt{TC}_{\phi}$} & $\max\{\text{TTB}, \text{TTK}\} =\text{TTB} = 12$\\ \midrule
	\multicolumn{4}{l}{\textbf{Priority rule}}\\ 
	$\mathtt{TV}_{\varphi}$  & & \multicolumn{2}{@{}l}{$9$}\\
	\makecell[l]{$\rho_{\sigma_j}$
		{over}\\ $[\mathtt{TV}_\varphi, h]$ }&  &\multicolumn{2}{@{}l}{\makecell[l]{$\sigma_1 \colon\minus0.221$, $\sigma_2\colon\minus0.628$,  $\sigma_3 \colon\minus0.389$, $\cdots$, \\ $\sigma_{10} \colon\minus0.773$,
			$\sigma_{11} \colon0.965$, $\sigma_{12} \colon\minus0.741$, \\ $\sigma_{13} \colon\minus0.741$, 
			$\sigma_{14} \colon\minus0.773$, $\sigma_{15} \colon\minus1.000$}}\\ 	
	
	\multirow{2.5}{*}{\makecell[c]{ \textit{1. iteration}\\\cmark}} & \multicolumn{2}{l}{$\phi$} & \makecell[l]{$\llbracket \sigma_1\rrbracket = \llbracket \sigma_2\rrbracket = \llbracket \sigma_3\rrbracket = \dots = \llbracket \sigma_9\rrbracket = \bot$,  \\ $\llbracket\sigma_{10}\rrbracket =\top$, $\llbracket\sigma_{14}\rrbracket =\top$}\\
	& \multicolumn{2}{l}{$\mathtt{TC}_{\phi}$} & $\max\{\text{TTB}, \text{TTK}\} =\text{TTB} =5$\\
	\bottomrule
\end{tabular}\label{tab:prop_robust}
\vspace{-4mm}
\end{table}

\begin{figure*}[t!]
	\centering
	\begin{subfigure}[t]{\textwidth}
		\centering\footnotesize
		\def\svgwidth{0.99\linewidth}
		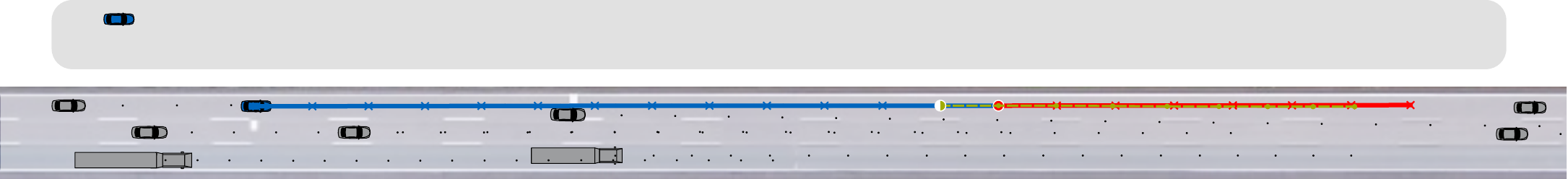
		\caption{Scenario at $k=0$ with the initial and repaired trajectories.}\vspace{1.5mm}
		\label{fig:rg1_initial}
	\end{subfigure}
	\hfill
	\begin{subfigure}[t]{\textwidth}
		\centering
		\def\svgwidth{0.99\linewidth}
		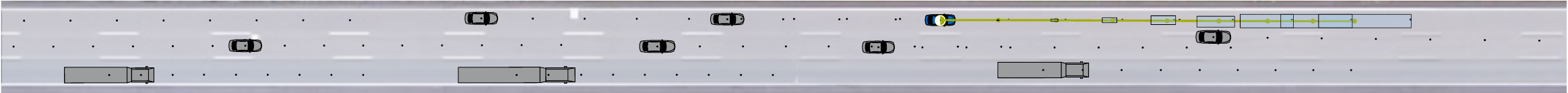
		\caption{Scenario at $\mathtt{TC}_{\phi_{\text{G1, G3}}}=12$ with the repaired trajectory, {generated within the specification-compliant reachable sets that satisfy $\phi_{\text{G1, G3}}$.}}\vspace{1.5mm}
		\label{fig:rg1_tc}
	\end{subfigure}
	\hfill
	\begin{subfigure}[t]{\textwidth}
		\centering
		\def\svgwidth{0.99\linewidth}
		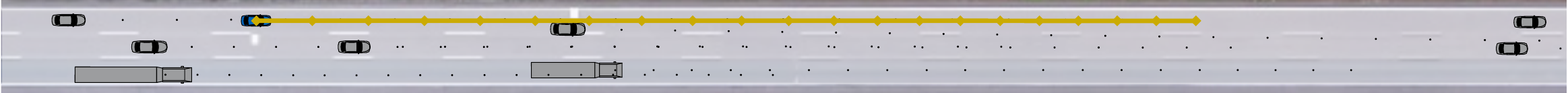
		\caption{{Scenario at $k=0$ with the} replanned trajectory using MICP-based planner, {complying with $\varphi_{\text{G1, G3}}$}.}\vspace{1.5mm}
		\label{fig:interstate_replan}
	\end{subfigure}
	\begin{subfigure}[t]{\textwidth}
		\centering
		\def\svgwidth{0.99\linewidth}
		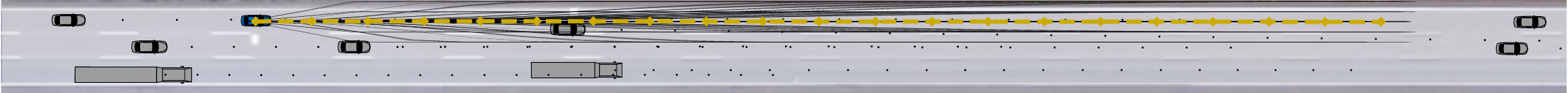
		\caption{{Scenario at $k=0$ with the} replanned trajectory using sampling-based planner, {complying with $\varphi_{\text{G1, G3}}$}.}\vspace{1.5mm}
		\label{fig:interstate_replan_sample}
	\end{subfigure}\hfill
	\begin{subfigure}[t]{0.32\textwidth}
		\centering\footnotesize
		\def\svgwidth{\linewidth}
		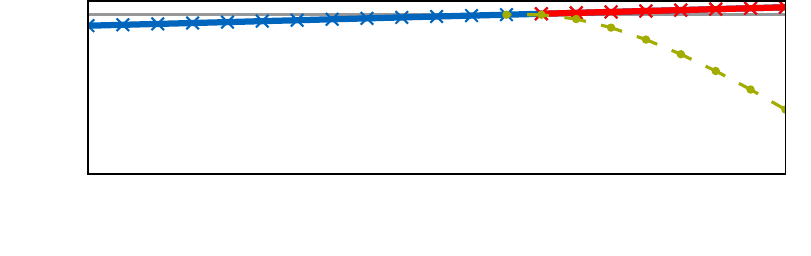
		\caption{Velocity profile of the repaired trajectory.}
		\label{fig:interstate_v}
	\end{subfigure}\hfill
	\begin{subfigure}[t]{0.32\textwidth}
		\centering\footnotesize
		\def\svgwidth{\linewidth}
		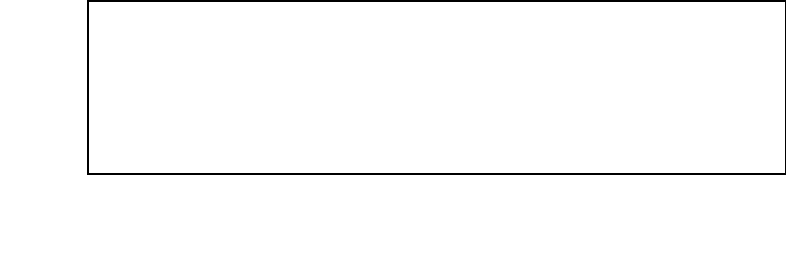
		\caption{Velocity profile for MICP-based replanning.}
		\label{fig:interstate_replan_v}
	\end{subfigure}\hfill
	\begin{subfigure}[t]{0.32\textwidth}
		\centering\footnotesize
		\def\svgwidth{\linewidth}
		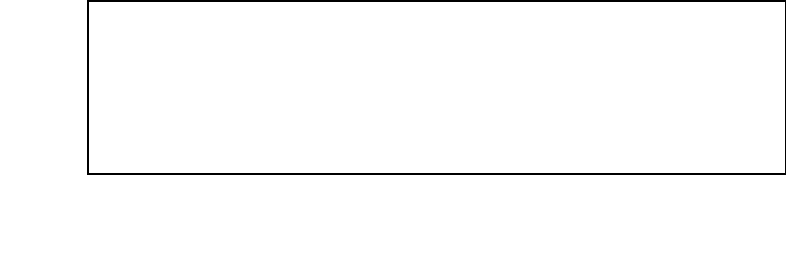
		\caption{Velocity profile for sampling-based  replanning.}
		\label{fig:interstate_replan_v_2}
	\end{subfigure}
	\caption{Highway scenario where the ego vehicle violates safe-distance and speed-limit rules. (a) and (b) show the initial and repaired trajectories, with a velocity comparison in (e). (c) and (d) present the replanned trajectories using MICP-based and sampling-based planners, respectively, with their velocity profiles shown in (f) and (g).}
	\vspace{-3mm}
	\label{fig:rg13-general}
\end{figure*}
\subsection{Multiple Traffic Rule Violations}\label{subsec:multi}
We begin by considering a highway scenario from the highD dataset, where the initial trajectory of the ego vehicle violates multiple traffic rules.  As illustrated in Fig.~\ref{fig:rg13-general}, these violations include the safe-distance rule $\varphi_{\text{G1}}$ and the speed-limit rule $\varphi_{\text{G3}}$ \cite{Maierhofer2020a}. The violated rules are combined into the conjunctive formula $\varphi_{\text{G1, G3}} =\varphi_{\text{G1}} \wedge \varphi_{\text{G3}} $, where
\begin{equation}\label{eq:rule_rG13}
	\begin{aligned}
	\varphi_{\text{G1}} \!= \! \mathbf{G}\big( &\mathrm{in\_same\_lane}(\boldsymbol{x}_{\mathtt{ego}}, \boldsymbol{x}_{\mathtt{obs}})\! \wedge\! \mathrm{behind}(\boldsymbol{x}_{\mathtt{ego}}, \boldsymbol{x}_{\mathtt{obs}}) \\
		& \wedge \lnot \mathbf{O}_{[0, t_\text{c}]} \big(\mathrm{cut\_in}(\boldsymbol{x}_{\mathtt{obs}}, \boldsymbol{x}_{\mathtt{ego}}) \wedge \\& \hspace{3.8em} \mathbf{P}(\lnot \mathrm{cut\_in}(\boldsymbol{x}_{\mathtt{obs}}, \boldsymbol{x}_{\mathtt{ego}}))\big)\\
		& \!\!\Rightarrow \mathrm{keeps\_safe\_distance\_prec}(\boldsymbol{x}_{\mathtt{ego}}, \boldsymbol{x}_{\mathtt{obs}}) \big) \ \text{and}\\
				\varphi_{\text{G3}}\! = \! \mathbf{G}\big(& \mathrm{keeps\_lane\_speed\_limit}(\boldsymbol{x}_{\mathtt{ego}}) \wedge \\
		& \mathrm{keeps\_type\_speed\_limit}(\boldsymbol{x}_{\mathtt{ego}}) \wedge\\
		&\mathrm{keeps\_fov\_speed\_limit}(\boldsymbol{x}_{\mathtt{ego}}) \wedge\\
		&  \mathrm{keeps\_braking\_speed\_limit}(\boldsymbol{x}_{\mathtt{ego}})\big)\text{.}
\end{aligned}
\end{equation}

\noindent  By abstracting the formula (cf. Sec.~\ref{sec:rule_monitoring}), we obtain the input for the SAT solver as:
\begin{flalign*}
	\varphi^{\mathtt{P}}_{\text{G1, G3}} &=  \underbrace{( \sigma_1\vee  \sigma_2 \vee \sigma_3 \vee  \sigma_4 )}_{\varphi_{\text{G1}}} \wedge  \underbrace{\sigma_5 \wedge  \sigma_6 \wedge  \sigma_7 \wedge  \sigma_8}_{\varphi_{\text{G3}}}, \\
	{\text{with}} \qquad  \sigma_1 &\coloneqq \mathbf{G}\left(\mathrm{keeps\_safe\_distance\_prec}(\boldsymbol{x}_{\mathtt{ego}}) \right), \\
	\sigma_2 &\coloneqq \mathbf{G}\left( \lnot\mathrm{behind}(\boldsymbol{x}_{\mathtt{ego}}, \boldsymbol{x}_{\mathtt{obs}}) \right), \\
	\sigma_3 &\coloneqq \mathbf{G}\left( \lnot \mathrm{in\_same\_lane}(\boldsymbol{x}_{\mathtt{ego}}, \boldsymbol{x}_{\mathtt{obs}}) \right), \\
	\sigma_4 &\coloneqq \mathbf{G}\left( \mathbf{O}_{[0, t_\text{c}]} \big(\mathrm{cut\_in}(\boldsymbol{x}_{\mathtt{obs}}, \boldsymbol{x}_{\mathtt{ego}}) \right.\\
	& \hspace{60pt} \left.\wedge \mathbf{P}(\lnot \mathrm{cut\_in}(\boldsymbol{x}_{\mathtt{obs}}, \boldsymbol{x}_{\mathtt{ego}}))\big) \right),\\
	\sigma_5 &\coloneqq \mathbf{G}\left( \mathrm{keeps\_lane\_speed\_limit}(\boldsymbol{x}_{\mathtt{ego}}) \right),\\
	\sigma_6 &\coloneqq \mathbf{G}\left( \mathrm{keeps\_type\_speed\_limit}(\boldsymbol{x}_{\mathtt{ego}}) \right),\\
	\sigma_7 &\coloneqq \mathbf{G}\left( \mathrm{keeps\_fov\_speed\_limit}(\boldsymbol{x}_{\mathtt{ego}}) \right),\\
	\sigma_8 &\coloneqq \mathbf{G}\left( \mathrm{keeps\_braking\_speed\_limit}(\boldsymbol{x}_{\mathtt{ego}}) \right).
\end{flalign*}
Using the SMT framework, we obtain the repaired trajectory, as shown in Fig.~\ref{fig:rg1_initial}, with its key parameters listed in Tab.~\ref{tab:prop_robust}. In the first iteration, the SAT solver suggests a change in the truth valuations of the propositions $\sigma_3$ and $\sigma_{8}$, which correspond to the maneuvers of braking, accelerating, and steering.  In the $\mathcal{T}$-solver, we determine that $\mathtt{TC}_{\phi_{\text{G1, G3}}}=\text{TTB}$ and obtain the specification-compliant reachable sets depicted in Fig.~\ref{fig:rg1_tc}. Consequently, the ego vehicle performs a braking maneuver to comply with the violated rules, which is evident from the velocity profile in Fig.~\ref{fig:interstate_v}.

\subsection{Intersection Traffic Rule Violation}
\label{subsec:intersection}
In this section, we demonstrate our trajectory repair approach against typical intersection rules using inD scenarios.
\subsubsection{Stop-Line Rule}\label{subsec:stop_line} 
\begin{figure}[t!]
	\centering
	\vspace{-2mm}
	\begin{subfigure}[t]{0.48\columnwidth}
		\centering\footnotesize
		\def\svgwidth{0.99\linewidth}
\begingroup%
  \makeatletter%
  \providecommand\color[2][]{%
    \errmessage{(Inkscape) Color is used for the text in Inkscape, but the package 'color.sty' is not loaded}%
    \renewcommand\color[2][]{}%
  }%
  \providecommand\transparent[1]{%
    \errmessage{(Inkscape) Transparency is used (non-zero) for the text in Inkscape, but the package 'transparent.sty' is not loaded}%
    \renewcommand\transparent[1]{}%
  }%
  \providecommand\rotatebox[2]{#2}%
  \newcommand*\fsize{\dimexpr\f@size pt\relax}%
  \newcommand*\lineheight[1]{\fontsize{\fsize}{#1\fsize}\selectfont}%
  \ifx\svgwidth\undefined%
    \setlength{\unitlength}{524.40002441bp}%
    \ifx\svgscale\undefined%
      \relax%
    \else%
      \setlength{\unitlength}{\unitlength * \real{\svgscale}}%
    \fi%
  \else%
    \setlength{\unitlength}{\svgwidth}%
  \fi%
  \global\let\svgwidth\undefined%
  \global\let\svgscale\undefined%
  \makeatother%
  \begin{picture}(1,1.05873376)%
    \lineheight{1}%
    \setlength\tabcolsep{0pt}%
    \put(0,0){\includegraphics[width=\unitlength,page=1]{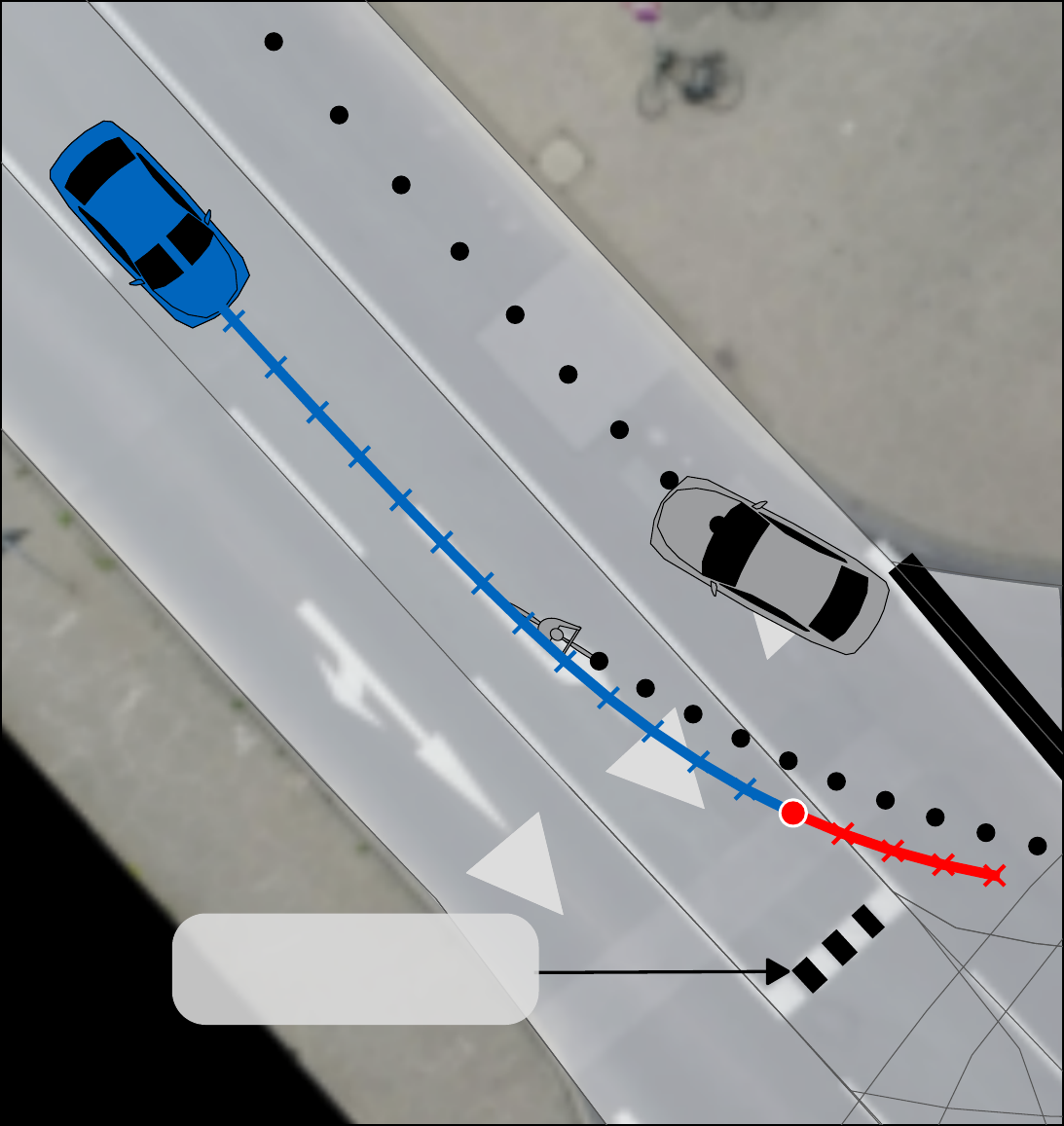}}%
    \put(0.20895528,0.12478146){\color[rgb]{0,0,0}\makebox(0,0)[lt]{\lineheight{1.25}\smash{\begin{tabular}[t]{l}Stop line\end{tabular}}}}%
  \end{picture}%
\endgroup%

		\caption{Scenario at $k=0$.}
		\label{fig:rin1_initial}
	\end{subfigure}
	\hfill
	\begin{subfigure}[t]{0.48\linewidth}
		\centering\footnotesize
		\def\svgwidth{0.99\linewidth}
		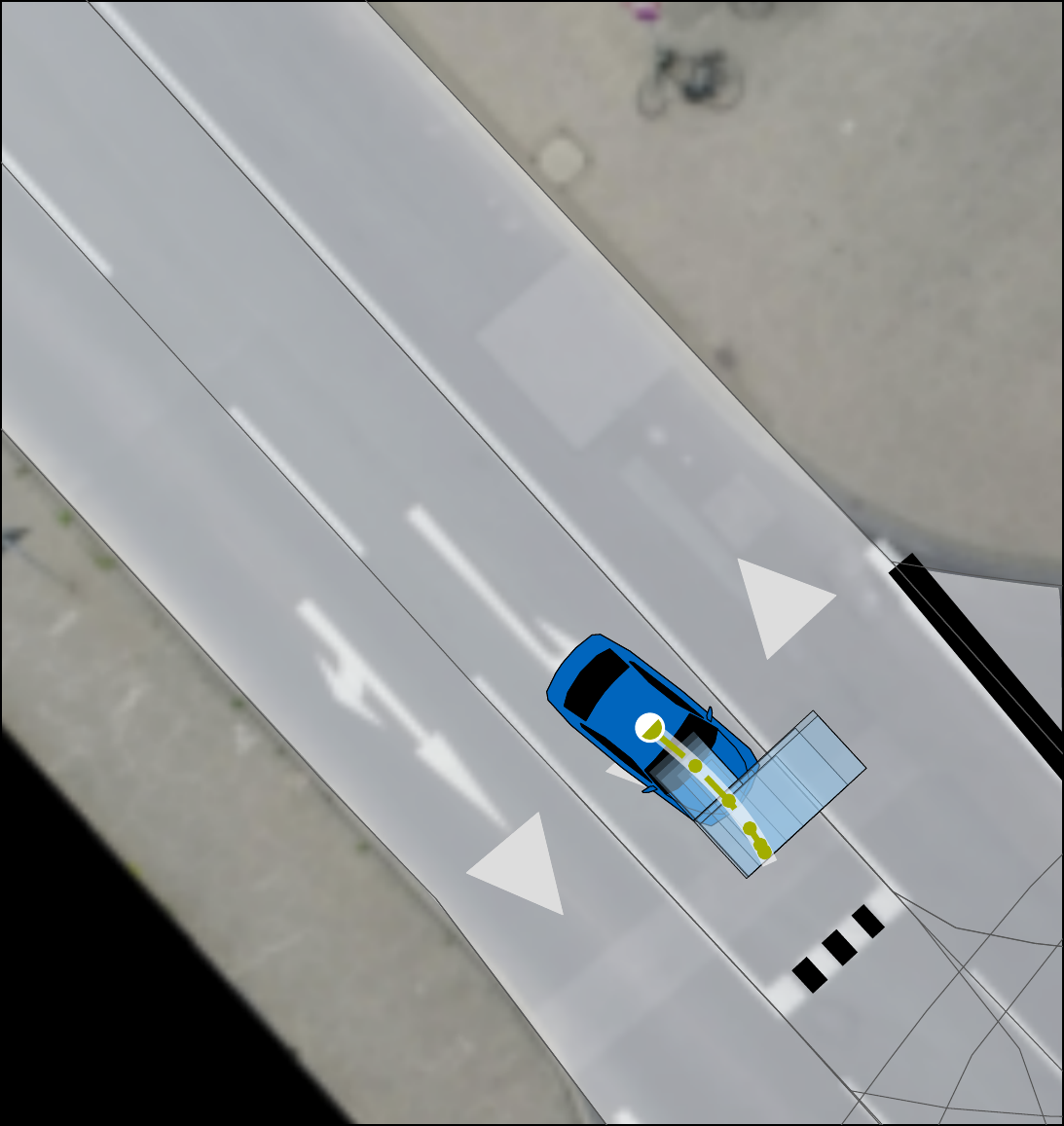
		\caption{Scenario at $\mathtt{TC}_{\phi_{\text{IN1}}}=12$.}
		\label{fig:rin1_repair}
	\end{subfigure}
	\hfill
	\caption{Rural scenario where the initial trajectory of the ego vehicle in (a) violates the stop-line rule, which is repaired by our approach in (b). The legends are the same as in Fig.~\ref{fig:rg13-general}.}
	\vspace{-4mm}
	\label{fig:rini_ini}
\end{figure}

Consider the stop-line rule in (\ref{eq:rule_rin1}), where a violation occurs if the ego vehicle fails to stop for $t_\text{slw}$ behind the stop line before crossing it, as shown in Fig.~\ref{fig:rin1_initial}. In the first iteration of the repair process, the SAT solver returns the solution  $\llbracket \sigma_1\rrbracket =\top$ (cf. (\ref{eq:cnf_sat}) and Tab.~\ref{tab:prop_robust}), which does not correspond to any maneuvers due to the presence of a past-time temporal operator in $\sigma_1$. Consequently, the second iteration yields the solution $\llbracket \sigma_1\rrbracket=\bot$, $\llbracket \sigma_2\rrbracket =\top$, leading to braking and kick-down maneuvers, resulting in $\mathtt{TC}_{\phi_{\text{IN1}}}=12$.  After performing trajectory repair, the initial trajectory is adjusted to slow down from $\mathtt{TC}_{\phi}$ and come to a full stop before the stop line (cf. Fig.~\ref{fig:rin1_repair}).

\begin{figure*}[t!]
	\centering
	\begin{subfigure}[t]{0.32\linewidth}
		\centering\footnotesize
		\def\svgwidth{0.99\linewidth}
\begingroup%
  \makeatletter%
  \providecommand\color[2][]{%
    \errmessage{(Inkscape) Color is used for the text in Inkscape, but the package 'color.sty' is not loaded}%
    \renewcommand\color[2][]{}%
  }%
  \providecommand\transparent[1]{%
    \errmessage{(Inkscape) Transparency is used (non-zero) for the text in Inkscape, but the package 'transparent.sty' is not loaded}%
    \renewcommand\transparent[1]{}%
  }%
  \providecommand\rotatebox[2]{#2}%
  \newcommand*\fsize{\dimexpr\f@size pt\relax}%
  \newcommand*\lineheight[1]{\fontsize{\fsize}{#1\fsize}\selectfont}%
  \ifx\svgwidth\undefined%
    \setlength{\unitlength}{575bp}%
    \ifx\svgscale\undefined%
      \relax%
    \else%
      \setlength{\unitlength}{\unitlength * \real{\svgscale}}%
    \fi%
  \else%
    \setlength{\unitlength}{\svgwidth}%
  \fi%
  \global\let\svgwidth\undefined%
  \global\let\svgscale\undefined%
  \makeatother%
  \begin{picture}(1,0.96556524)%
    \lineheight{1}%
    \setlength\tabcolsep{0pt}%
    \put(0,0){\includegraphics[width=\unitlength,page=1]{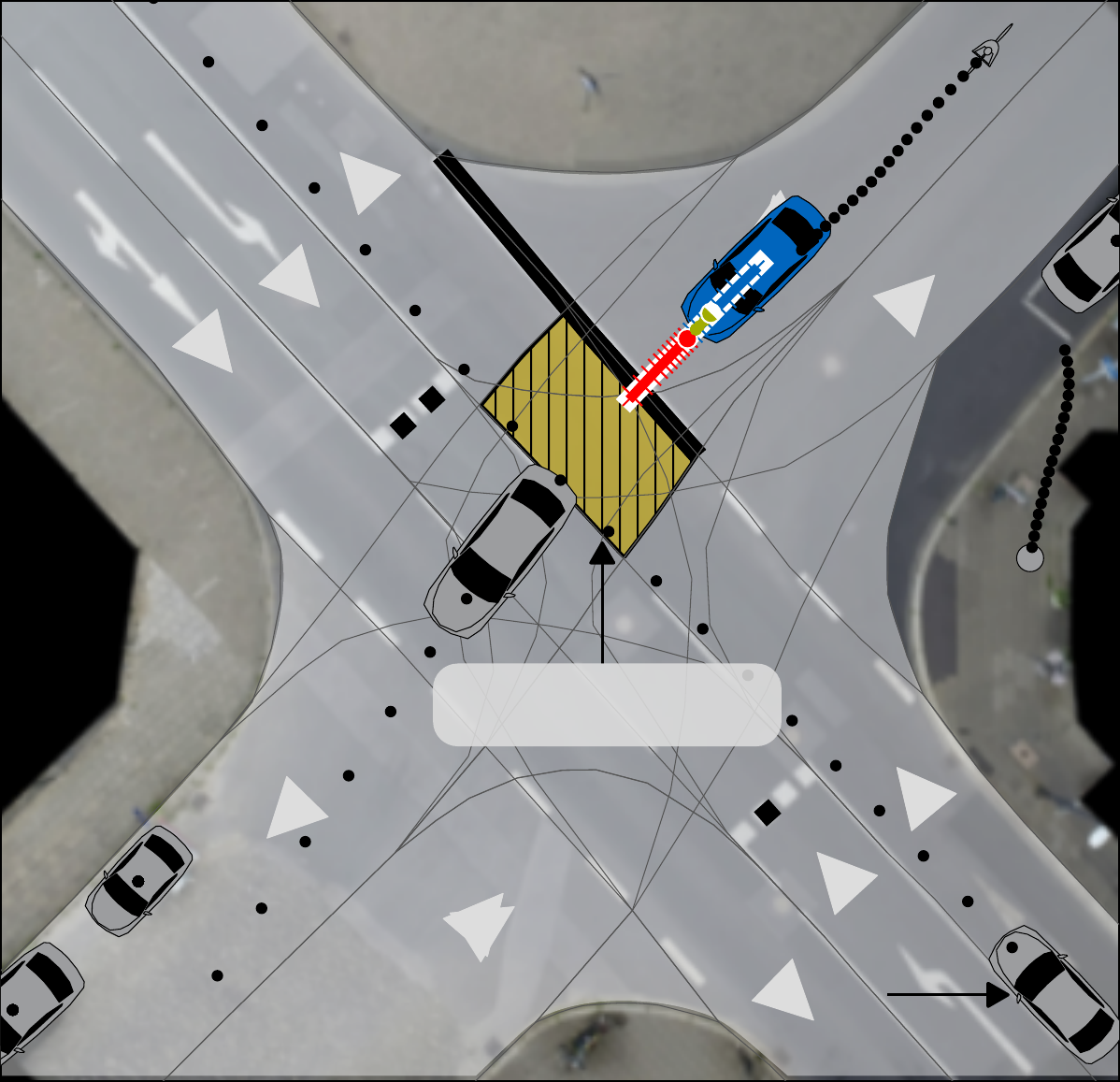}}%
    \put(0.41314636,0.32201419){\color[rgb]{0,0,0}\makebox(0,0)[lt]{\lineheight{1.25}\smash{\begin{tabular}[t]{l}Conflict area\end{tabular}}}}%
    \put(0,0){\includegraphics[width=\unitlength,page=2]{int_in4_ini.pdf}}%
    \put(0.32498943,0.06578502){\color[rgb]{0,0,0}\makebox(0,0)[lt]{\lineheight{1.25}\smash{\begin{tabular}[t]{l}Rule-relevant obstacle\end{tabular}}}}%
  \end{picture}%
\endgroup%

		\caption{Initial and repaired trajectories at $k=0$.}
		\label{fig:rin4_initial}
	\end{subfigure}
	\hfill
	\begin{subfigure}[t]{0.32\linewidth}
		\centering\footnotesize
		\def\svgwidth{0.99\linewidth}
\begingroup%
  \makeatletter%
  \providecommand\color[2][]{%
    \errmessage{(Inkscape) Color is used for the text in Inkscape, but the package 'color.sty' is not loaded}%
    \renewcommand\color[2][]{}%
  }%
  \providecommand\transparent[1]{%
    \errmessage{(Inkscape) Transparency is used (non-zero) for the text in Inkscape, but the package 'transparent.sty' is not loaded}%
    \renewcommand\transparent[1]{}%
  }%
  \providecommand\rotatebox[2]{#2}%
  \newcommand*\fsize{\dimexpr\f@size pt\relax}%
  \newcommand*\lineheight[1]{\fontsize{\fsize}{#1\fsize}\selectfont}%
  \ifx\svgwidth\undefined%
    \setlength{\unitlength}{575bp}%
    \ifx\svgscale\undefined%
      \relax%
    \else%
      \setlength{\unitlength}{\unitlength * \real{\svgscale}}%
    \fi%
  \else%
    \setlength{\unitlength}{\svgwidth}%
  \fi%
  \global\let\svgwidth\undefined%
  \global\let\svgscale\undefined%
  \makeatother%
  \begin{picture}(1,0.96556524)%
    \lineheight{1}%
    \setlength\tabcolsep{0pt}%
    \put(0,0){\includegraphics[width=\unitlength,page=1]{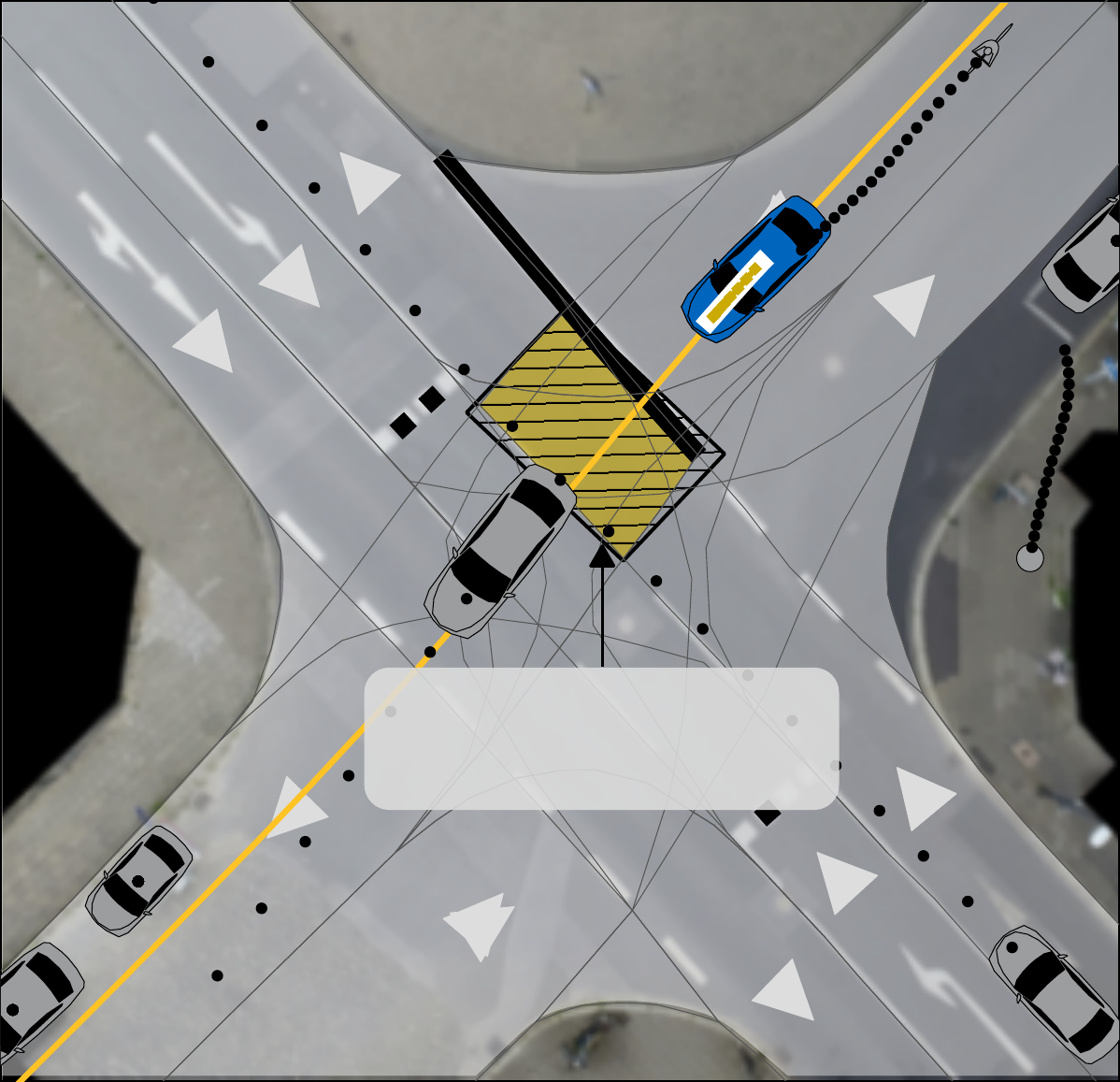}}%
    \put(0.41113466,0.26487563){\color[rgb]{0,0,0}\makebox(0,0)[lt]{\lineheight{1.25}\smash{\begin{tabular}[t]{l}conflict area\end{tabular}}}}%
    \put(0.35254896,0.31414103){\color[rgb]{0,0,0}\makebox(0,0)[lt]{\lineheight{1.25}\smash{\begin{tabular}[t]{l}Overapproximated\end{tabular}}}}%
    \put(0,0){\includegraphics[width=\unitlength,page=2]{int_in4_replan.pdf}}%
    \put(0.24592854,0.06301548){\color[rgb]{0,0,0}\makebox(0,0)[lt]{\lineheight{1.25}\smash{\begin{tabular}[t]{l}$\Gamma$\end{tabular}}}}%
  \end{picture}%
\endgroup%

		\caption{Replanned trajectory using MICP.}
		\label{fig:rin4_replan}
	\end{subfigure}
	\hfill
	\begin{subfigure}[t]{0.32\linewidth}
		\centering\footnotesize
		\def\svgwidth{0.99\linewidth}
\begingroup%
  \makeatletter%
  \providecommand\color[2][]{%
    \errmessage{(Inkscape) Color is used for the text in Inkscape, but the package 'color.sty' is not loaded}%
    \renewcommand\color[2][]{}%
  }%
  \providecommand\transparent[1]{%
    \errmessage{(Inkscape) Transparency is used (non-zero) for the text in Inkscape, but the package 'transparent.sty' is not loaded}%
    \renewcommand\transparent[1]{}%
  }%
  \providecommand\rotatebox[2]{#2}%
  \newcommand*\fsize{\dimexpr\f@size pt\relax}%
  \newcommand*\lineheight[1]{\fontsize{\fsize}{#1\fsize}\selectfont}%
  \ifx\svgwidth\undefined%
    \setlength{\unitlength}{575bp}%
    \ifx\svgscale\undefined%
      \relax%
    \else%
      \setlength{\unitlength}{\unitlength * \real{\svgscale}}%
    \fi%
  \else%
    \setlength{\unitlength}{\svgwidth}%
  \fi%
  \global\let\svgwidth\undefined%
  \global\let\svgscale\undefined%
  \makeatother%
  \begin{picture}(1,0.96556524)%
    \lineheight{1}%
    \setlength\tabcolsep{0pt}%
    \put(0,0){\includegraphics[width=\unitlength,page=1]{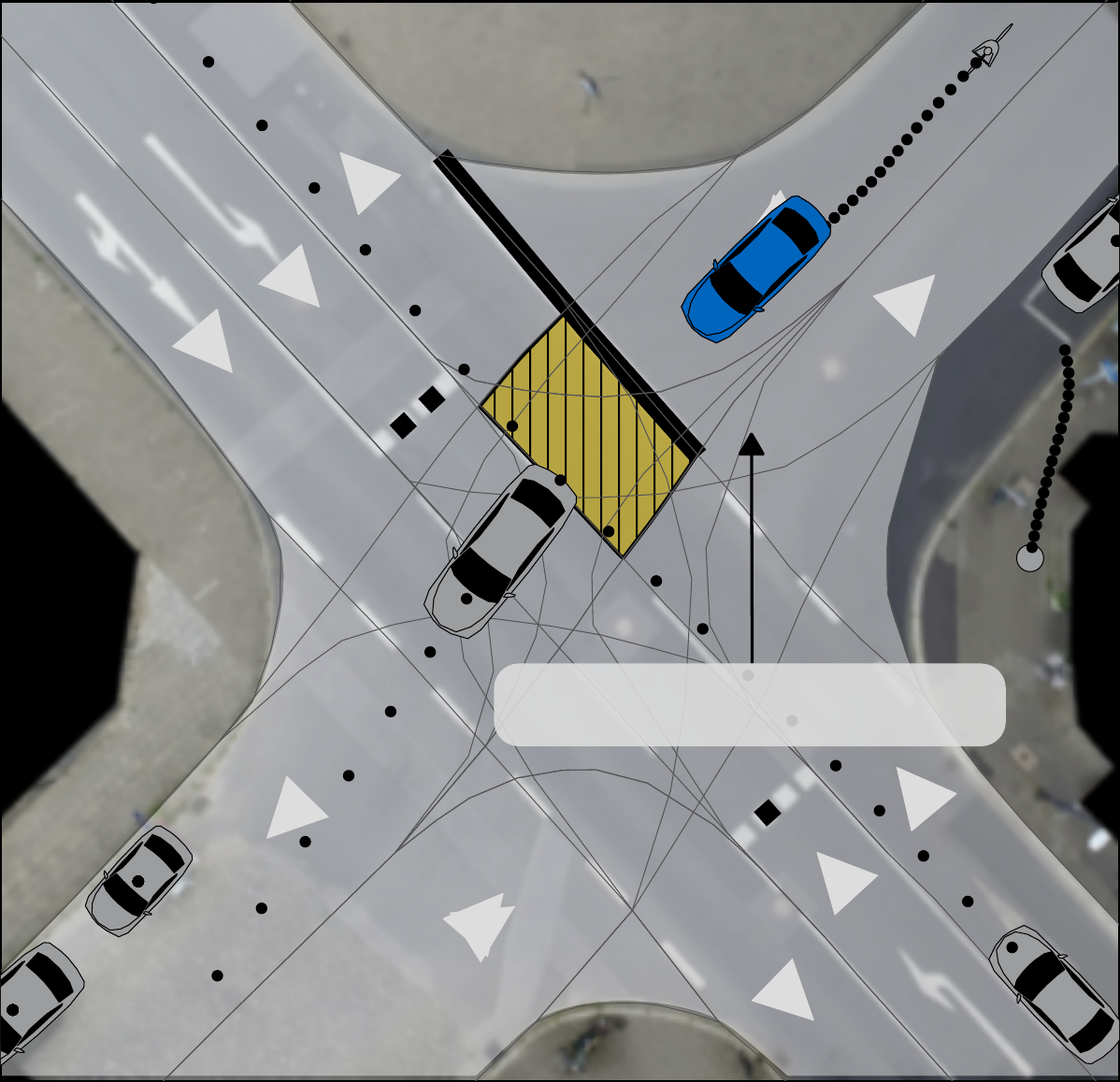}}%
    \put(0.46817854,0.32201419){\color[rgb]{0,0,0}\makebox(0,0)[lt]{\lineheight{1.25}\smash{\begin{tabular}[t]{l}Sampled trajectories\end{tabular}}}}%
    \put(0,0){\includegraphics[width=\unitlength,page=2]{int_in4_replan_sample.pdf}}%
  \end{picture}%
\endgroup%

		\caption{Replanned trajectory using sampling.}\vspace{1.5mm}
		\label{fig:rin4_replan_2}
	\end{subfigure}
	\begin{subfigure}[t]{0.32\linewidth}
		\centering\footnotesize
		\def\svgwidth{0.99\linewidth}
		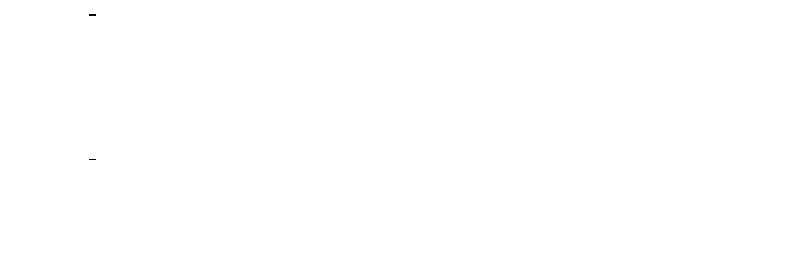
		\caption{Repaired velocity profile.}
		\label{fig:rin4_v}
	\end{subfigure}
	\hfill
	\begin{subfigure}[t]{0.32\linewidth}
		\centering\footnotesize
		\def\svgwidth{0.99\linewidth}
		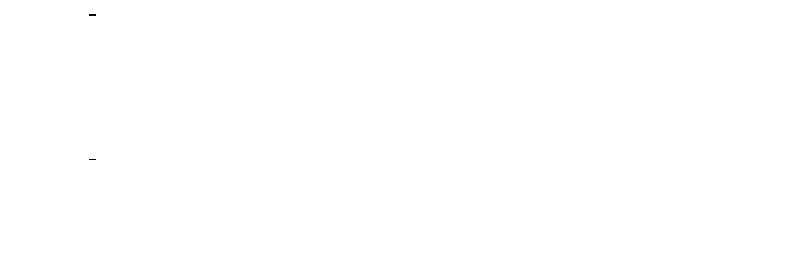
		\caption{Replanned velocity profile using MICP.}
		\label{fig:rin4_v_replan}
	\end{subfigure}
	\hfill
	\begin{subfigure}[t]{0.32\linewidth}
		\centering\footnotesize
		\def\svgwidth{0.99\linewidth}
		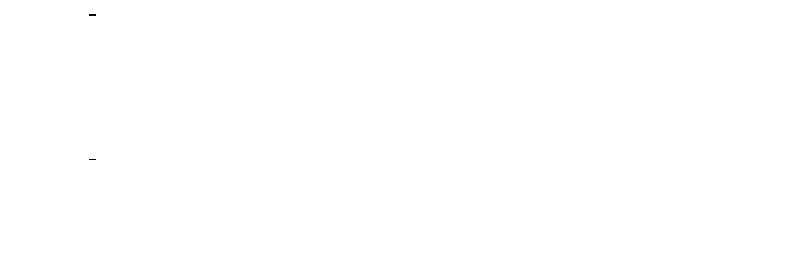
		\caption{Replanned velocity profile using sampling.}
		\label{fig:rin4_v_replan_2}
	\end{subfigure}
	\caption{Intersection scenario where the initially planned trajectory for the ego vehicle violates the priority rule by entering the conflict area without yielding to a higher-priority obstacle. (a) shows the initial positions, and (d) the repaired velocities. (b) and (c) illustrate the replanned trajectories in terms of positions, while (e) and (f) show the corresponding velocities. The same legends from Fig.~\ref{fig:rg13-general} are used for consistency.}
	\vspace{-4mm}
	\label{fig:rin4_ini}
\end{figure*}
\subsubsection{Priority Rule} \label{subsec:priority} Next, we consider a more complex rule -- rural priority rule $\varphi_{\text{IN4}}$ \cite{maierhofer2022formalization}, which prohibits the ego vehicle from entering an intersection if this action would pose a danger to another vehicle with the right of way:  
\footnotetext[9]{ $\mathrm{obs\_straight\_ego\_straight\_obs\_prio}(\boldsymbol{x}_{\mathtt{obs}}, \boldsymbol{x}_{\mathtt{ego}})$ is an abbreviation of the subformula $ \mathrm{going\_straight}(\boldsymbol{x}_{\mathtt{obs}}) \wedge  \mathrm{going\_straight}(\boldsymbol{x}_{\mathtt{ego}}) 
	\wedge \mathrm{has\_priority}(\boldsymbol{x}_{\mathtt{obs}}, \boldsymbol{x}_{\mathtt{ego}})$ from \cite[Tab. VI]{maierhofer2022formalization}, which allows us to make the formula more compact. This abbreviation format similarly applies to all combinations of turning directions.}
\vspace{-2mm}
\begin{align*}
		\varphi_{\text{IN4}} & =
		\\& \mathbf{G}\big((\mathrm{obs\_straight\_ego\_straight\_obs\_prio}(\boldsymbol{x}_{\mathtt{obs}}, \boldsymbol{x}_{\mathtt{ego}})\footnotemark \\ & \vee \mathrm{obs\_straight\_ego\_left\_turns\_obs\_prio}(\boldsymbol{x}_{\mathtt{obs}}, \boldsymbol{x}_{\mathtt{ego}})\\
		& \vee \mathrm{obs\_straight\_ego\_right\_turns\_obs\_prio}(\boldsymbol{x}_{\mathtt{obs}}, \boldsymbol{x}_{\mathtt{ego}})\\
		& \left. \vee \dots \textit{combinations of turning directions}\dots\right)\\
		 & \!\!\!\!\!\! \Rightarrow \left( \mathbf{G}\big(\mathrm{not\_endanger\_intersection}(\boldsymbol{x}_{\mathtt{ego}}, \boldsymbol{x}_{\mathtt{obs}})\big) \right. \\ & \left.\enspace\,\, \lor	\lnot  \mathrm{on\_lanelet\_with\_type\_intersection}(\boldsymbol{x}_{\mathtt{ego}})
		\right) \big),
\end{align*}
\text{with}
	\begin{align*}
		&\mathrm{not\_endanger\_intersection}(\boldsymbol{x}_{\mathtt{ego}}, \boldsymbol{x}_{\mathtt{obs}}) \coloneqq\\
		  &\quad \big(\mathrm{in\_intersection\_conflict\_area}(\boldsymbol{x}_{\mathtt{ego}}, \boldsymbol{x}_{\mathtt{obs}})\\
		  &\enspace\ \Rightarrow (
		  \lnot \mathrm{causes\_braking\_intersection}(\boldsymbol{x}_{\mathtt{ego}}, \boldsymbol{x}_{\mathtt{obs}})\\ & \quad \wedge\lnot \mathbf{F}_{[0, t_{\mathrm{ib}}]}(\mathrm{in\_intersection\_conflict\_area}(\boldsymbol{x}_{\mathtt{obs}},\boldsymbol{x}_{\mathtt{ego}})) )\big)\\
		  & \wedge\big(\mathrm{in\_intersection\_conflict\_area}( \boldsymbol{x}_{\mathtt{obs}}, \boldsymbol{x}_{\mathtt{ego}})\\
		  &\quad \Rightarrow \lnot\mathbf{F}_{[0, t_{\mathrm{ia}}]}(\mathrm{in\_intersection\_conflict\_area}(\boldsymbol{x}_{\mathtt{ego}}, \boldsymbol{x}_{\mathtt{obs}}))
		  \big).
	\end{align*}

\noindent As illustrated in Fig.~\ref{fig:rin4_initial}, the ego vehicle violates $\varphi_{\text{IN4}}$ as its initial trajectory enters the conflict area -- the overlapping region of the lanes -- with the rule-relevant obstacle. Following formula rewriting and distributive decomposition (cf.~Sec.~\ref{sec:rule_monitoring}), we obtain $\varphi^\mathtt{D}_{\text{IN4}}$ as:
\begin{equation*}
\begin{aligned}\label{eq:rule_rin4_p}
	\varphi^\mathtt{D}_{\text{IN4}} \sledom &\ ( \sigma_1 \wedge \sigma_2 \wedge  \sigma_3 \wedge \dots) \vee  (\sigma_{10} \wedge \sigma_{13}) \vee (\sigma_{10} \wedge \sigma_{14}) \\ & \vee (\sigma_{11} \wedge \sigma_{12} \wedge \sigma_{13}) \vee (\sigma_{11} \wedge \sigma_{12} \wedge \sigma_{14}) \vee \sigma_{15},
\end{aligned}
\end{equation*}
with \vspace{-2mm}
\begin{align*}
		\sigma_1 &\coloneqq \mathbf{G}\left(\lnot \mathrm{obs\_straight\_ego\_straight\_obs\_prio}(\cdot)\right), \\
	\sigma_2 &\coloneqq \mathbf{G}\left(\lnot\mathrm{obs\_straight\_ego\_left\_turns\_obs\_prio}(\cdot) \right), \\
		\sigma_3 &\coloneqq \mathbf{G}\left(\lnot\mathrm{obs\_straight\_ego\_right\_turns\_obs\_prio}(\cdot) \right), \\& \dots\\
	\sigma_{10} &\coloneqq \mathbf{G}\left(\lnot\mathrm{in\_intersection\_conflict\_area}(\boldsymbol{x}_{\mathtt{ego}}, \boldsymbol{x}_{\mathtt{obs}}) \right), \\
	\sigma_{11} &\coloneqq \mathbf{G}\left( \lnot \mathrm{causes\_braking\_intersection}(\cdot) \right),\\
	\sigma_{12} &\coloneqq \mathbf{G}\left(\lnot \mathbf{F}_{[0, t_{\mathrm{ib}}]}(\right. \\ & \hspace{1.65cm} \left. \mathrm{in\_intersection\_conflict\_area}(\boldsymbol{x}_{\mathtt{obs}}, \boldsymbol{x}_{\mathtt{ego}})) \right),\\
	\sigma_{13} &\coloneqq \mathbf{G}\left(\lnot\mathrm{in\_intersection\_conflict\_area}(\boldsymbol{x}_{\mathtt{obs}}, \boldsymbol{x}_{\mathtt{ego}}) \right), \\
		\sigma_{14} &\coloneqq \mathbf{G}\left(\lnot \mathbf{F}_{[0, t_{\mathrm{ia}}]}(\right. \\ & \hspace{1.65cm} \left.\mathrm{in\_intersection\_conflict\_area}(\boldsymbol{x}_{\mathtt{ego}}, \boldsymbol{x}_{\mathtt{obs}}) \right),\\
	\sigma_{15} &\coloneqq \mathbf{G}\left(\lnot\mathrm{on\_lanelet\_with\_type\_intersection}(\cdot) \right).
\end{align*}
Next, $\varphi^\mathtt{D}_{\text{IN4}}$ is transformed into  $\varphi^\mathtt{P}_{\text{IN4}}$ in CNF, which serves as the input to the online SAT solver. After detecting the violation, the SAT solver first generates a partial solution $\phi_{\text{IN4}}$ (see Tab.~\ref{tab:prop_robust}),  which serves as input for the $\mathcal{T}$-solver. This solution corresponds to adjustments in the truth valuations of the propositions $\sigma_{10}$ and $\sigma_{14}$. 
After a single iteration, 
our approach successfully enables the ego vehicle to stop and yield to the obstacle with higher priority crossing the same intersection (see Fig.~\ref{fig:rin4_initial} and Fig.~\ref{fig:rin4_v}).

\subsection{Comparison with Related Work}\label{subsec:comp}
To further support our claims in Sec.~\ref{sec:introduction}, we compare our approach with a full trajectory replanning strategy, highlighting the advantages of partially adapting the initial trajectory. {Note that collision avoidance is enforced as a default requirement for all resulting trajectories, unless stated otherwise.}

\subsubsection{Replanning Using Mixed-Integer Convex Programming} The first comparison is with a state-of-the-art temporal logic motion planner\footnote{The implementation is based on \url{https://github.com/vincekurtz/stlpy}.} based on mixed-integer convex programming (MICP) \cite{raman2014model, kurtz2022mixed}, which encodes STL specifications in NNF over convex predicates using binary variables. To align with the repair setup (cf. Sec.~\ref{subsec:vm}), the dynamics of the ego vehicle are modeled as fourth-order integrators for both longitudinal and lateral directions along $\Gamma$. For simple predicates that map state variables to real values, we parameterize them based on the definition provided in \cite[Sec.~IV-C]{Halder2023}, facilitating the formulation of convex constraints. In more complex cases, we employ approximations to evaluate robustness while maintaining the soundness of the approach \cite[Prop. 1]{YuanfeiLinMPR}. For instance, for $\lnot\mathrm{in\_intersection\_conflict\_area}(\cdot)$, we overapproximate the exact conflict area with a polygon defined by its extreme (maximum and minimum) vertex coordinates in the curvilinear coordinate system (cf. Fig.~\ref{fig:rin4_replan}). The robustness  is then computed as the signed distance to its longitudinal and lateral bounds. Furthermore, to primarily investigate the impact of the rules on computation time, we limit the collision avoidance constraints to obstacles within the same lane $\mathcal{L}_{\mathtt{dir}}$ as the ego vehicle.
{To solve the optimization problem, we use the default configuration and the Gurobi solver \cite{gurobi} as provided in \cite{kurtz2022mixed}, which is in line with our setup.}

Fig.~\ref{fig:interstate_replan} and Fig.~\ref{fig:rin4_replan} show the replanned results using MICP for the rules $\varphi_{\text{G1, G3}}$ and $\varphi_{\text{IN4}}$, respectively.
From Tab.~\ref{tab:prop_robust_tc}, it is evident that the runtime increases with the complexity of the formula. This is particularly noticeable for rule $\varphi_{\text{G1, G3}}$, where the nonlinear predicate $\mathrm{keeps\_safe\_distance\_prec}(\cdot)$ is approximated using piecewise-linear functions \cite[Sec.~IV-C]{Halder2023}, resulting in more constraints.  In comparison, our trajectory repair approach consistently produces feasible solutions in significantly less time, regardless of the formula length.
Moreover, similar to (\ref{eq:problem_def}), the MICP optimization problem also minimizes the cost function $J$ and yields velocity profiles comparable to those from trajectory repair. This demonstrates consistency in the outcomes across both approaches (cf.~Fig.~\ref{fig:rg13-general} and Fig.~\ref{fig:rin4_ini}).

\begin{table}[!t]\centering\footnotesize
	\caption{{Performance} comparison between trajectory repair and replanning. Arrows indicate desired performance direction -- $\downarrow$ indicates lower is better, $\uparrow$ indicates higher is better, with the best performance highlighted in {bold}.}
	\renewcommand{\arraystretch}{1.2}
	\begin{tabular}{@{}l@{\hspace{0.8cm}}c@{\hspace{0.7cm}}c@{\hspace{0.8cm}}c@{}} \toprule
		{\textbf{Method}} & {\textbf{Multi-violation}} & {\textbf{Stop-line}} &  {\textbf{Priority}} \\\midrule
		\multicolumn{4}{l}{\textbf{Comp. Time}~$\downarrow$ (single scenario)}  \\
		{Repair} & $\boldsymbol{127}$\unit{ms} &  $\boldsymbol{88}$\unit{ms} &  $\boldsymbol{193}$\unit{ms} \\
		{Replan using MICP} & {$1708$}\unit{ms} & {$699$}\unit{ms} & {$2401$}\unit{ms}\\
		{Replan using sampling} & {$1408$}\unit{ms} & {$218$}\unit{ms} & {$1812$}\unit{ms}\\\midrule
		\multicolumn{4}{l}{{\textbf{Success Rate}~$\uparrow$ ($150$ scenarios)}} \\
		{Repair} & {$\boldsymbol{98.5\%}$} & {$\boldsymbol{98.9\%}$} & {$\boldsymbol{75.0\%}$} \\
		{Replan using MICP} & {$97.7\%$} & {$69.7\%$} & {$33.3\%$} \\
		{Replan using sampling} & {$98.4\%$} & {$67.0\%$} & {$32.4\%$} \\
		\bottomrule
	\end{tabular}\label{tab:prop_robust_tc}\vspace{-4mm}
\end{table}
We further evaluate our approach by comparing it to the MICP-based replanning method across $150$ randomly selected highD and inD scenarios (cf. Sec.~\ref{subsec:scenario}), with the performance depicted in Fig.~\ref{fig:comp_runtime} and {the lower part of Tab.~\ref{tab:prop_robust_tc}}. Our approach demonstrates exceptional efficiency, achieving a runtime that is $90.7\%$ faster than MICP-based replanning (cf. Fig.~\ref{fig:comp_runtime}). This remarkable performance is driven by the decomposition of rules into smaller subformulas and the effective application of convex optimization techniques. Although the runtime varies with the scenario configuration, particularly with the length of the violation segments, i.e., $h - \mathtt{TV}_{\varphi}$, the worst-case runtime for our method remains superior to that of the MICP approach. {Moreover, $66.7\%$ of the MICP replanning attempts for the $\varphi_{\text{IN4}}$ violation fail due to the excessive overapproximation of the conflict area, resulting in an empty solution space (cf.~Tab.~\ref{tab:prop_robust_tc}). A similar failure pattern is observed for $\varphi_{\text{IN1}}$, where the stop line associated with the predicate $\mathrm{stop\_line\_in\_front}(\boldsymbol{x}_{\mathtt{ego}})$ is overapproximated as a polygon aligning with the reference path $\Gamma$. The requirement for convex predicates limits the scalability of MICP-based approaches, as many rule predicates are inherently non-convex.}

\subsubsection{Replanning Using Sampling-Based Planner}\label{subsub:sampling} We adapt the popular sampling-based motion planner\footnote{The code base is obtained from \url{https://commonroad.in.tum.de/tools/commonroad-reactive-planner}.} described in~\cite{werling2012optimal}, which generates, {in parallel}, a finite set of quintic polynomials as candidate trajectories connecting the current state to sampled goal states. The polynomials are then ranked based on the cost function $J$ and {sequentially checked for} feasibility -- including drivability and collision avoidance -- and compliance with the violated rules. The planner terminates upon identifying a feasible solution, which is the trajectory with the minimum cost that satisfies these checks, and returns it as the replanned trajectory. {Note that, apart from adding rule monitoring for each polynomial, all other settings remain unchanged from the original implementation.}
\begin{figure}[t!]
	\vspace{-6mm}
	\centering\footnotesize
	\includesvg[width=0.87\columnwidth]{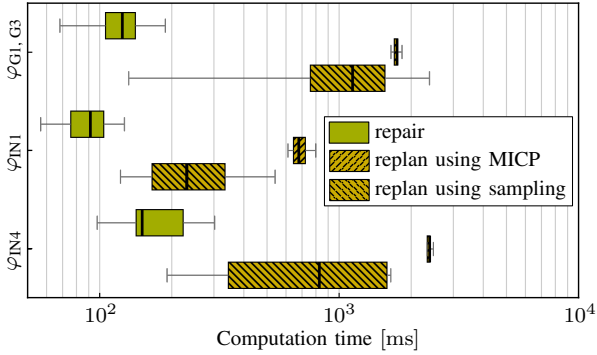}
	\caption{Comparison of runtime performance between our approach and MICP-based and sampling-based replanning over $150$ scenarios. Outliers are omitted from the box plot for enhanced clarity.}	\vspace{-4mm}
	\label{fig:comp_runtime}
\end{figure}

The exemplary replanned results for rules $\varphi_{\text{G1, G3}}$ and $\varphi_{\text{IN4}}$ are shown in Fig.~\ref{fig:interstate_replan_sample} and Fig.~\ref{fig:rin4_replan_2}, respectively, each require over $1\unit{\second}$ (cf. Tab.~\ref{tab:prop_robust_tc}).
For $\varphi_{\text{G1, G3}}$, although a sampled trajectory that maintains the initial velocity is valid (cf. Fig.~\ref{fig:interstate_replan_v_2}), it is not prioritized during the checking process because the high initial acceleration of the ego vehicle is penalized by the cost function.  In the case of  $\varphi_{\text{IN4}}$, the initial use of $270$ samples fails to yield a feasible trajectory, as the ego vehicle is already nearing the conflict area, resulting in a limited solution space. Moreover, the scenario requires the vehicle to decelerate to a full stop; however, the polynomial velocity profile imposes further limitations on the available options (cf. Fig.~\ref{fig:rin4_v_replan_2}). Therefore, the sample number is gradually increased to $5508$ before a solution is found. In contrast, our approach finds feasible trajectories with a stable runtime across all three examples, each taking less than $200\unit{ms}$.

Next, we compare our approach with the sampling-based replanning method across the selected $150$ scenarios. As shown in Fig.~\ref{fig:comp_runtime}, the runtime deviation for the sampling-based approach is significantly greater than that of trajectory repair, exceeding it by an impressive factor of $17.8$.
The comparison clearly illustrates that sampling-based planners often struggle to efficiently find rule-compliant trajectories. This challenge arises primarily from the discretization of the state space and the design of the cost function \cite[(4) and (12)]{werling2012optimal}, {which does not inherently prioritize high-cost maneuvers, such as braking and lane changing.}  In contrast, the runtime of our trajectory repair approach is minimally affected by the cost function selection. 
{In addition, from Tab.~\ref{tab:prop_robust_tc}, we can observe that the sampling-based replanning achieves a success rate comparable to the MICP-based approach, both of which are lower than that of our method. The failures can be attributed to the fact that the sampling-based approach is only resolution-complete. This limitation also makes it difficult to distinguish between the absence of  feasible solutions and a failure to find one.}
{In contrast, the SMT solver provides interpretable outputs, such as reasoned propositions, allowing us to better analyze how the solving process terminates.}
\begin{figure}[t!]
	\centering\footnotesize
	\includesvg[width=0.865\columnwidth]{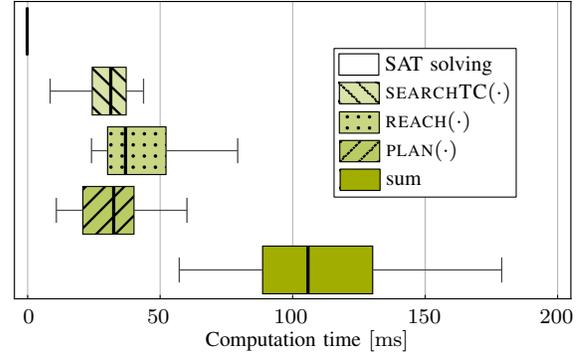}
	\caption{Computation times of our approach over $150$ scenarios, accounting for the total runtime across all iterations within the SMT solver. Outliers are excluded from the box plot for improved clarity.}
	\label{fig:repair_runtime}
	\vspace{-4mm}
\end{figure}

\subsection{Discussions}\label{subsec:disuccsion}
{In this section, we highlight the computational benefits and success rate of our approach, present an ablation study comparing it to our prior work \cite{YuanfeiLin2021, yuanfei2022}, and discuss its limitations.}
\subsubsection{Computation Time}\label{subsec:comp_time}
We illustrate the computation times of components in our trajectory repair approach across the $150$ benchmarks in Fig.~\ref{fig:repair_runtime}. {Since we leverage robustness-guided heuristics within the SMT solver, we primarily report empirical runtime rather than theoretical complexity.} The total mean computation time is $114.2\unit{ms}$, with over $96\%$ of cases staying below $200\unit{ms}$, underscoring its real-time capability. The $\mathtt{TC}_{\phi}$ computation, reachability analysis, and optimization-based trajectory repair account for the majority of the time, with mean values of $37.5\unit{ms}$, $43.4\unit{ms}$, and $33.3\unit{ms}$, respectively. Meanwhile, the SAT solving takes an average of $30\unit{\us}$. This demonstrates that, even for complex rules in critical urban scenarios, our approach maintains efficient performance across all components. To optimize runtime performance, the codebase could be fully implemented in C++; however, the current performance is already sufficient for our test environment.

\begin{figure}[t!]
	\centering
	\begin{subfigure}[t]{0.51\columnwidth}
		\centering\scriptsize
		\includesvg[width=\textwidth]{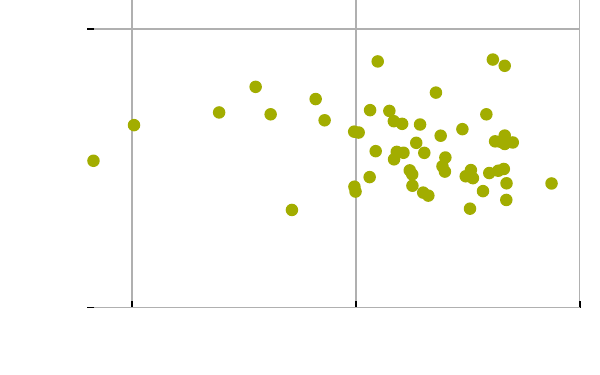}
		\caption{Repair.}
		\label{fig:repair_ttc}
	\end{subfigure}
	\begin{subfigure}[t]{0.455\columnwidth}
		\centering\scriptsize
		\includesvg[width=\textwidth]{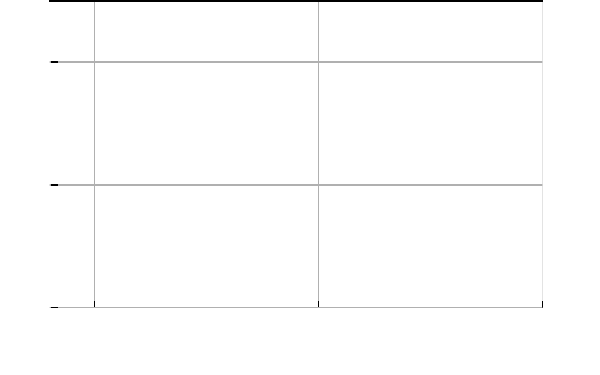}
		\caption{Replan using sampling.}
		\label{fig:replan_ttc}
	\end{subfigure}
	\caption{Relationship between computation time and TTC as an indicator of scenario criticality.}\vspace{-4mm}
	\label{fig:relation_ttc}
\end{figure}

{To highlight the effectiveness of reachability analysis in reducing computation time during critical situations, we compare the runtime performance of the benchmark scenarios between our approach and the sampling-based planner (cf.~Sec.~\ref{subsub:sampling}), using time-to-collision (TTC) \cite{lin2023commonroad} as the criticality measure. The stop-line rule is excluded from this comparison because, unlike TTC, it does not involve interactions with other obstacles. The results in Fig.~\ref{fig:relation_ttc} with linear regression demonstrate that the computation time for our approach decreases as TTC decreases, indicating improved performance in more challenging scenarios. In contrast, the computation time for the sampling-based planner increases as the scenarios become more critical when traffic rules are taken into account.}

\subsubsection{{Success Rate}}\label{subsubsec:success}
{Since we verify the rule compliance of the repaired trajectory before returning it (cf. Sec.~\ref{subsec:repair}), our approach is sound.} 
	In addition, it achieves a high success rate across a diverse set of scenarios and rule types, as shown in Tab.~\ref{tab:prop_robust_tc}.
{The failed repairs are primarily due to three reasons: (i) the scenarios are recorded such that no rule-compliant solution exists from certain initial states, (ii) stricter rule decomposition in (\ref{eq:approxi}), and (iii) the reachable sets are overapproximated. The third issue is particularly prominent for priority rules, where the conflict area between vehicles from different directions (cf.~Fig.~\ref{fig:rin4_ini}) may not align with the reference path for the ego vehicle, causing  looser reachable set approximations. }
\subsubsection{{Ablation Studies}}
{Additionally, we conduct three ablation studies to assess the necessity of individual submodules in our approach and their impact on overall performance, using the same scenarios as in Sec.~\ref{subsec:comp}.}
\paragraph{Model Predictive Robustness}
{Unlike our previous study~\cite{yuanfei2022}, which used model-free robustness~\cite{ruleSTL} as a heuristic for SAT solving, we adopt model predictive robustness in this work (cf. Sec.~\ref{sec:sat_solver}). Tab.~\ref{tab:ablation} shows ablation results for the safe-distance rule $\varphi_{\text{G1}}$ (cf. Sec.~\ref{subsec:multi}), demonstrating the improvements in solver efficiency. With model predictive robustness, we observe a reduced average number of SMT solver iterations, resulting in a lower overall computation time. Furthermore, our approach can be easily extended to other rules without the need to define new robustness functions for additional predicates -- unlike the method in~\cite{yuanfei2022}.
}

\paragraph{Time-To-Comply Computation}
{To showcase the benefits of computing the time-to-comply in the branching mechanism, we list the computation times in Tab.~\ref{tab:ablation} when branching directly from the initial state, i.e., $k_{\mathrm{cut}}$ is always set to $0$ (cf.~(\ref{eq:kprime})). In addition, we evaluate trajectory smoothness using the absolute jerk cost, defined as the sum of the jerk over the entire obtained trajectories.
For multi-rule violations and the stop-line rule, our approach yields better runtime performance and trajectory smoothness. However, for the priority rule, branching from the initial state can result in a slightly more efficient computation compared to our method. This is because the small rule-compliant solution space leads to an exhaustive search for the time-to-comply, which can be beneficial for reachability analysis \cite[Sec. VIII-D]{liu2023specification}.  This observation highlights the need for properly determining $k_{\mathrm{cut}}$, which remains an open direction for future work. In addition, there is no significant difference in trajectory smoothness, as it is already incorporated into the dynamic constraints during the cut-off state computation (cf. Def.~\ref{def:tc}).}

\paragraph{Reachability Analysis} {We compare the use of reachability analysis with the baseline approach that omits it, relying instead on the informed closed-loop RRT (informed CL-RRT) from~\cite{YuanfeiLin2021} to generate repaired trajectories. During the tree expansion, we enforce traffic rule compliance as an additional requirement alongside collision avoidance. As shown in Tab.~\ref{tab:ablation}, the informed CL-RRT approach exhibits longer and less predictable runtimes compared to our method across all rules due to its probabilistic nature. This is particularly evident in scenarios with a smaller solution space, such as those involving violations of priority rules, which consistently result in timeouts. In contrast, our approach leverages reachability analysis to precompute the feasible solution space, enabling more efficient trajectory generation.}
\begin{table}[!t]\centering\footnotesize

		\caption{{Ablation results on the subcomponents of our approach, with the best performance highlighted in bold. Reported times are presented as the mean $\pm$ standard deviation. Arrows indicate that the smaller the value, the better the performance.}}
		\renewcommand{\arraystretch}{1.2}
		\begin{tabular}{@{}l@{\hspace{0.15cm}}cc@{\hspace{0.15cm}}cc@{\hspace{0.15cm}}cc@{}} \toprule
			\textbf{Method} & \multicolumn{3}{@{}c}{\textbf{Num. of iter.}~$\downarrow$} & \multicolumn{3}{@{}c}{\textbf{Comp. time}~$\downarrow$} \\\hline
			\makecell[l]{W/o model predictive rob. \\ for $\varphi_{\text{G1}}$ (versus \cite{yuanfei2022} using \\model-free robustness)} & \multicolumn{3}{@{}c}{${1.13}$} &  \multicolumn{3}{@{}c}{${157}\pm99$\unit{ms}}  \\\hline
			Our approach & \multicolumn{3}{@{}c}{$\boldsymbol{1.08}$} & \multicolumn{3}{@{}c}{$\boldsymbol{127}\pm72$\unit{ms}} \\\midrule
			{\textbf{Comp. time}~$\downarrow$} & \multicolumn{2}{@{}c}{\textbf{Multi-violation}} & \multicolumn{2}{@{}c}{\textbf{Stop-line}} & \multicolumn{2}{@{}c@{}}{\textbf{Priority}} \\\hline
			\makecell[l]{W/o time-to-comply \\(i.e., $k_{\mathrm{cut}} = 0$)} & \multicolumn{2}{@{}c}{{$179\pm34$}\unit{ms}} & \multicolumn{2}{@{}c}{{$115\pm 26$}\unit{ms}} & \multicolumn{2}{@{}c@{}}{{$\boldsymbol{117}\pm78$}\unit{ms}}\\
			\makecell[l]{W/o reachability analysis\footnotemark\\ (versus \cite{YuanfeiLin2021} using informed\\ CL-RRT)} & \multicolumn{2}{@{}c}{{$428\pm405$}\unit{ms}} & \multicolumn{2}{@{}c}{{$492\pm426$}\unit{ms}} & \multicolumn{2}{@{}c@{}}{{timeout}}\\\hline 
			Our approach & \multicolumn{2}{@{}c}{{$\boldsymbol{120}\pm35$}\unit{ms}} & \multicolumn{2}{@{}c}{{$\boldsymbol{91}\pm16$}\unit{ms}} & \multicolumn{2}{@{}c@{}}{{${181}\pm 65$}\unit{ms}}\\\midrule
			{\textbf{Jerk cost}~$\downarrow$} & \multicolumn{2}{@{}c}{\textbf{Multi-violation}} & \multicolumn{2}{@{}c}{\textbf{Stop-line}} & \multicolumn{2}{@{}c@{}}{\textbf{Priority}} \\\hline
			\makecell[l]{W/o time-to-comply \\(i.e., $k_{\mathrm{cut}} = 0$)} & \multicolumn{2}{@{}c}{{$160.9$}} & \multicolumn{2}{@{}c}{{$102.2$}} & \multicolumn{2}{@{}c@{}}{{$\boldsymbol{27.2}$}}\\\hline 
			Our approach & \multicolumn{2}{@{}c}{{$\boldsymbol{123.7}$}} & \multicolumn{2}{@{}c}{{$\boldsymbol{90.6}$}} & \multicolumn{2}{@{}c@{}}{$34.1$}\\
			\bottomrule
		\end{tabular}\label{tab:ablation}
	\vspace{-3mm}
\end{table}
\footnotetext{
	Results over $1\unit{s}$ are excluded as timeouts. }
\subsubsection{{Limitations}}
{
	The SMT solver and branching mechanism in trajectory repair may introduce suboptimality compared to fully
	replanning the trajectory. This suboptimality can be managed by incorporating optimality considerations into the SAT solving process, carefully designing the cost function \cite[Rem.~4]{werling2012optimal}, and applying more restrictive input constraints in the time-to-comply computation. 
	Another limitation is the completeness due to the overapproximation discussed in Sec.~\ref{subsubsec:success}, which leads to false positives when identifying repaired trajectories. {One possible remedy, if time permits, is to reduce conservatism by relaxing the strictness of the rule decomposition and improving the accuracy of the reachable set overapproximations.} However, in practice, obtaining a viable solution efficiently is often prioritized. In all failure cases, a fail-safe mechanism is essential to guarantee safety, particularly in highly dynamic environments~\cite{pek2020fail}.} 
\begin{figure}[t!]
	\centering
	\def\svgwidth{0.99\columnwidth}
	\begin{subfigure}[t]{\columnwidth}
		\centering\scriptsize
		\def\svgwidth{\linewidth}
		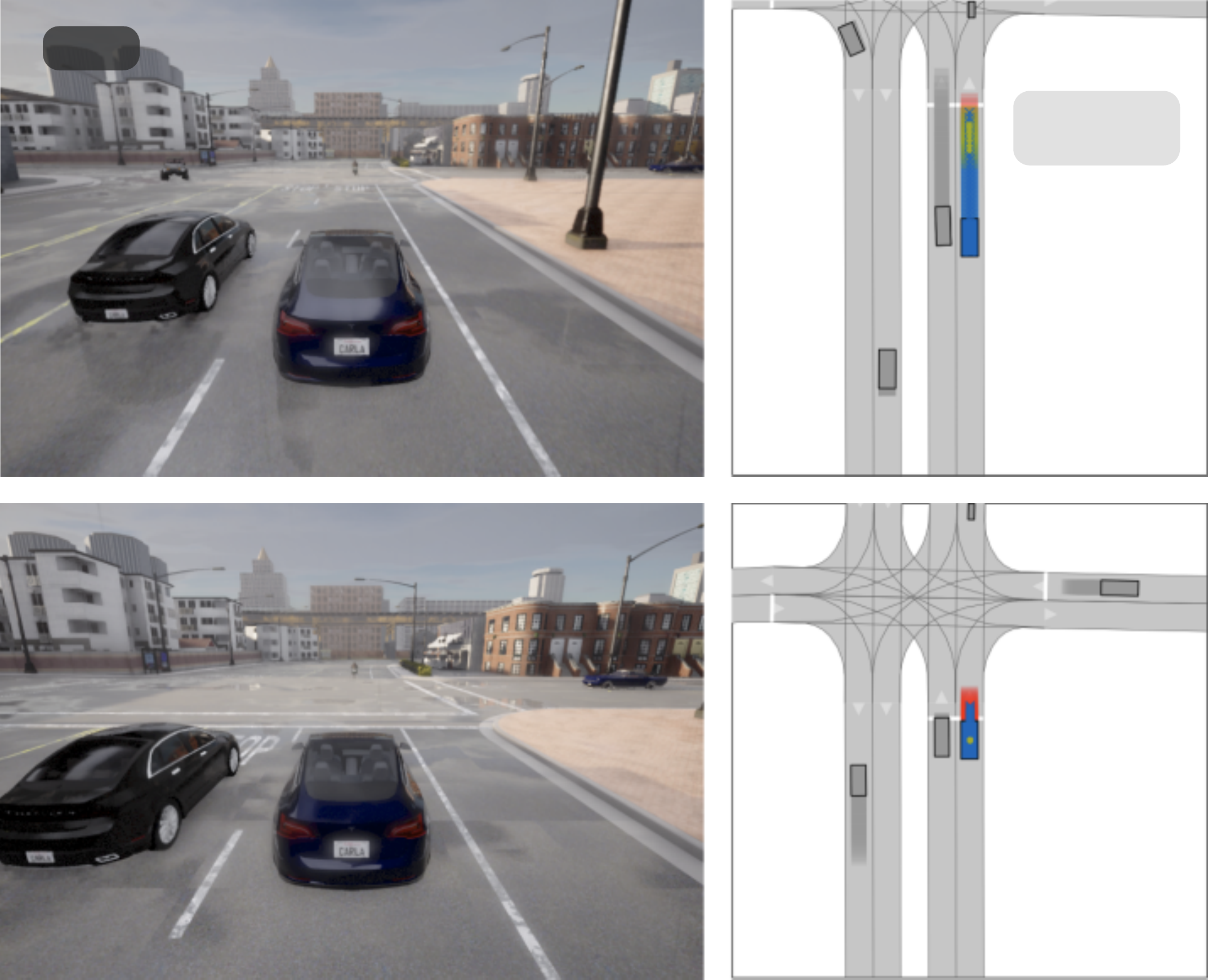
		\caption{CARLA and CommonRoad scenarios at two time steps.}
		\label{fig:carla_over}
	\end{subfigure}
	\begin{subfigure}[t]{\columnwidth}
		\centering\scriptsize\vspace{2mm}
		\includesvg[width=\textwidth]{./figures/carla_rin1_measure_test.svg}\vspace{-1mm}
		\caption{Nominal and measurement data.}
		\label{fig:carla_in1_measure}
	\end{subfigure}
	\caption{Repair for the stop-line rule, where the ego vehicle successfully waits in front of the stop line for  $3\unit{s}$. (a) shows wide-angle snapshots from the CARLA simulator focusing on the ego vehicle, with the CommonRoad scenarios visualized in a bird's-eye view on the right-hand side. (b) presents selected experimental data, where $y$ denotes the vertical Cartesian position.}\vspace{-3mm}
	\label{fig:CARLA_RIN1}
\end{figure}

\begin{figure}[t!]
	\centering
	\def\svgwidth{0.99\columnwidth}
	\begin{subfigure}[t]{\columnwidth}
		\centering\scriptsize
		\def\svgwidth{\linewidth}
\begingroup%
  \makeatletter%
  \providecommand\color[2][]{%
    \errmessage{(Inkscape) Color is used for the text in Inkscape, but the package 'color.sty' is not loaded}%
    \renewcommand\color[2][]{}%
  }%
  \providecommand\transparent[1]{%
    \errmessage{(Inkscape) Transparency is used (non-zero) for the text in Inkscape, but the package 'transparent.sty' is not loaded}%
    \renewcommand\transparent[1]{}%
  }%
  \providecommand\rotatebox[2]{#2}%
  \newcommand*\fsize{\dimexpr\f@size pt\relax}%
  \newcommand*\lineheight[1]{\fontsize{\fsize}{#1\fsize}\selectfont}%
  \ifx\svgwidth\undefined%
    \setlength{\unitlength}{1029.27499846bp}%
    \ifx\svgscale\undefined%
      \relax%
    \else%
      \setlength{\unitlength}{\unitlength * \real{\svgscale}}%
    \fi%
  \else%
    \setlength{\unitlength}{\svgwidth}%
  \fi%
  \global\let\svgwidth\undefined%
  \global\let\svgscale\undefined%
  \makeatother%
  \begin{picture}(1,0.39459988)%
    \lineheight{1}%
    \setlength\tabcolsep{0pt}%
    \put(0,0){\includegraphics[width=\unitlength,page=1]{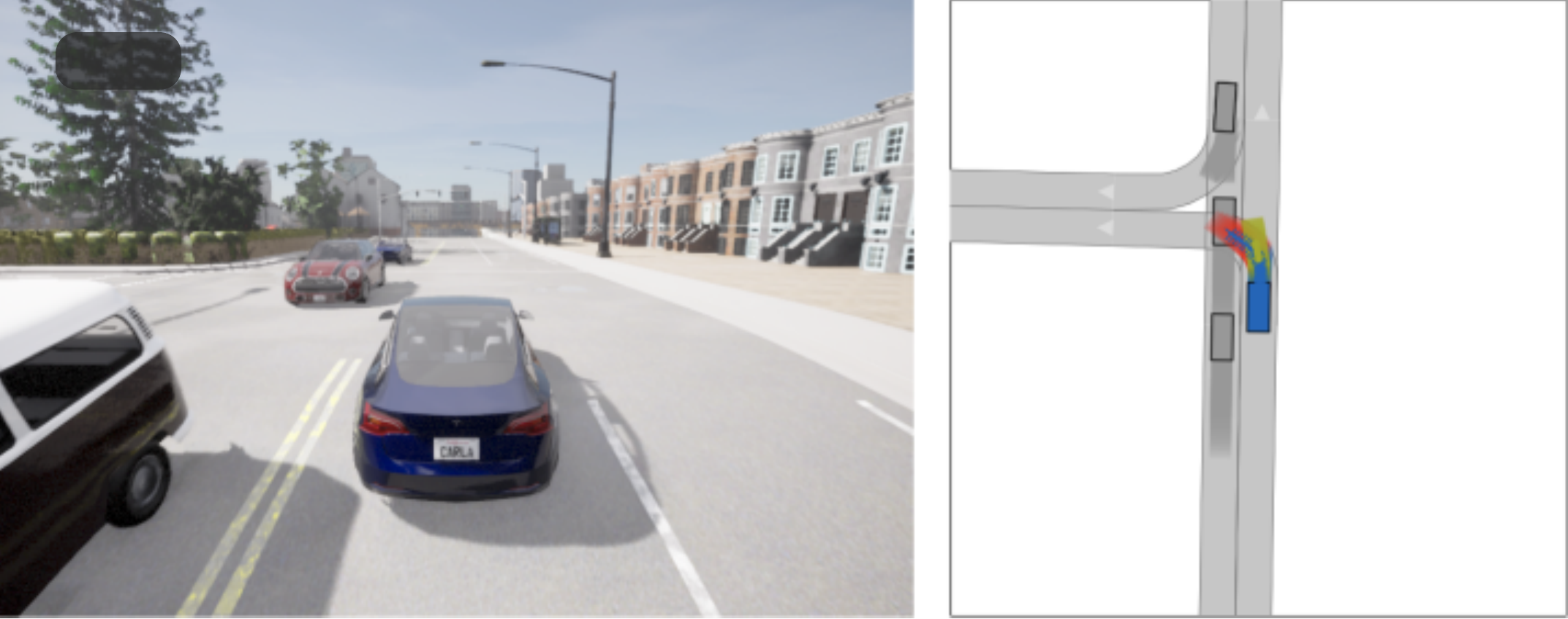}}%
    \put(0.04823012,0.3490913){\color[rgb]{1,1,1}\makebox(0,0)[lt]{\lineheight{1.25}\smash{\begin{tabular}[t]{l}$7.5\unit{s}$\end{tabular}}}}%
    \put(0,0){\includegraphics[width=\unitlength,page=2]{carla_rin5.pdf}}%
  \end{picture}%
\endgroup%

		\caption{CARLA and CommonRoad scenarios at the repair time step.}
		\label{fig:carla_over_rin5}
	\end{subfigure}
	\begin{subfigure}[t]{\columnwidth}
		\centering\scriptsize
		\vspace{2mm}
		\includesvg[width=\linewidth]{./figures/carla_rin5_measure_test.svg}
		\caption{Nominal and measurement data.}
		\label{fig:carla_in5_measure}
	\end{subfigure}
	\caption{Repair for the left-turn rule, where the ego vehicle successfully yields to another obstacle inside the intersection. (a) shows a snapshot from the CARLA simulator and CommonRoad scenario, with legends matching those in Fig.~\ref{fig:CARLA_RIN1}. (b) presents experimental data for the ego vehicle, where $x$ is the horizontal Cartesian position, and $\theta$ is the orientation.}
	\vspace{-2mm}
	\label{fig:CARLA_RIN5}
\end{figure}
\subsection{CARLA Simulation}\label{subsec:carla} 
We now integrate the proposed trajectory repair approach into the CARLA simulator \cite{Dosovitskiy17}, employing the planning algorithm described in \cite{kochdumper2024real} as the nominal planner. The prediction module utilizes a constant acceleration model to forecast the future trajectories of surrounding traffic participants. Specifically, the planning algorithm combines a decision module based on reachability analysis with the solution of an optimal control problem to generate feasible trajectories. However, it explicitly accounts for only a limited subset of traffic rules, such as speed limits and traffic lights. 
For our experiments, we use the urban downtown environment with CARLA map ID Town~03 and configure the traffic manager in autopilot mode to simulate realistic urban traffic scenarios. The planned trajectories are continuously monitored against formalized urban traffic rules \cite{maierhofer2022formalization} and are repaired by our approach whenever a violation occurs. To address discrepancies between the high-fidelity vehicle model in CARLA and the simplified vehicle models used in both the nominal planner and our repair approach, we incorporate a feedback controller to compensate for model uncertainties and disturbances. 

Fig.~\ref{fig:CARLA_RIN1} and Fig.~\ref{fig:CARLA_RIN5} display the repair results in the CARLA simulator for the stop-line rule $\varphi_{\text{IN1}}$ (cf. Sec.~\ref{subsec:stop_line}) and the left-turn rule $\varphi_{\text{IN5}}$, respectively. The latter rule restricts the ego vehicle from making a left turn without priority unless it can safely enter the oncoming lane without posing a risk to approaching vehicles. For a detailed formalization of $\varphi_{\text{IN5}}$, we refer readers to  \cite[Tab.~VI]{maierhofer2022formalization}. 
Without requiring any tuning of the nominal planner, we adapt the planned trajectory to a rule-compliant solution space through consecutive trajectory repair, even when the nominal planner consistently tends to violate the rule (cf. Fig.~\ref{fig:carla_in1_measure}). These results highlight the robustness of our approach in a closed-loop environment, emphasizing its readiness for real-world deployment.

\subsection{Real-World Vehicle Deployment}\label{subsec:realworld} 

\begin{figure}[t!]
	\centering
	\begin{subfigure}[!t]{\columnwidth}
		\centering\scriptsize
		\def\svgwidth{\linewidth}
		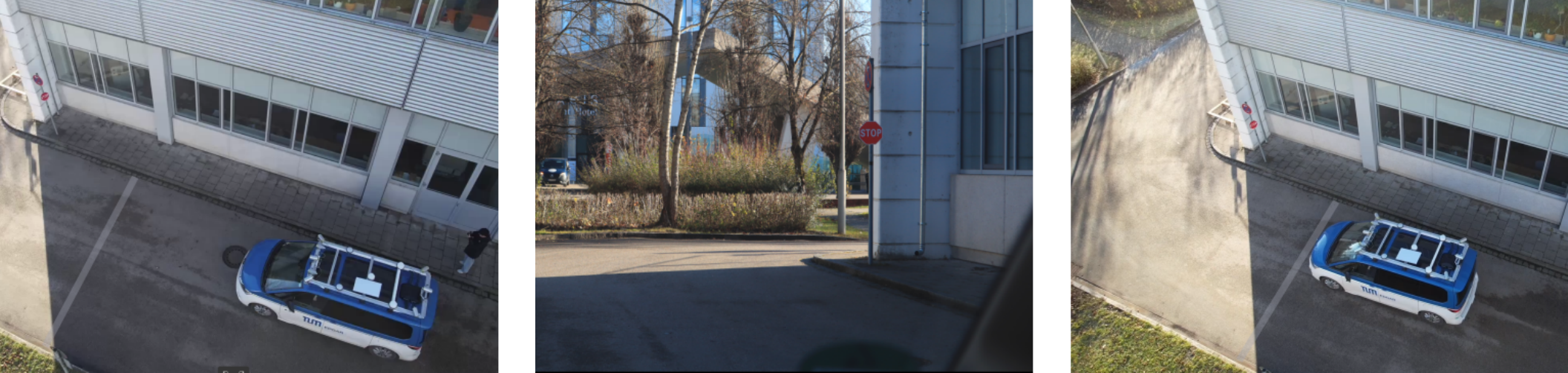
		\caption{Camera images of the experiment.}\vspace{1.5mm}
		\label{fig:autoware_images}
	\end{subfigure}
	\begin{subfigure}[t]{\columnwidth}
		\centering\scriptsize
		\def\svgwidth{\linewidth}
\begingroup%
  \makeatletter%
  \providecommand\color[2][]{%
    \errmessage{(Inkscape) Color is used for the text in Inkscape, but the package 'color.sty' is not loaded}%
    \renewcommand\color[2][]{}%
  }%
  \providecommand\transparent[1]{%
    \errmessage{(Inkscape) Transparency is used (non-zero) for the text in Inkscape, but the package 'transparent.sty' is not loaded}%
    \renewcommand\transparent[1]{}%
  }%
  \providecommand\rotatebox[2]{#2}%
  \newcommand*\fsize{\dimexpr\f@size pt\relax}%
  \newcommand*\lineheight[1]{\fontsize{\fsize}{#1\fsize}\selectfont}%
  \ifx\svgwidth\undefined%
    \setlength{\unitlength}{656.55914547bp}%
    \ifx\svgscale\undefined%
      \relax%
    \else%
      \setlength{\unitlength}{\unitlength * \real{\svgscale}}%
    \fi%
  \else%
    \setlength{\unitlength}{\svgwidth}%
  \fi%
  \global\let\svgwidth\undefined%
  \global\let\svgscale\undefined%
  \makeatother%
  \begin{picture}(1,0.36668448)%
    \lineheight{1}%
    \setlength\tabcolsep{0pt}%
    \put(0,0){\includegraphics[width=\unitlength,page=1]{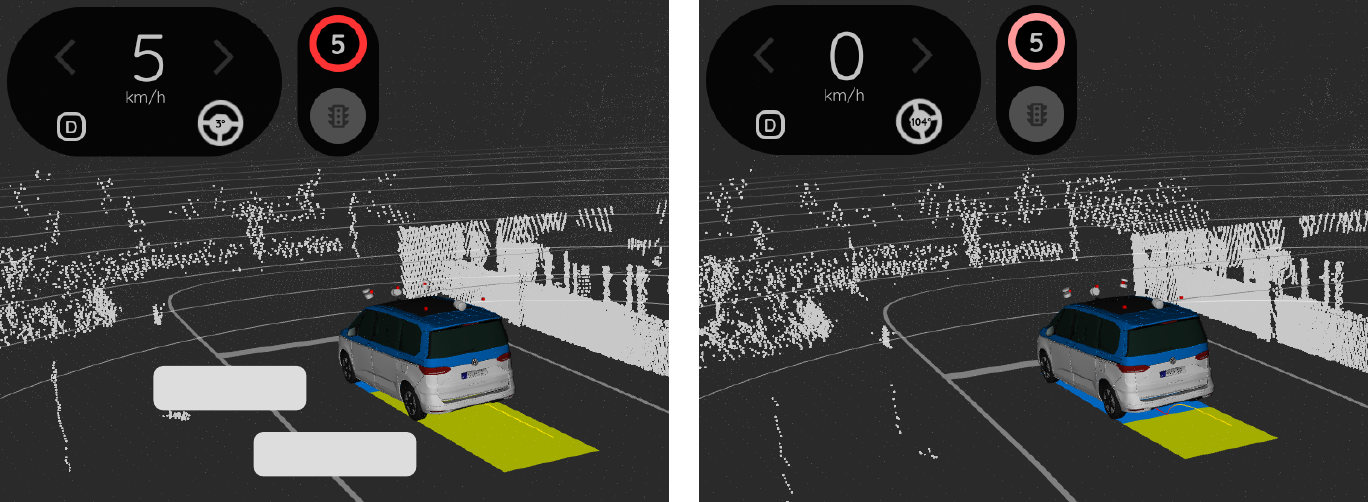}}%
    \put(0.12326139,0.07540642){\color[rgb]{0,0,0}\rotatebox{-0.09351281}{\makebox(0,0)[lt]{\lineheight{1.25}\smash{\begin{tabular}[t]{l}Planned\end{tabular}}}}}%
    \put(0.19522592,0.02706152){\color[rgb]{0,0,0}\rotatebox{-0.09351281}{\makebox(0,0)[lt]{\lineheight{1.25}\smash{\begin{tabular}[t]{l}Repaired\end{tabular}}}}}%
    \put(0,0){\includegraphics[width=\unitlength,page=2]{autoware_repair.pdf}}%
    \put(0.40971585,0.32600445){\color[rgb]{1,1,1}\makebox(0,0)[lt]{\lineheight{1.25}\smash{\begin{tabular}[t]{l}$60\unit{s}$\end{tabular}}}}%
    \put(0.92169233,0.32600445){\color[rgb]{1,1,1}\makebox(0,0)[lt]{\lineheight{1.25}\smash{\begin{tabular}[t]{l}$67\unit{s}$\end{tabular}}}}%
    \put(0,0){\includegraphics[width=\unitlength,page=3]{autoware_repair.pdf}}%
  \end{picture}%
\endgroup%

		\caption{Autoware scenarios at two time steps.}\vspace{.5mm}
		\label{fig:autoware_repair}
	\end{subfigure}
	\begin{subfigure}[t]{\columnwidth}
		\centering\scriptsize
		\vspace{2mm}
		\includesvg[width=\textwidth]{./figures/Autoware_data_test.svg}
		\caption{Nominal and measurement data where $a$ is the acceleration.}
		\label{fig:autoware_measure}
	\end{subfigure}
	\caption{Situation from our real-world experiments where the nominal planner for the ego vehicle neglects the stop-line rule, with its trajectory continuously monitored and effectively repaired by our approach.}	\vspace{-4mm}
	\label{fig:aw}
\end{figure}

We integrate our approach into the EDGAR research vehicle \cite{karle2023edgar}, a Volkswagen T7 Multivan equipped with the necessary sensors and hardware for fully autonomous test runs. Fig.~\ref{fig:aw} illustrates the real-world test, showcasing a stop line in the driveway of the ego vehicle.  The repair algorithm is incorporated via the Autoware Universe middleware software stack \cite{kato2018autoware}, with the nominal planner provided by the CommonRoad-Autoware interface toolbox \cite{wursching2024simplifying}, which does not explicitly account for any formalized traffic rules. Trajectories generated by the repairer, if they violate a rule, or the nominal planner, if they do not, are then communicated to the controller through ROS2 \cite{macenski2022robot}. This setup allows the repair approach to operate within the same software environment as previous experiments. 
As shown in Fig.~\ref{fig:autoware_measure}, through the application of our repair algorithm, the ego vehicle stops precisely in front of the marked stop line for approximately $5\unit{s}$. 
The duration exceeds the required $3\unit{s}$ (cf. Tab.~\ref{tab:params}) due to the heavy computational load of the sensor components, which slows down the planning-monitoring-repair pipeline. While reducing $t_\mathrm{slw}$ in the rule formula (cf. (\ref{eq:rule_rin1})) or optimizing the entire software stack could mitigate this issue, these aspects are beyond the scope of our work.
 Additionally, we observe similar driving behavior in the real-world deployment and the CARLA simulator, with differences primarily in the robustness of the controller (cf.~Fig.~\ref{fig:carla_in1_measure} and Fig.~\ref{fig:autoware_measure}). Therefore, we can rely on CARLA for further development before deploying the repair algorithm in real-world scenarios.

\section{Conclusions}
\label{sec:conclusions}

To the best of our knowledge, we present the first trajectory repair framework for automated vehicles designed to ensure compliance with any formalized traffic rules, provided a feasible solution exists. Our approach modularizes rule formulas and leverages SMT to determine repair strategies, while reachability analysis prunes the search space after branching from the initial trajectory, enhancing computational efficiency. Although incorporating if-else conditions into the planning module can manage specific traffic rules, they become impractical for comprehensive compliance across all traffic situations.  Our adaptive repair mechanism, independent of the nominal planner, systematically rectifies rule-violating trajectories, similar to how public monitoring systems detect and address violations to uphold order on roads. 
Through comparisons with state-of-the-art methods and extensive experiments in challenging rules and real-world conditions,  we believe our approach marks a significant advancement in enhancing safety and fostering public trust in automated driving.

\section*{Acknowledgment}
The authors gratefully acknowledge the financial support provided by the German Research Foundation (DFG) under grants AL 1185/7-1 and AL 1185/20-1. The authors also extend their gratitude to the EDGAR team for the complete development of the vehicle and their valuable support during the driving experiments. Special thanks go to M. Wetzlinger for his insightful feedback on earlier drafts of this article and to A. Li for conducting the drone filming for the experiments.

\bibliographystyle{IEEEtran}
{\bibliography{repair}}

\begin{IEEEbiography}[{\includegraphics[width=1in,height=1.25in,clip,keepaspectratio]{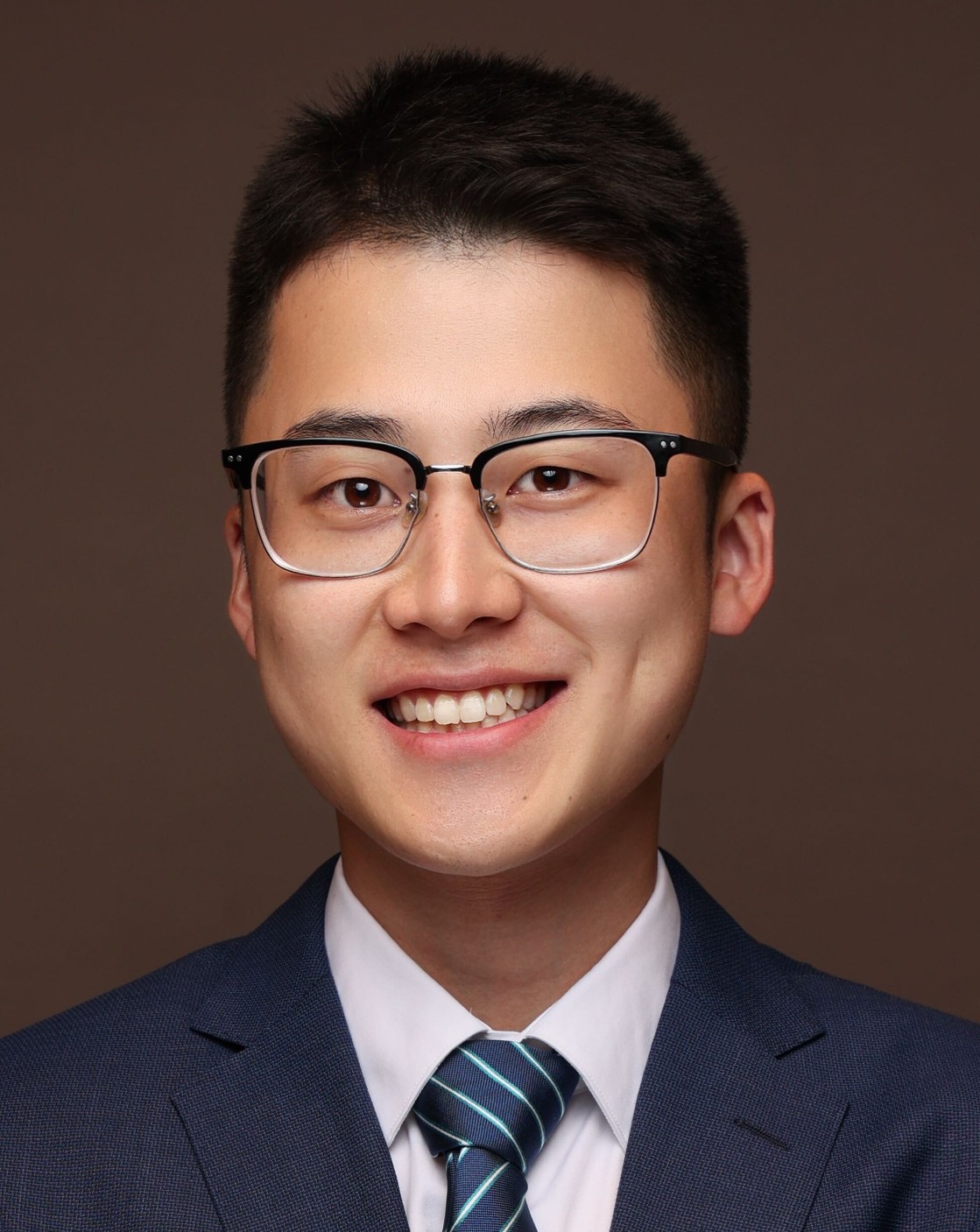}}]{Yuanfei Lin} 	received his B.Eng. degree in Automotive Engineering from Tongji University, China, in 2018, and dual M.Sc. degrees in Mechanical Engineering and Mechatronics and Robotics at the Technical University of Munich, Germany, in 2020.  In 2023, he was a visiting scholar at the University of California, Berkeley, USA. He completed his Ph.D. in Computer Science at the Technical University of Munich, Germany, in 2025. His research interests include motion planning, formal methods, and large language models for automated vehicles.
\end{IEEEbiography}
\begin{IEEEbiography}[{\includegraphics[width=1in,height=1.25in,clip,keepaspectratio]{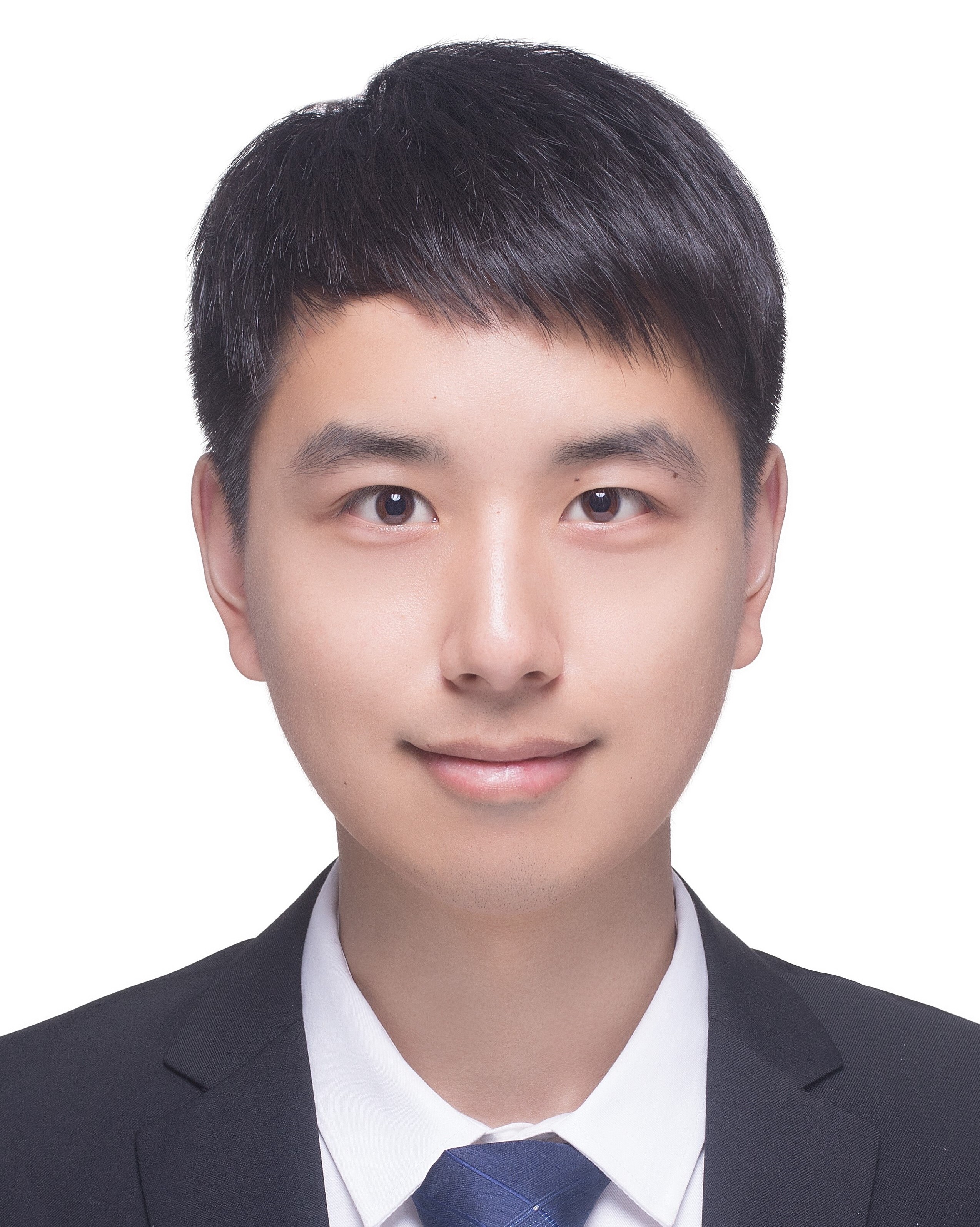}}]{Zekun Xing} is currently a Ph.D. student in the Chair of Automatic Control Engineering at the Technical University of Munich, Germany. He received his B.Eng. degree in Automotive Engineering and Service from Tongji University, China, in 2021, and M.Sc. degree in Mechatronics and Robotics at the Technical University of Munich, Germany, in 2023. His research interests include motion prediction and planning for autonomous vehicles, decision-making in interactive driving scenarios, and model predictive control with application in automated driving.
\end{IEEEbiography}
\begin{IEEEbiography}[{\includegraphics[width=1in,height=1.25in,clip,keepaspectratio]{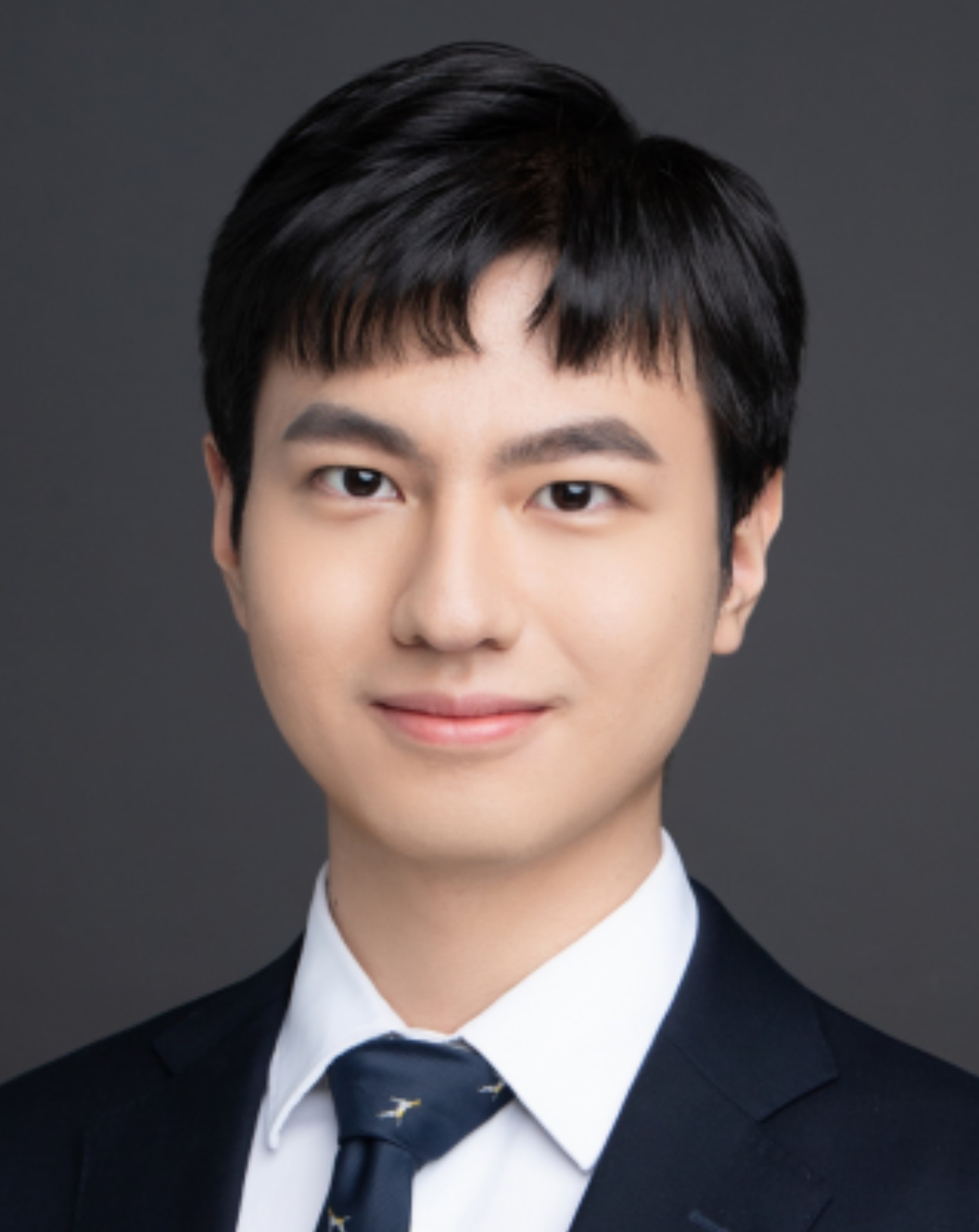}}]{Xuyuan Han} 
	received his B.Eng. degree in Mechatronics from Tongji University, China, in 2022, and M.Sc. degree in Robotics, Cognition, Intelligence at the Technical University of Munich, Germany, in 2025. His research interests include motion planning for autonomous vehicles, the integration of vision-language models in end-to-end autonomous driving systems, and embodied intelligence.
\end{IEEEbiography}
\begin{IEEEbiography}[{\includegraphics[width=1in,height=1.25in,clip,keepaspectratio]{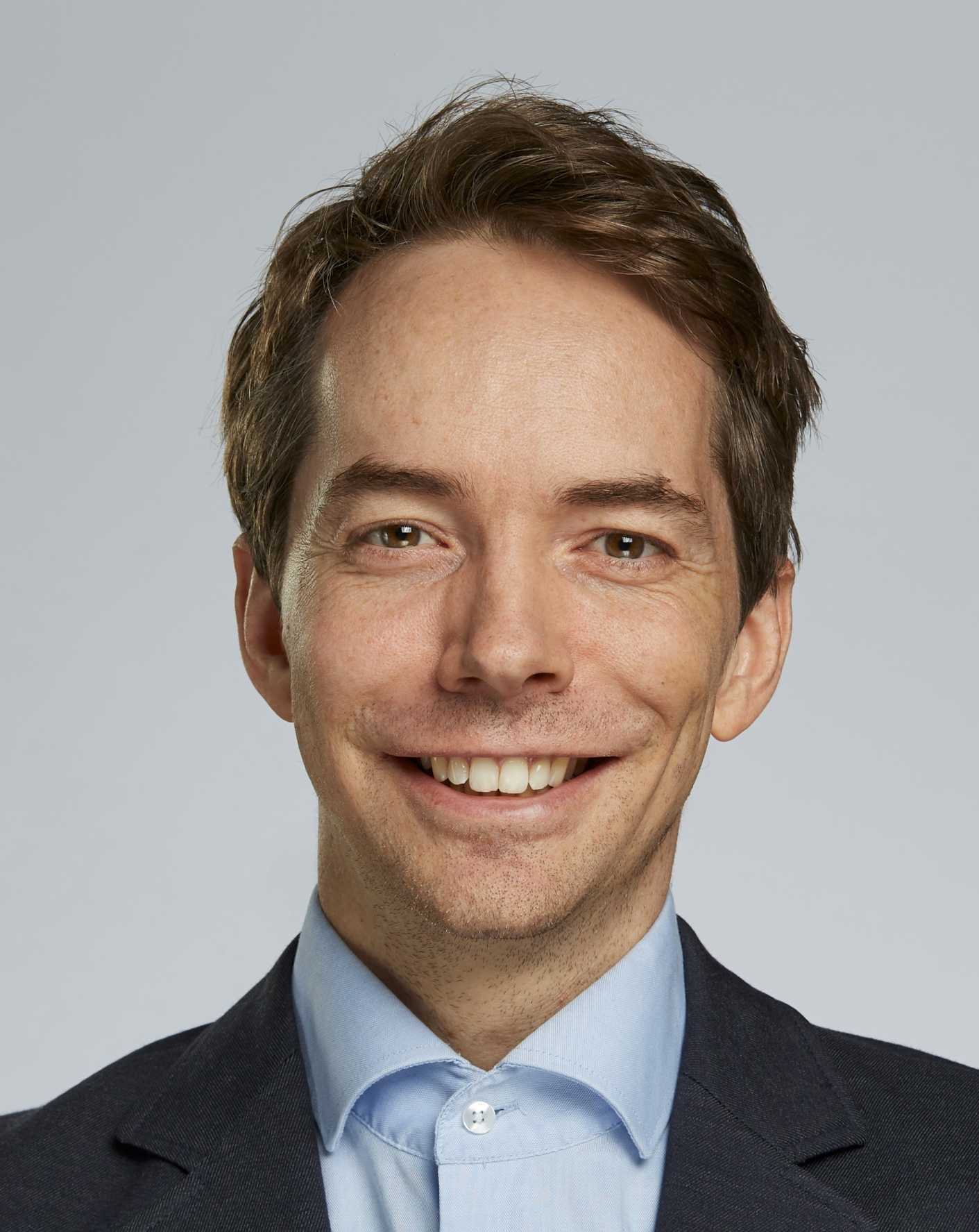}}]
	{Matthias Althoff} is an associate professor in computer science at the Technical University of Munich, Germany. He received his diploma engineering degree in Mechanical
	Engineering in 2005, and his Ph.D. degree in Electrical Engineering in
	2010, both from the Technical University of Munich, Germany.
	From 2010 to 2012 he was a postdoctoral researcher at Carnegie Mellon University,
	Pittsburgh, USA, and from 2012 to 2013 an assistant professor at Technische Universit\"at Ilmenau, Germany. His research interests include formal verification of continuous and hybrid systems, reachability analysis, planning algorithms, nonlinear control, automated vehicles, and power systems.
\end{IEEEbiography}

\vfill

\end{document}